\RequirePackage{booktabs} 

\documentclass[sn-basic]{sn-jnl}

\usepackage{natbib}
\usepackage{array}
\usepackage{pgfplots}
\usepackage{caption} 
\usepackage{hyperref}
\usepackage{makecell}
\usepackage{hhline}
\usepackage{colortbl}
\usepackage{tabularx}
\usepackage{subcaption}
\usetikzlibrary{patterns}
\usepackage{natbib}

\usepackage{graphicx}%
\usepackage{multirow}%
\usepackage{amsmath,amssymb,amsfonts}%
\usepackage{amsthm}%
\usepackage{mathrsfs}%
\usepackage[title]{appendix}%
\usepackage{textcomp}%
\usepackage{manyfoot}%
\usepackage{booktabs}%
\usepackage{algorithm}%
\usepackage{algorithmicx}%
\usepackage{algpseudocode}%
\usepackage{listings}%

\usepackage{lscape}

\newif\ifshowrevision 
\showrevisionfalse 

\ifshowrevision
  \newcommand{\revision}[1]{\textcolor{red}{#1}} 
\else
  \newcommand{\revision}[1]{#1} 
\fi

\newif\ifshowrevisionB 
\showrevisionBfalse 

\ifshowrevisionB
  \newcommand{\revisionB}[1]{\textcolor{red}{#1}} 
\else
  \newcommand{\revisionB}[1]{#1} 
\fi

\theoremstyle{thmstyleone}%
\theoremstyle{thmstyletwo}%

\theoremstyle{thmstylethree}%

\begin{document}

\title{A Comprehensive Survey of \revision{Deep Learning for} Time Series Forecasting: Architectural Diversity and Open Challenges}


\author[1,3]{\fnm{Jongseon} \sur{Kim}}\email{hallooawooye@snu.ac.kr}
\equalcont{}

\author[1,4]{\fnm{Hyungjoon} \sur{Kim}}\email{khjn81@snu.ac.kr}
\equalcont{These authors contributed equally to this work.}

\author[2]{\fnm{HyunGi} \sur{Kim}}\email{rlagusrl0128@snu.ac.kr}

\author[1]{\fnm{Dongjun} \sur{Lee}}\email{elite1717@snu.ac.kr}

\author*[1, 2]{\fnm{Sungroh} \sur{Yoon}}\email{sryoon@snu.ac.kr}

\affil[1]{\orgdiv{Interdisciplinary Program in Artificial Intelligence}, \orgname{Seoul National University}}

\affil[2]{\orgdiv{Department of Electrical and Computer Engineering}, \orgname{Seoul National University}}

\affil[3]{\orgdiv{R\&D Department}, \orgname{LG Chem}}

\affil[4]{\orgdiv{R\&D Department}, \orgname{Samsung SDI}}



\abstract{
Time series forecasting is a critical task that provides key information for decision-making across various fields, such as economic planning, supply chain management, and medical diagnosis.
After the use of traditional statistical methodologies and machine learning in the past, various fundamental deep learning architectures such as MLPs, CNNs, RNNs, and GNNs have been developed and applied to solve time series forecasting problems.
However, the structural limitations caused by the inductive biases of each deep learning architecture constrained their performance.
Transformer models, which excel at handling long-term dependencies, have become significant architectural components for time series forecasting.
However, recent research has shown that alternatives such as simple linear layers can outperform Transformers. These findings have opened up new possibilities for using diverse architectures\revision{, ranging from fundamental deep learning models to emerging architectures and hybrid approaches}.
In this context of exploration into various models, the architectural modeling of time series forecasting has now entered a renaissance.
This survey not only provides a historical context for time series forecasting but also offers comprehensive and timely analysis of the movement toward architectural diversification.
By comparing and re-examining various deep learning models, we uncover new perspectives and presents the latest trends in time series forecasting, including the emergence of hybrid models, diffusion models, Mamba models, and foundation models.
By focusing on the inherent characteristics of time series data, we also address open challenges that have gained attention in time series forecasting, such as channel dependency, distribution shift, causality, and feature extraction.
This survey explores vital elements that can enhance forecasting performance through diverse approaches.
\revision{These contributions help lower entry barriers for newcomers by providing a systematic understanding of the diverse research areas in time series forecasting (TSF), while offering seasoned researchers broader perspectives and new opportunities through in-depth exploration of TSF challenges.}
}




\maketitle

\section{Introduction}\label{sec1}


\begin{figure}[!h]
\centering
\includegraphics[width=1.0\textwidth, trim=0cm 0cm 0cm 0cm, clip]{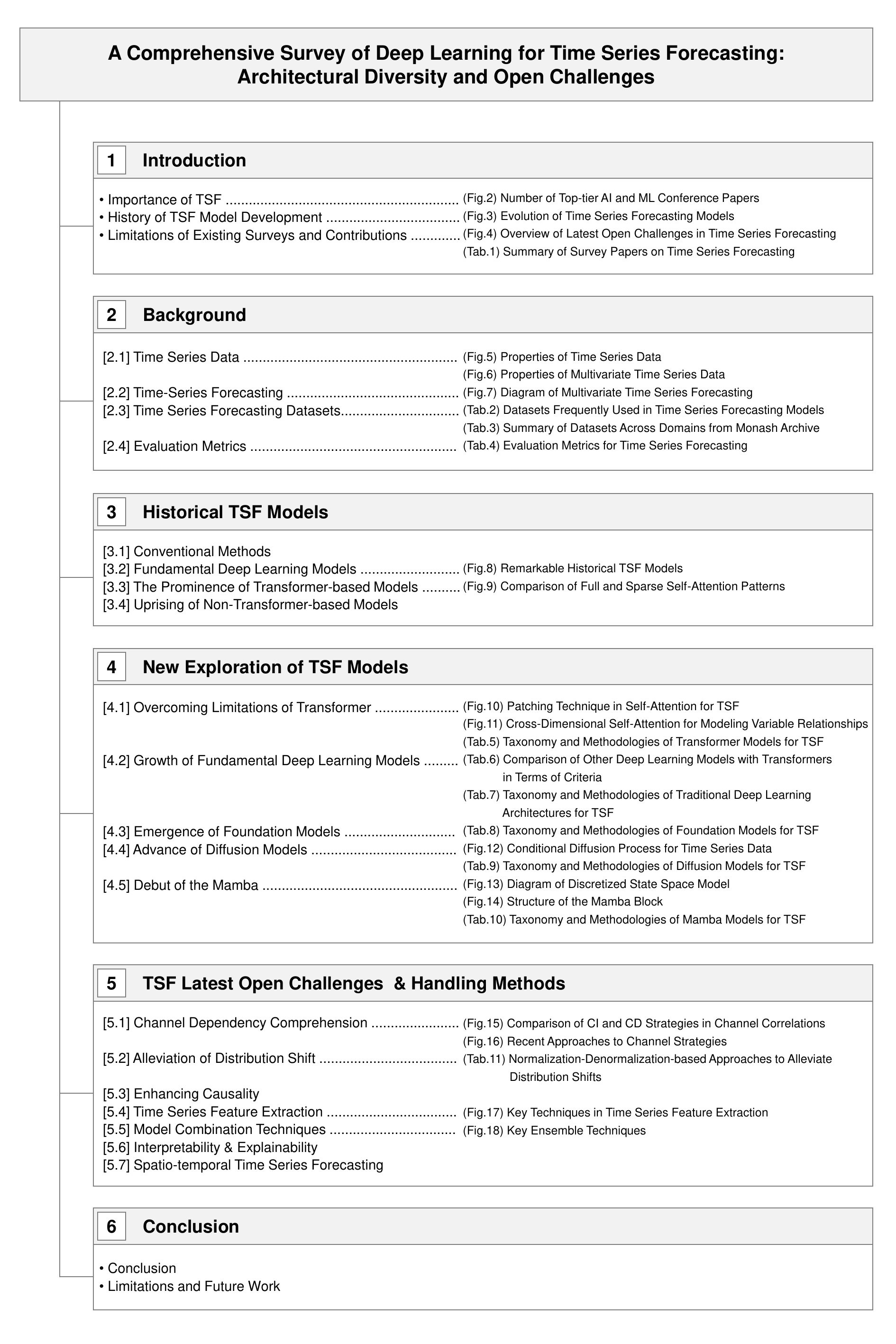}
\caption{Overview of This Survey}
\label{OverviewOfThisSurvey}

\begin{picture}(0,0)

    \put(175,604){\makebox(0,0)[c]{\textbf{\fontsize{8}{12}\selectfont(Section \ref{sec1})}}}
    \put(175,529){\makebox(0,0)[c]{\textbf{\fontsize{8}{12}\selectfont(Section \ref{sec2})}}}
    \put(175,433){\makebox(0,0)[c]{\textbf{\fontsize{8}{12}\selectfont(Section \ref{sec3})}}}
    \put(175,358){\makebox(0,0)[c]{\textbf{\fontsize{8}{12}\selectfont(Section \ref{sec4})}}}
    \put(175,204){\makebox(0,0)[c]{\textbf{\fontsize{8}{12}\selectfont(Section \ref{sec5})}}}
    \put(175,86){\makebox(0,0)[c]{\textbf{\fontsize{8}{12}\selectfont(Section \ref{sec6})}}}
\end{picture}
\end{figure}

Time series forecasting (TSF) is a task that predicts future values based on sequential historical data \citep{cryer1986time}.
It is utilized as a key decision-making tool in various fields, such as economics and finance, supply chain management, transportation, energy, weather, and healthcare \citep{DANESE2011204, IntroductionToFinancialForecasting, trafficForecasting, ElectricityForecasting, climateForecasting, healthForecasting}.
Such applications offer various opportunities, including cost reduction, increased efficiency, and enhanced competitiveness \citep{DANESE2011204}.
\revision{The diversity and complexity of time series data} make forecasting challenging.
In addition to apparent information, \revision{hidden patterns and irregular values pose challenges in learning temporal dependencies. \citep{contiformer}}

In multivariate problems, additional factors such as channel correlation make the task even more difficult (Section \ref{sec2-1}).
Furthermore, time series data exhibits different characteristics depending on the domain, and the various times and environments in which the series is collected result in significantly different patterns (Section \ref{sec2-3}).
Because of this, TSF problems exhibit limited model generalizability, requiring diverse architectures and approaches.
The increasingly complicated TSF problems are presenting researchers with growing challenges, which has recently led to the active development of new methodologies and algorithms to address these issues \citep{TSFwithDL}.

\revision{TSF is closely connected to core AI challenges, and solving these problems plays a crucial role in advancing the field of AI.
For instance, the generalization of AI models is essential for ensuring stable performance on new data.
However, time series data presents unique characteristics and domain-specific patterns that make generalization particularly challenging \citep{han2023improving}.
Solutions to generalization in TSF provide new insights that are applicable to other fields, such as Natural Language Processing (NLP) and Computer Vision (CV).
The distribution of real-world time series data can change over time, and models must effectively handle these changes (Section \ref{sec5-2}). TSF plays a crucial role in addressing distribution shift issues, and the techniques developed to solve these problems can extend to other AI domains, such as image, video, and NLP.
Moreover, attempts to understand models are crucial for improving their trustworthiness in real-world applications \citep{han2023impact}.
In mission-critical fields such as healthcare and finance, the explainability of time series prediction models is essential.
For example, explaining how stock price predictions or patient vital sign forecasts are made can help users better understand and trust the model's decisions.
Thus, TSF research in explainability provides practical solutions for other AI applications (Section \ref{sec5-6}).
Recently, TSF has evolved beyond simple forecasting, establishing itself as a practical platform for addressing key AI challenges.
Solving these problems within TSF highlights the potential for AI to develop in a complementary and synergistic way.
}

\revision{Fig. \ref{Fig_TSFpapers} shows the number of papers on time series forecasting presented at major AI and machine learning conferences. This explosive increase demonstrates the growing importance of TSF research in the AI and machine learning fields.
}
As various studies addressing time series forecasting problems are being actively conducted, survey papers are also being frequently published.
Over time, numerous survey papers have systematically organized the vast landscape of TSF, offering in-depth research that has provided valuable guidance and direction for researchers.
However, existing survey papers still have room for improvement, particularly in addressing the inevitable increase in model diversity and the open challenges in the field\revision{, as summarized in Table \ref{tab:SurveyList}}.
\revision{This survey aims to reorganize the extensive discourse on TSF by providing essential and comprehensive information.
At the same time, it explores model diversity to deliver the latest TSF trends and uniquely highlights open challenges, clearly presenting them to readers.}


\medskip
\bigskip

\small
\begin{center}

\begin{tikzpicture}
    \centering
    \begin{axis}[
        width=0.8\textwidth,
        height=0.45\textwidth,
        ylabel={Number of Papers},
        y label style={yshift=-1em}, 
        symbolic x coords={2020, 2021, 2022, 2023, 2024},
        xtick=data,
        ybar stacked,
        bar width=25pt,
        ymin=0,
        enlarge x limits=0.1,
        ytick pos=left,
        xtick pos=left,
        ymajorgrids=true, 
        axis line style={gray!100, thick}, 
        legend style={at={(0,0.605)}, 
        anchor=south west, 
        font=\scriptsize, 
        cells={anchor=west}, 
        fill=white}, 
        nodes={scale=0.8, transform shape},
        every axis plot/.append style={
            fill opacity=1,
            draw=black,
            thin,
            preaction={fill=white} 
        },
        area legend,
        tick style={gray!100},
        label style={gray!100},
        axis on top=false 
    ]

        \addplot [fill={rgb: red,254; green,25; blue,25}] coordinates {(2020,4) (2021,3) (2022,7) (2023,8) (2024,28)}; 
        \addplot [fill=orange] coordinates {(2020,1) (2021,1) (2022,2) (2023,5) (2024,13)}; 
        \addplot [fill=yellow] coordinates {(2020,0) (2021,2) (2022,1) (2023,2) (2024,13)}; 
        \addplot [fill={rgb: red,60; green,180; blue,75}] coordinates {(2020,3) (2021,5) (2022,3) (2023,7) (2024,7)}; 
        \addplot [fill={rgb: red,0; green,130; blue,200}] coordinates {(2020,0) (2021,1) (2022,4) (2023,4) (2024,8)}; 
        \addplot [fill={rgb: red,145; green,30; blue,180}] coordinates {(2020,2) (2021,1) (2022,4) (2023,3) (2024,12)}; 
        
        \legend{
            NeurIPS: {\small Conference on Neural Information Processing Systems}, 
            ICLR: {\small International Conference on Learning Representations}, 
            ICML: {\small International Conference on Machine Learning}, 
            AAAI: {\small Association for the Advancement of Artificial Intelligence}, 
            IJCAI: {\small International Joint Conference on Artificial Intelligence}, 
            KDD: {\small Knowledge Discovery and Data Mining}
        }
    \end{axis}
\end{tikzpicture}


    \smallskip
    
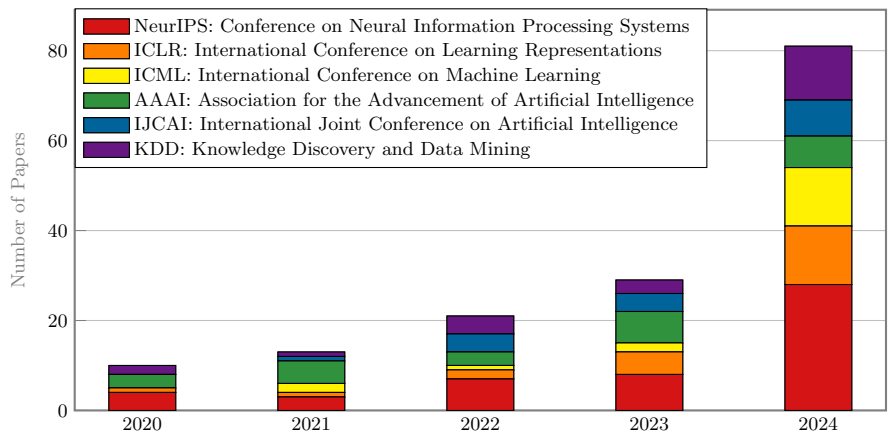
\captionof{figure}{
    Number of Top-tier AI and ML Conference Papers on Time Series Forecasting
    }

    \label{Fig_TSFpapers}
\end{center}

\normalsize

\begin{figure}[]
\centering
\rotatebox{90}{ 
    \begin{minipage}{\textheight}
    \captionsetup{justification=raggedright, singlelinecheck=false} 
    \caption{Evolution of Time Series Forecasting Models}
    \label{Evolution}
    \vspace{0.5cm} 
    \includegraphics[width=\textheight, trim=0cm 5.1cm 0cm 5.1cm, clip, keepaspectratio]{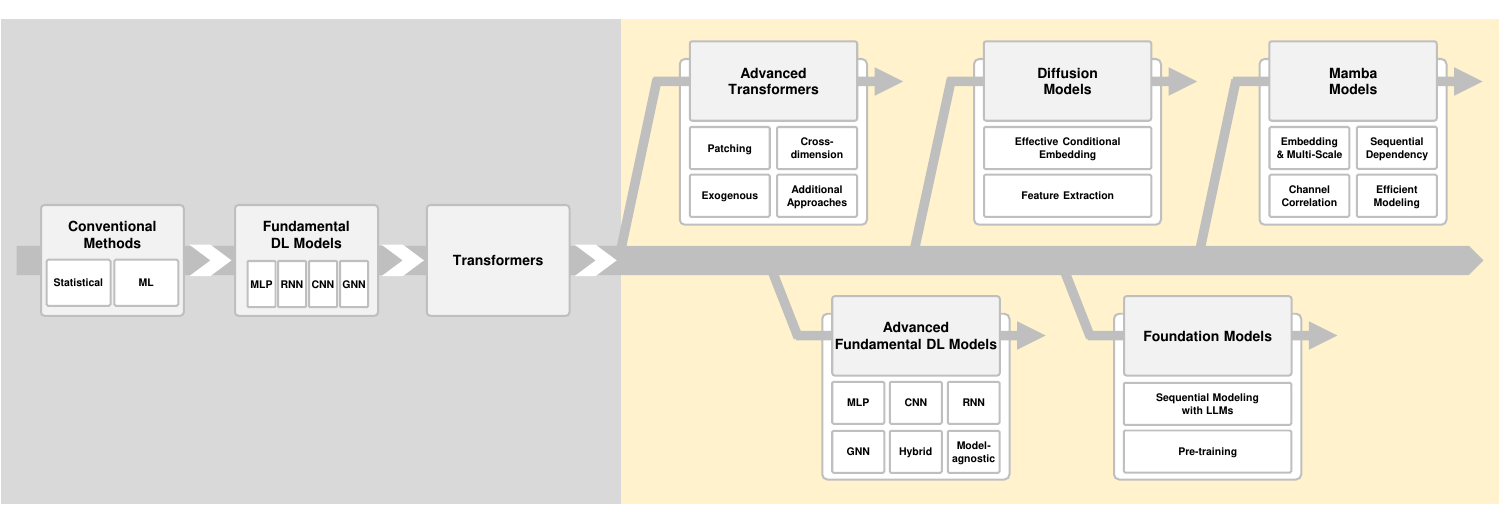}

    \begin{picture}(0,0)
        \put(53,104){\makebox(0,0)[c]{\textbf{\fontsize{6}{10}\selectfont(Section \ref{sec3-1})}}}
        \put(143,104){\makebox(0,0)[c]{\textbf{\fontsize{6}{10}\selectfont(Section \ref{sec3-2})}}}
        \put(233,104){\makebox(0,0)[c]{\textbf{\fontsize{6}{10}\selectfont(Section \ref{sec3-3})}}}
        
        \put(360,206){\makebox(0,0)[c]{\textbf{\fontsize{6}{10}\selectfont(Section \ref{sec4-1})}}}
        \put(427,87){\makebox(0,0)[c]{\textbf{\fontsize{6}{10}\selectfont(Section \ref{sec4-2})}}}
        \put(498,206){\makebox(0,0)[c]{\textbf{\fontsize{6}{10}\selectfont(Section \ref{sec4-3})}}}
        \put(562,91){\makebox(0,0)[c]{\textbf{\fontsize{6}{10}\selectfont(Section \ref{sec4-4})}}}
        \put(632,206){\makebox(0,0)[c]{\textbf{\fontsize{6}{10}\selectfont(Section \ref{sec4-5})}}}
    \end{picture}
    
    \end{minipage}
}
\end{figure}

Models for TSF have undergone various stages of development over an extended period.
In the past, statistical methods based on moving averages were predominantly used, which later evolved into traditional approaches such as Exponential Smoothing and ARIMA \citep{box1970forecasting}.
Machine learning techniques such as Tree Models \citep{quinlan1986induction} and Support Vector Machines (SVM) \citep{cortes1995support} have also been frequently used, but they had limitations in learning complex nonlinear patterns (Section \ref{sec3-1}).
With the increase in available data and advancements in hardware computing power, various deep learning architectures such as MLPs \citep{MLP}, RNNs \citep{RNN}, CNNs \citep{LeNet}, and GNNs \citep{GNN} were developed, enabling the learning of more complex patterns.
However, the performance of these early deep learning architectures was constrained by their intrinsic designs.
To overcome these structural limitations, variants such as Long Short-Term Memory (LSTM) \citep{LSTM} and Temporal Convolutional Networks (TCN) \citep{TCN} have been widely utilized (Section \ref{sec3-2}).
Transformers \citep{Transformer}, known for their ability to handle long-term dependencies, have demonstrated excellent performance in natural language processing and have been naturally extended to time series data as well.
While Transformers have shown good performance in TSF and become widely popular, recent cases have shown that simple linear models can outperform Transformer models (Section \ref{sec3-3}).
As a result, there has been a significant increase in reconsidering fundamental deep learning methodologies, along with growing interest in various architectures such as foundation models, diffusion models, and Mamba models (Section \ref{sec4-2}, \ref{sec4-3}, \ref{sec4-4}, \ref{sec4-5}).
The Transformer model continues to improve in performance and still plays a significant role (Section \ref{sec4-1}).
In this way, TSF has entered a renaissance of modeling, with various methodologies actively competing without being dominated by any single approach (Fig. \ref{Evolution}).
In this context, this survey offers two major strengths that set it apart from previous TSF survey papers.

\normalsize

First, we focus on the inevitable diversification of architectures, providing a timely and comprehensive view to understand the current trend of architecture diversification.
Existing TSF survey papers, such as \citep{TrasformersInTS, LLMforTSsurvey, DiffForTS, RiseofDiff, mamba360}, focus on providing detailed analyses of specific architectures but have limitations when it comes to broadly comparing diversified architectures, including newly emerging ones.
This paper systematically compares the developmental progress of various architectures (MLPs, CNNs, RNNs, GNNs, Transformer, Diffusion, foundation models, Mamba) and analyzes the strengths, weaknesses, and contributions of each.
In addition, it addresses the performance of hybrid models that combine the strengths of multiple architectures, clearly highlighting key trends in TSF.
With these contributions, readers can effectively understand the continuously evolving trends and directions of advancement in the field.
Through this, it lowers the entry barrier for newcomers to TSF and provides a comprehensive roadmap that opens up new research opportunities for established researchers.

\begin{figure}[!h]
\centering
\includegraphics[width=1\textwidth, trim=0cm 0cm 0cm 0cm, clip]{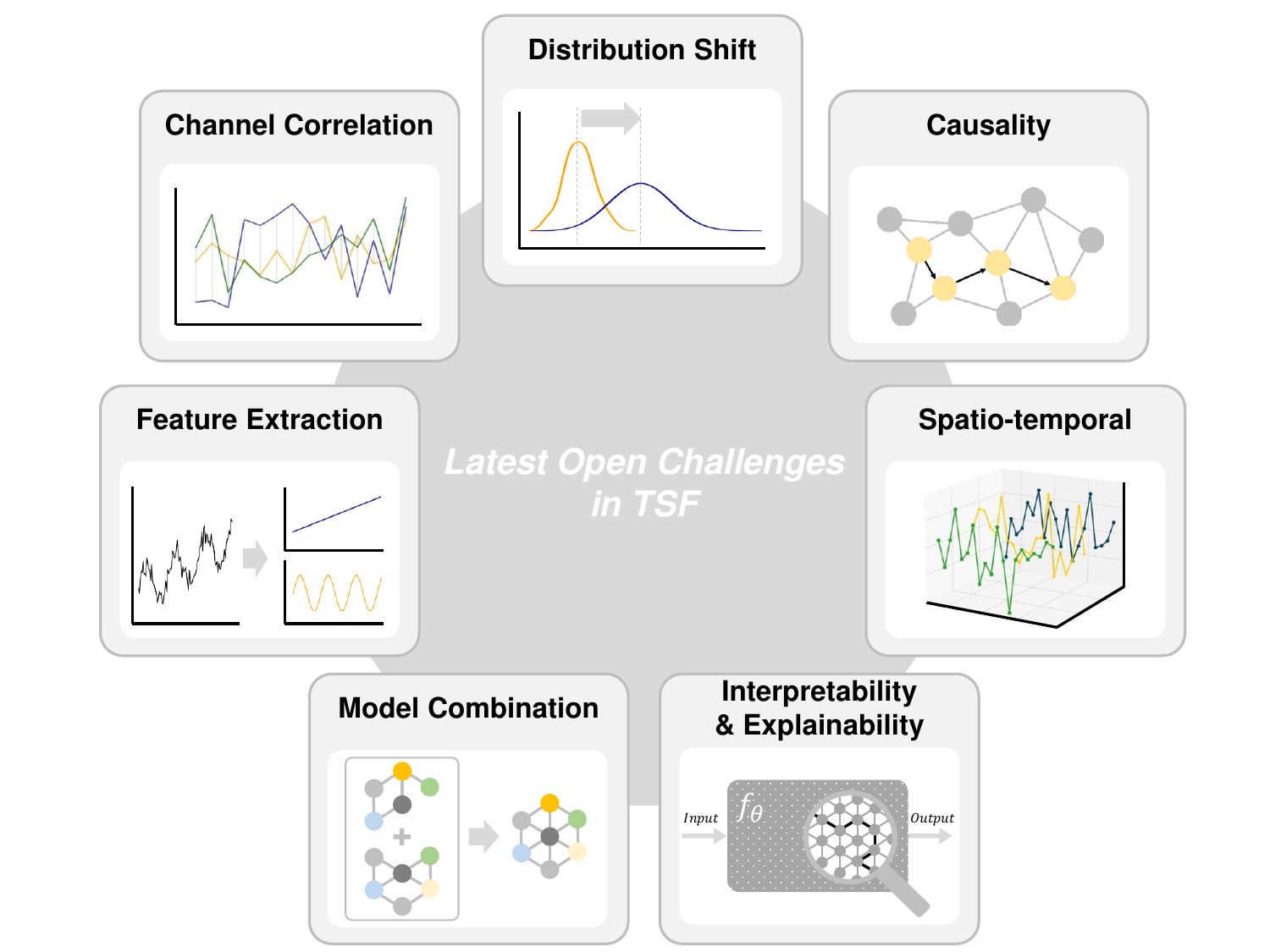}
\begin{picture}(0,0)(0,0)
    \put(-115,287){\makebox(0,0)[c]{\textbf{\fontsize{6}{10}\selectfont(Section \ref{sec5-1})}}}
    \put(3,313){\makebox(0,0)[c]{\textbf{\fontsize{6}{10}\selectfont(Section \ref{sec5-2})}}}
    \put(122,287){\makebox(0,0)[c]{\textbf{\fontsize{6}{10}\selectfont(Section \ref{sec5-3})}}}
    \put(-129,186){\makebox(0,0)[c]{\textbf{\fontsize{6}{10}\selectfont(Section \ref{sec5-4})}}}
    \put(-56,86){\makebox(0,0)[c]{\textbf{\fontsize{6}{10}\selectfont(Section \ref{sec5-5})}}}
    \put(66,80){\makebox(0,0)[c]{\textbf{\fontsize{6}{10}\selectfont(Section \ref{sec5-6})}}}
    \put(133,186){\makebox(0,0)[c]{\textbf{\fontsize{6}{10}\selectfont(Section \ref{sec5-7})}}}
\end{picture}

\caption{Overview of Latest Open Challenges in Time Series Forecasting}\label{Overview of Latest Issues in Time Series Forecasting}
\end{figure}

Second, we explore from the perspective of open challenges.
Although numerous advanced architectures have resolved many issues, the core challenges in TSF continue to persist.
In particular, issues such as channel correlation, distribution shifts, causality, and feature extraction (Section \ref{sec5}) remain significant challenges that need to be addressed (Fig. \ref{Overview of Latest Issues in Time Series Forecasting}).
This survey explores the latest methodologies aimed at addressing these challenges and provides readers with valuable insights for problem-solving.
\revision{While some existing surveys address these TSF challenges, they often focus on only a few problems.
In contrast, this paper takes an issue-driven approach, deeply examining seven key TSF problems and organizing new solutions.}

\begin{sidewaystable} 
\centering
\captionsetup{justification=raggedright, singlelinecheck=false} 
\caption{Summary of Survey Papers on Time Series Forecasting} 
\label{tab:SurveyList}
\renewcommand{\arraystretch}{1.2}

\arrayrulecolor[gray]{0.8} 
\begin{tabular}{!{\color{black}\vrule width 1pt} >{\centering\arraybackslash}m{4cm}|>{\raggedright\arraybackslash}m{10cm}|>{\centering\arraybackslash}m{2cm}|>{\centering\arraybackslash}m{2cm}|>{\centering\arraybackslash}m{1.3cm}|>{\centering\arraybackslash}m{2.5cm}!{\color{black}\vrule width 1pt}}

\arrayrulecolor{black}\specialrule{1pt}{0pt}{0pt} 

\rowcolor[HTML]{808080} 
{\color[HTML]{FFFFFF} \textbf{Articles}} & 
\multicolumn{1}{>{\centering\arraybackslash}m{10cm}}{\color[HTML]{FFFFFF} \textbf{Focus}} & 
{\color[HTML]{FFFFFF} \textbf{Broad Architecture Review}} & 
{\color[HTML]{FFFFFF} \textbf{\revision{TSF Challenges}}} & 
{\color[HTML]{FFFFFF} \textbf{Recent Work}} & 
{\color[HTML]{FFFFFF} \textbf{Reference}} \\
\arrayrulecolor[gray]{0.8}\hline

Time-series forecasting with deep learning: a survey & · Encoder-Decoder structures and Hybrid models \newline · Interpretability and causal inference for decision support & \checkmark & \revision{\checkmark} &  & \cite{TSFwithDL} \\ \hline
Forecast Methods for Time Series Data: A Survey & · \revision{Categorization of various} forecasting methods: statistical, ML, and DL \newline · Challenges in preprocessing, modeling, and parallel computation & \checkmark &  &  & \cite{ForecastMethodsforTS} \\ \hline
Deep Learning for Time Series Forecasting: Tutorial and Literature Survey & · Key components of DL TSF \newline · Practical applicability & \checkmark &  &  & \cite{DLforTSF} \\ \hline
A Review on Deep Sequential Models for Forecasting Time Series Data & · Common deep sequential models \newline · Guidelines on implementation, application, and optimization & \checkmark &  &  & \cite{ReviewOnDeepSequential} \\ \hline
Transformers in Time Series: A Survey & · Transformer structure variations and architectural improvements &  & \revision{\checkmark} &  & \cite{TrasformersInTS} \\ \hline
Long Sequence Time-Series Forecasting with Deep Learning: A Survey & · Definition of long sequence forecasting challenges from various perspectives \newline · New classification system and performance evaluation methods & \checkmark & \revision{\checkmark} &  & \cite{LSTFwithDeep} \\ \hline
Machine Learning Advances for Time Series Forecasting & · Economic/financial TSF \newline · Categorized into linear and nonlinear & \checkmark &  &  & \cite{MLforTS} \\ \hline
Diffusion Models for Time-Series Applications: A Survey & · Diffusion models in time series analysis &  &  &  & \cite{DiffForTS} \\ \hline
The Rise of Diffusion Models in Time-Series Forecasting & · Diffusion-based forecasting models with case studies &  &  & \checkmark & \cite{RiseofDiff} \\ \hline
Foundation Models for Time Series Analysis: A Tutorial and Survey & · Application of foundation models in time series analysis &  &  & \checkmark & \cite{FMforTS} \\ \hline
Large Language Models for Time Series: A Survey & · Categorization of methodologies for LLMs in time series \newline · Explanation of bridging modality gaps &  &  & \checkmark & \cite{LLMforTSsurvey} \\ \hline
A Survey of Time Series Foundation Models & · Approaches to Foundation Models and LLMs in time series analysis &  &  & \checkmark & \cite{surveyOfTSFM} \\ \hline
Mamba-360: Survey of State Space Models & · Emphasis on SSM advantages in long time series \newline · Comparison of domain-specific applications and performance &  &  & \checkmark & \cite{mamba360} \\ \hline
\textbf{This Survey} & · A comprehensive overview of the evolution of TSF models, introducing innovative architectures \newline · Discussion of critical open challenges in forecasting tasks, along with advanced solutions to address them & \checkmark & \revision{\checkmark} & \checkmark &  \\
\arrayrulecolor{black}\specialrule{1.0pt}{0pt}{0pt}

\end{tabular}
\end{sidewaystable}

\normalsize


This survey covers the fundamental concepts of time series data and the problem definition of forecasting in Section \ref{sec2}, followed by an examination of the evolution of past methodologies in Section \ref{sec3}.
Section \ref{sec4} analyzes the key features of the latest models, and finally, Section \ref{sec5}, explores the open challenges in TSF and their solutions.
Through this, readers will gain a broad understanding of the past and present of TSF research and acquire new ideas for future studies.

\section{Background}\label{sec2}

In this section, before exploring time series forecasting models, we explain the definition and key characteristics of time series data to provide the necessary background. We also define the problem of time series forecasting tasks, discussing related datasets and evaluation metrics. This establishes the basic concepts of time series forecasting and provides preliminary information to help understand the models discussed in the following sections.

\subsection{Time-Series Data}\label{sec2-1}
Time series data is a collection of sequential data points gathered at regular time intervals, representing a series of observations of phenomena that change over time. This temporal continuity allows for the understanding and analysis of phenomena that evolve according to time order. Each data point represents the state or value at a specific moment, and through the observation of these data points, various patterns such as long-term trends, seasonality, cyclicality, and irregularities can be recognized. These patterns provide valuable information for predicting future values or detecting changes at critical moments.
By extracting and studying the meaningful information provided by time series data, \revision{researchers can develop practical applications to address challenges} in various disciplines.

\begin{figure}[!h]
\centering
\includegraphics[width=1.0\textwidth, trim=1cm 3cm 1cm 3cm, clip]{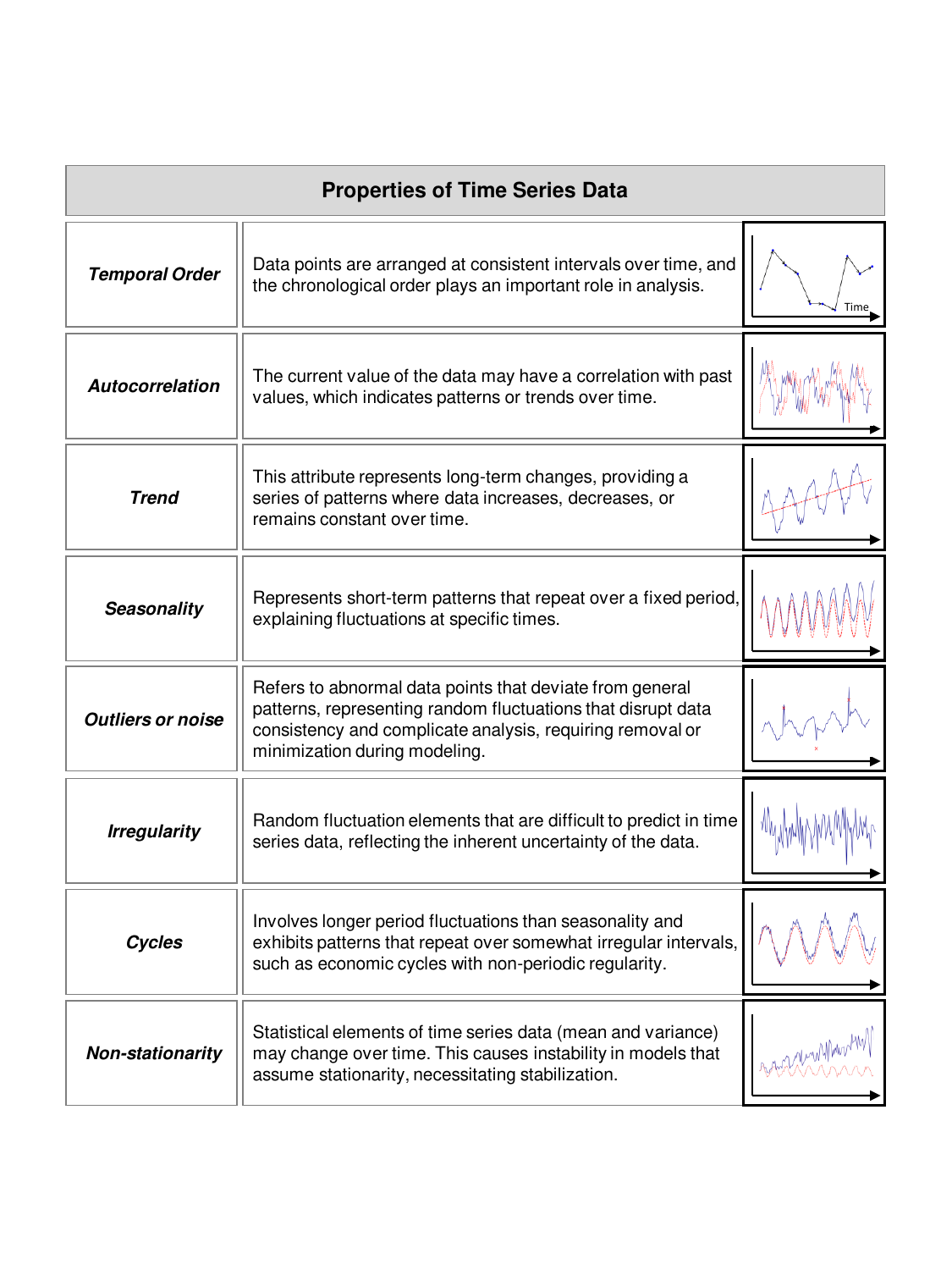}
\caption{Properties of Time Series Data}
\label{Properties of Time Series Data}
\end{figure}

\paragraph{Characteristics of Time Series Data}
Time series data encapsulates various characteristics that play a critical role in explaining the diverse patterns and fluctuations within time series.
Understanding these characteristic elements is essential for analyzing and predicting data. The key properties are explained in Fig. \ref{Properties of Time Series Data}.

\begin{figure}[!h]
\centering
\includegraphics[width=1.0\textwidth, trim=1cm 8.5cm 1cm 8.5cm, clip]{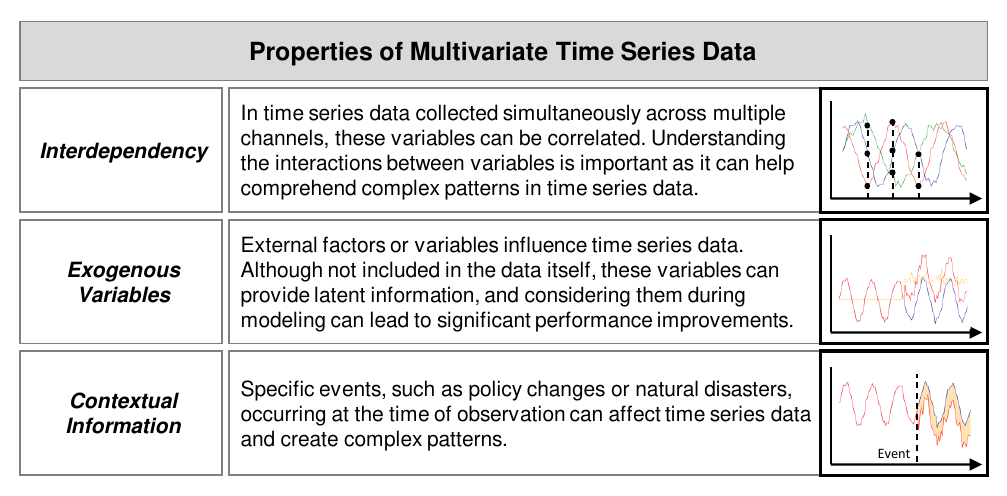}
\caption{Properties of Multivariate Time Series Data}
\label{Properties of Multivariate Time Series Data}
\end{figure}

In time series data, the above features frequently appear in a mixed form.
Therefore, decomposition is commonly used to separate the components for detailed analysis, or distribution shift alleviation methods are widely applied.
Many time series datasets provide information on multivariate variables.
Sometimes, these data provide additional information that univariate data cannot, and it is important to understand this for many problems.
The main properties are explained in Fig. \ref{Properties of Multivariate Time Series Data}.

These additional features convey the complexity of the real world in greater detail and sometimes provide reasons for significant pattern changes.
Considering this information can lead to more precise time series analysis and forecasting.

\subsection{Time-Series Forecasting}\label{sec2-2}
Time series forecasting involves analyzing historical time-series data to predict future values. This process entails analyzing inherent temporal patterns in the data, such as trends and seasonality, using past observations to estimate future values. The goal of time series forecasting is to support decision-making by providing information about an uncertain future, such as predicting future demand changes or fluctuations in prices. Time series forecasting can employ various methodologies, including statistical methods, machine learning, and deep learning. These methodologies focus on capturing the characteristics of data as it changes over time.

\subparagraph{\textbf{Univariate Time Series Forecasting (UTSF)}}
Univariate forecasting refers to making predictions using only one variable. For example, predicting the next day's temperature based solely on past temperature data from a weather station is a univariate forecast. The advantage of univariate forecasting is that the models are simple and computationally efficient. Since the data consists of a single variable, the model is relatively straightforward and easy to understand, making data collection and management easier. However, it may only utilize limited information as it cannot account for important external factors or interactions between different variables.
\revision{\begin{align}
    \hat{x}_{t+1:t+h} = f({x}_{t-p:t}) \label{eq:utsf}
\end{align}}
\revision{The formula representing UTSF is shown in Eq. (\ref{eq:utsf}). The model takes the past \( p+1 \) time steps, specifically the data from \( t-p \) to \( t \), as input to make predictions. Here, \( \hat{x}_{t+1:t+h} \) represents the predicted values from \( t+1 \) to \( t+h \). Single-step forecasting predicts only one value when \( h=1 \), while multi-step forecasting predicts multiple values when \( h>1 \).}

\subparagraph{\textbf{Multivariate Time Series Forecasting (MTSF)}}
Multivariate forecasting involves making predictions using multiple variables simultaneously. For instance, in weather forecasting, predicting the next day's temperature by considering various variables such as temperature, humidity, and wind speed is an example of multivariate forecasting. By incorporating interactions and correlations between multiple variables, multivariate forecasting can capture complex relationships, offering higher predictive accuracy. However, these models tend to be more complex, require more data, and can be more challenging to handle, increasing the risk of overfitting.

\begin{figure}[H]
\centering
\includegraphics[width=0.8\textwidth, trim=0cm 2.5cm 0cm 4.5cm, clip]{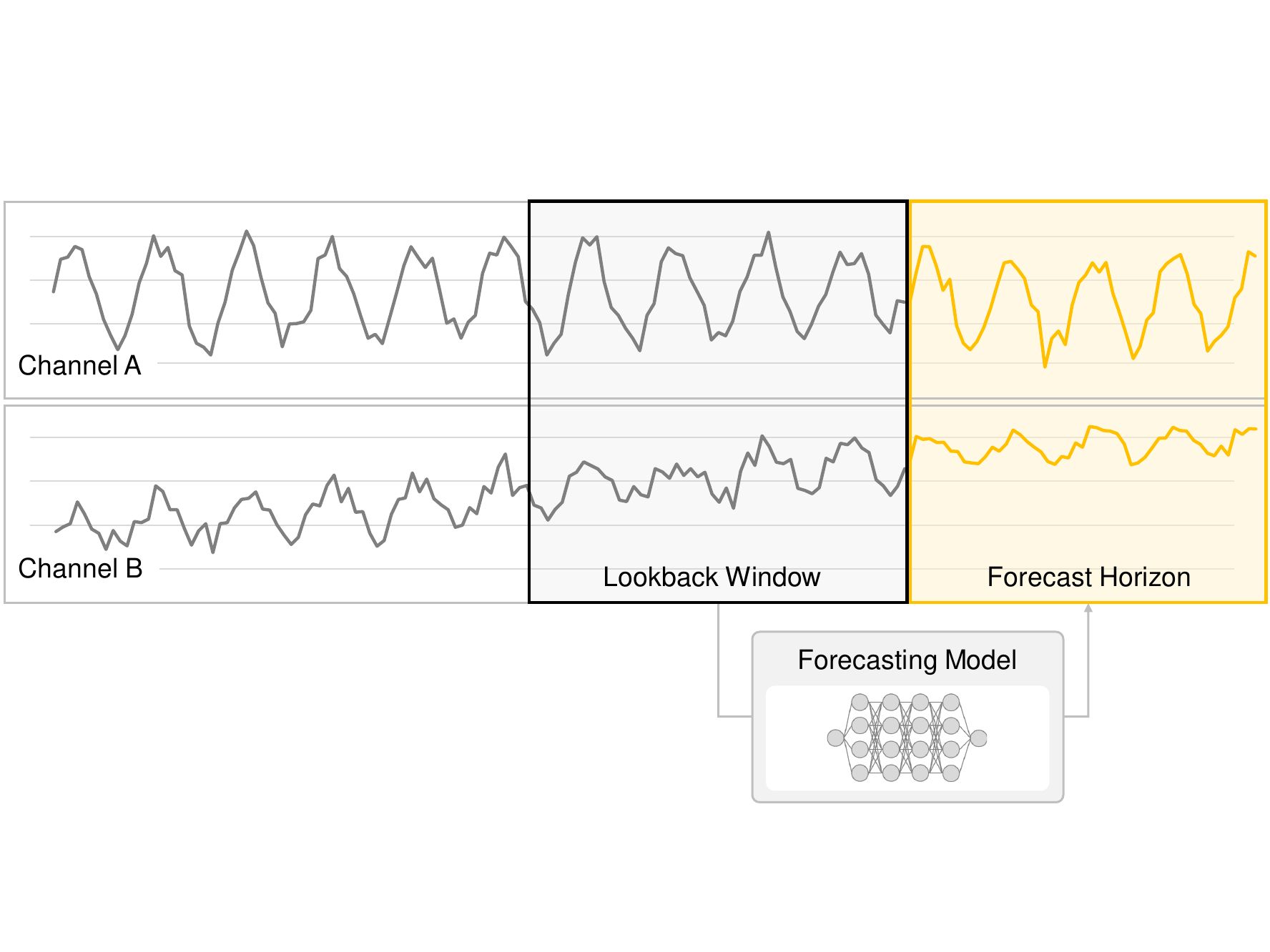}
\caption{Diagram of Multivariate Time Series Forecasting}\label{Diagram of Multivariate Time Series Forecasting}
\end{figure}

To incorporate other influencing variables, we can extend the model to a multivariate approach.
\revision{\begin{align}
    \hat{\mathbf{X}}_{t+1:t+h} = f(\mathbf{X}_{t-p:t}) \label{eq:mtsf}
\end{align}}
\revision{The formula representing MTSF is shown in Eq. (\ref{eq:mtsf}).
\( \mathbf{X}_{t-p:t} \) is a set of vectors over the past \( p+1 \) time steps, where each vector contains the values of multiple variables at time \( t \). This model takes multiple variables from the past \( p+1 \) time steps, specifically \( \mathbf{X}_{t-p:t} \), as input to generate the multivariate predicted values from \( t+1 \) to \( t+h \), denoted as \( \hat{\mathbf{X}}_{t+1:t+h} \).}
This multivariate approach allows for more comprehensive modeling by considering both the target variable's history and the effects of other relevant factors. Fig. \ref{Diagram of Multivariate Time Series Forecasting} illustrates how a multivariate forecasting model uses a past lookback window to predict future intervals.

\subparagraph{\textbf{Short-Term Time Series Forecasting (STSF)}}
Short-term time series forecasting focuses on predictions for the near future, making it suitable for tasks that require quick responses, such as immediate operational planning or short-term decision-making. The models are simple, making them easy to train and implement, and they often demonstrate relatively high accuracy. However, because the forecast range is short, it cannot capture long-term trends or complex variations, which limits its applicability.

\subparagraph{\textbf{Long-Term Time Series Forecasting (LTSF)}}
Long-term time series forecasting deals with predictions for the distant future, with forecast horizons increasingly extending to several months, years, or beyond. It is valuable for long-term strategy planning, investment decisions, and policy-making, addressing many real-world problems. By identifying long-term trends and cycles, organizations can prepare accordingly, highlighting its significance. However, predicting the distant future is challenging, and extensive research is being conducted to improve accuracy.

\subsection{Time Series Forecasting Datasets}\label{sec2-3}
Models for Time Series Forecasting (TSF) are primarily trained and evaluated using open benchmark datasets. Open datasets offer data that has already been collected across various fields, allowing researchers to use them without going through separate data collection processes. Additionally, researchers can objectively compare the performance of different models and convincingly demonstrate the contributions of their developed models.

Datasets frequently used by time series forecasting models in various domains, such as energy consumption, traffic, weather, exchange rates, and disease outbreaks, are listed in Table \ref{tab:datasets}. These time series datasets are typically collected in real-time via sensors, recorded as transaction logs in financial markets, accumulated as server logs, or gathered through periodic surveys.

\begin{table}[!h]
\centering

\captionsetup{justification=raggedright,singlelinecheck=false}
\caption{Datasets Frequently Used in Time Series Forecasting Models}
\label{tab:datasets}

\footnotesize 
\renewcommand{\arraystretch}{1.3} 
\arrayrulecolor[gray]{0.8} 

\begin{tabular}{!{\color{black}\vrule width 1pt} >{\centering\arraybackslash}m{1.3cm}|>{\centering\arraybackslash}m{1.3cm}|>{\centering\arraybackslash}m{1cm}|m{0.35\textwidth}|>{\centering\arraybackslash}m{1.5cm}|>{\centering\arraybackslash}m{0.12\textwidth}!{\color{black}\vrule width 1pt}}

\arrayrulecolor{black}\specialrule{1.0pt}{0pt}{0pt} 

\rowcolor[HTML]{808080}
{\color[HTML]{FFFFFF} \textbf{Dataset}} & 
{\color[HTML]{FFFFFF} \textbf{Channel}} & 
{\color[HTML]{FFFFFF} \textbf{Length}} & 
\multicolumn{1}{c}{\cellcolor[HTML]{808080}\centering {\color[HTML]{FFFFFF} \textbf{Description}}} &  
{\color[HTML]{FFFFFF} \textbf{Frequency}} & 
\multicolumn{1}{c|}{\cellcolor[HTML]{808080}{\color[HTML]{FFFFFF} \textbf{Source}}} \\
\arrayrulecolor[gray]{0.8} 

\makecell{ETTm1, \\ ETTm2} & 7 & 69680 & Electricity Transformer Temperature data from two counties in China for 2 years & 15 mins & \href{https://github.com/zhouhaoyi/ETDataset}{ETDataset} \\ \hline

\makecell{ETTh1, \\ ETTh2} & 7 & 17420 & Electricity Transformer Temperature data from two counties in China for 2 years & Hourly & \href{https://github.com/zhouhaoyi/ETDataset}{ETDataset} \\ \hline

Electricity & 321 & 26304 & Hourly electricity consumption data of clients from 2012 to 2014 & Hourly & \href{https://archive.ics.uci.edu/dataset/321/electricityloaddiagrams20112014}{UCI Archive} \\ \hline

Traffic & 862 & 17544 & Hourly road occupancy rates on San Francisco Bay area freeways for 48 months (2015-2016) & Hourly, Weekly & \href{https://pems.dot.ca.gov/}{PeMS} \\ \hline

Weather & 21 & 52696 & Meteorological data (e.g., humidity, temperature) in Germany & 10 mins & \href{https://www.bgc-jena.mpg.de/wetter/}{BGC Jena} \\ \hline

Exchange & 8 & 7588 & Daily exchange rates of eight countries from 1990 to 2016 & Daily & \href{https://github.com/laiguokun/multivariate-time-series-data}{Multivariate Time Series Data} \\ \hline

ILI & 7 & 966 & Weekly Influenza-Like Illness data from the U.S. CDC & Weekly & \href{https://gis.cdc.gov/grasp/fluview/fluportaldashboard.html}{CDC FluView} \\
\arrayrulecolor{black}\specialrule{1.0pt}{0pt}{0pt} 

\end{tabular}%

\end{table}

Time series data exist across a wide range of domains, each exhibiting unique characteristics.
This diversity is due to the nature of time series data, which addresses real-world problems in various fields and can display highly varied patterns depending on time and environmental factors.
For example, stock price data often show long-term upward or downward trends.
Stock market fluctuations tend to move in specific directions over the long term due to complex influences such as economic conditions, corporate performance, and policy changes.
In contrast, electricity consumption data exhibit pronounced periodicity throughout the day, with higher consumption during daytime hours and lower consumption at night.
There are also seasonal patterns, with different consumption rates during summer and winter.

Monash Time Series Forecasting Archive \citep{godahewa2021monash} points out that there is no comprehensive time series dataset archive and thus provides a thorough time series repository.
This repository offers 30 datasets for real-world data and competition purposes, covering various domains such as energy, economics, nature, and sales.
Details are specified in Table \ref{tab:dataset_summary}.

\begin{sidewaystable}
\centering
\captionsetup{justification=raggedright, singlelinecheck=false} 
\caption{Summary of Datasets Across Domains from Monash Archive \citep{godahewa2021monash}}
\label{tab:dataset_summary}
\footnotesize 
\renewcommand{\arraystretch}{1.6} 

\begin{tabular}{!{\color{black}\vrule width 1pt} >{\centering\arraybackslash}m{3cm}|>{\centering\arraybackslash}m{1.7cm}|>{\centering\arraybackslash}m{1.7cm}|>{\centering\arraybackslash}m{1.7cm}|>{\centering\arraybackslash}m{1.7cm}|>{\centering\arraybackslash}m{1.7cm}|>{\centering\arraybackslash}m{1.7cm}|>{\centering\arraybackslash}m{5cm}!{\color{black}\vrule width 1pt}}

\arrayrulecolor{black}\specialrule{1.0pt}{0pt}{0pt} 

\rowcolor[HTML]{808080} 
{\color[HTML]{FFFFFF} \textbf{Dataset}} & 
{\color[HTML]{FFFFFF} \textbf{Domain}} & 
{\color[HTML]{FFFFFF} \textbf{No. Series}} & 
{\color[HTML]{FFFFFF} \textbf{Min. Length}} & 
{\color[HTML]{FFFFFF} \textbf{Max. Length}} & 
{\color[HTML]{FFFFFF} \textbf{Competition}} & 
{\color[HTML]{FFFFFF} \textbf{Multivariate}} & 
{\color[HTML]{FFFFFF} \textbf{Source}} \\ 

\arrayrulecolor[gray]{0.8}\hline

M1 & Multiple & 1001 & 15 & 150 & Yes & No & \cite{makridakis1982accuracy} \\ \hline
M3 & Multiple & 3003 & 20 & 144 & Yes & No & \cite{makridakis2000m3} \\ \hline
M4 & Multiple & 100000 & 19 & 9933 & Yes & No & \cite{makridakis2020m4} \\ \hline
Tourism & Tourism & 1311 & 11 & 333 & Yes & No & \cite{athanasopoulos2011tourism} \\ \hline
CIF 2016 & Banking & 72 & 34 & 120 & Yes & No & \cite{vstvepnivcka2017results} \\ \hline
London Smart Meters & Energy & 5560 & 288 & 39648 & No & No & \href{https://www.kaggle.com/datasets/jeanmidev/smart-meters-in-london}{Jean-Michel, 2019} \\ \hline
Aus. Electricity Demand & Energy & 5 & 230736 & 232272 & No & No & \cite{godahewa_2021_4659727} \\ \hline
Wind Farms & Energy & 339 & 6345 & 527040 & No & No & \cite{godahewa_2021_4654909,godahewa_2021_4654858} \\ \hline
Dominick & Sales & 115704 & 28 & 393 & No & No & \cite{center2020dominick} \\ \hline
Bitcoin & Economic & 18 & 2659 & 4581 & No & No & \cite{godahewa_2021_5121965,godahewa_2021_5122101} \\ \hline
Pedestrian Counts & Transport & 66 & 576 & 96424 & No & No & \href{https://data.melbourne.vic.gov.au/explore/dataset/all-sensors-real-time-status/information/?disjunctive.dev_id}{City of Melbourne, 2020} \\ \hline
Vehicle Trips & Transport & 329 & 70 & 243 & No & No & \href{https://github.com/fivethirtyeight/uber-tlc-foil-response}{fivethirtyeight, 2015} \\ \hline
KDD Cup 2018 & Nature & 270 & 9504 & 10920 & Yes & No & \href{https://www.kdd.org/kdd2018/kdd-cup}{KDD Cup, 2018} \\ \hline
Weather & Nature & 3010 & 1332 & 65981 & No & No & \cite{sparks2020bomrang} \\ \hline
NN5 & Banking & 111 & 791 & 791 & Yes & Yes & \cite{taieb2012review} \\ \hline
Web Traffic & Web & 145063 & 803 & 803 & Yes & Yes & \href{https://www.kaggle.com/c/web-traffic-time-series-forecasting}{Google, 2017} \\ \hline
Solar & Energy & 137 & 52560 & 52560 & No & Yes & \href{https://www.nrel.gov/grid/solar-power-data.html}{Solar, 2020} \\ \hline
Electricity & Energy & 321 & 26304 & 26304 & No & Yes & \href{https://archive.ics.uci.edu/dataset/321/electricityloaddiagrams20112014}{UCI, 2020} \\ \hline
Car Parts & Sales & 2674 & 51 & 51 & No & Yes & \href{https://cran.r-project.org/web/packages/expsmooth/}{Hyndman, 2015} \\ \hline
FRED-MD & Economic & 107 & 728 & 728 & No & Yes & \cite{mccracken2016fred} \\ \hline
San Francisco Traffic & Transport & 862 & 17544 & 17544 & No & Yes & \href{https://pems.dot.ca.gov/}{Caltrans, 2020} \\ \hline
Rideshare & Transport & 2304 & 541 & 541 & No & Yes & \cite{godahewa_2021_5122114,godahewa_2021_5122232} \\ \hline
Hospital & Health & 767 & 84 & 84 & No & Yes & \href{https://cran.r-project.org/web/packages/expsmooth/}{Hyndman, 2015} \\ \hline
COVID Deaths & Nature & 266 & 212 & 212 & No & Yes & \href{https://github.com/CSSEGISandData/COVID-19}{Johns Hopkins University, 2020} \\ \hline
Temperature Rain & Nature & 32072 & 725 & 725 & No & Yes & \cite{godahewa_2021_5129073,godahewa_2021_5129091} \\ \hline
Sunspot & Nature & 1 & 73931 & 73931 & No & No & \href{https://www.sidc.be/SILSO/newdataset}{Sunspot, 2015} \\ \hline
Saugeen River Flow & Nature & 1 & 23741 & 23741 & No & No & \href{http://www.jenvstat.org/v04/i11}{McLeod and Gweon, 2013} \\ \hline
US Births & Nature & 1 & 7305 & 7305 & No & No & \cite{mcleod2013optimal} \\ \hline
Solar Power & Energy & 1 & 7397222 & 7397222 & No & No & \cite{godahewa_2021_4656027} \\ \hline
Wind Power & Energy & 1 & 7397147 & 7397147 & No & No & \cite{godahewa_2021_4656032} \\
\arrayrulecolor{black}\specialrule{1.0pt}{0pt}{0pt} 
\end{tabular}
\end{sidewaystable}

\normalsize
Time series data exhibit a greater variety of domain-specific characteristics compared to Natural Language Processing (NLP). These differences in dataset characteristics make time series forecasting problems more diverse and complex, reducing the generalizability of time series forecasting models. Consequently, developing foundation models for TSF is more challenging compared to NLP. Time series data lack the well-designed vocabulary and grammar found in NLP, and obtaining vast amounts of data is more difficult than in fields like Computer Vision (CV) or NLP. In the CV field, there are general benchmark datasets such as MNIST and ImageNet, while in NLP, there are datasets like GLUE (General Language Understanding Evaluation) and SQuAD (Stanford Question Answering Dataset). In contrast, the time series analysis field lacks large-scale general benchmark datasets and large-scale challenges like those in image and language domains. The most famous time series forecasting competition is the M-Competitions, which started in 1982 and has only been held six times to date. The paper ``Time Series Dataset Survey for Forecasting with Deep Learning'' \citep{datasetsurvey} highlights the lack of general benchmark datasets for TSF and analyzes the statistical characteristics of datasets across various domains, clustering them to provide easily accessible information for researchers. \revision{We do not delve into the characteristics of the datasets in detail, but instead provide a brief overview of the key datasets.} Recently, the development of foundation models for time series has begun, leading to the gradual emergence of general datasets for TSF. Details about TSF foundation models will be discussed in Section 4.3.

\subsection{Evaluation Metrics}\label{sec2-4}
The evaluation metrics used in TSF play a crucial role in objectively comparing and assessing model performance. Common metrics for deterministic models include Mean Absolute Error (MAE) and Mean Squared Error (MSE), which are easy to use because they allow intuitive comparisons of prediction errors. On the other hand, probabilistic models that can reflect uncertainty typically use Continuous Ranked Probability Score (CRPS), which measures the accuracy of distribution forecasts by evaluating the difference between predicted and actual distributions \citep{matheson1976scoring}. However, the limitations of these traditional evaluation metrics have been noted, leading to the proposal of various performance metrics \citep{thompson1990mse}.

The distinct characteristics of time series data necessitate the use of diverse metrics in TSF. It is essential to assess not only the magnitude of errors but also how well the model learns the temporal patterns and various features of time series data. Table \ref{tab:Evaluation Metrics for Time Series Forecasting} categorizes evaluation metrics by type and explains each metric. 


\revision{Error-based metrics for deterministic models measure the difference between predicted and actual values.
Explained variance metrics assess how well the model explains the variance of the data \citep{chicco2021coefficient}.
These metrics evaluate how well the model captures key patterns such as seasonality or trends in the data, thereby aiding in the structural understanding of the predictions.
Error metrics focus on accuracy, but they do not assess overfitting, whereas explained variance evaluates the model's generalization ability.
Model selection metrics help identify the optimal model \citep{portet2020primer}.
These metrics take into account factors such as sample size and the number of parameters, playing a key role in selecting the optimal model.
Thus, they are particularly useful for choosing the most appropriate model for a given dataset.
For instance, in noisy environments, MAE is preferred over MSE due to its reduced sensitivity to outliers.
Additionally, when evaluating non-stationary data, it is important to use metrics that effectively capture changes in trends and seasonality.
In such cases, cumulative error-based metrics or explained variance metrics can be useful.}

\revision{Probabilistic model evaluation metrics, on the other hand, account for uncertainty, enabling the assessment of both the confidence and variability of predictions.
These metrics help evaluate not only the predicted values but also the uncertainty surrounding them.
Beyond basic error-based metrics, several advanced metrics are used.
Interval Metrics assess the confidence intervals of the predictions, considering both the predicted values and their associated uncertainty \citep{khosravi2011prediction}.
These metrics are especially useful in fields such as financial market forecasting and risk management.
Quantile Metrics evaluate the accuracy of specific quantiles of predicted values, making them particularly useful when assessing how well various quantiles match the actual values in situations with high uncertainty \citep{koenker1978regression}.
Quantile Metrics are beneficial in areas such as weather forecasting and insurance, where forecasts are assessed across different risk levels.
Sharpness Metrics, which evaluate the width of predicted confidence intervals, are used to assess how narrow and accurate the prediction intervals are \citep{zhao2020individual}.
This metric is valuable in logistics and supply chain forecasting, where minimizing the width of prediction intervals is crucial.}

\revision{Thus, various evaluation metrics are crucial in assessing model performance, and each should be chosen based on the specific characteristics of the data and the requirements of the problem.
In real-world scenarios, the appropriate use of metrics allows for the selection of the most effective model, which can then be leveraged for efficient decision-making.
In contrast, in the field of deep learning research, MSE and MAE remain widely used due to their simplicity, intuitiveness, and the fact that they have been extensively utilized in previous studies, making them suitable for performance comparison \citep{LSTFwithDeep}.}

\begin{sidewaystable}[]
\centering
\captionsetup{justification=raggedright, singlelinecheck=false} 
\caption{Evaluation Metrics for Time Series Forecasting}

\label{tab:Evaluation Metrics for Time Series Forecasting}

\tiny 
\renewcommand{\arraystretch}{2.2} 

\begin{tabular}{!{\color{black}\vrule width 1pt} >{\centering\arraybackslash}m{1.5cm}|>{\centering\arraybackslash}m{3cm}|>{\centering\arraybackslash}m{4cm}|>{\centering\arraybackslash}m{5cm}|m{6cm}!{\color{black}\vrule width 1pt}}

\arrayrulecolor{black}\specialrule{1.0pt}{0pt}{0pt} 

\rowcolor[HTML]{808080} 
{\color[HTML]{FFFFFF} \textbf{Evaluation Type}} & 
{\color[HTML]{FFFFFF} \textbf{Category}} & 
{\color[HTML]{FFFFFF} \textbf{Metric}} & 
{\color[HTML]{FFFFFF} \textbf{Formula}} & 
\multicolumn{1}{c|}{\cellcolor[HTML]{808080}{\color[HTML]{FFFFFF} \textbf{Explanation}}} \\

\arrayrulecolor[gray]{0.8}\hline

\multirow{19}{*}{Deterministic} & \multirow{3}{*}{\revision{Error-Based: Absolute}} & Mean Absolute Error (MAE) & $\frac{1}{n} \sum_{t=1}^{n} |y_t - \hat{y}_t|$ & Measures the average magnitude of the errors in a set of predictions, without considering their direction.\\ \hhline{|~|~|-|-|-|}
& & Mean Squared Error (MSE) & $\frac{1}{n} \sum_{t=1}^{n} (y_t - \hat{y}_t)^2$ & Measures the average of the squared errors. \\ \hhline{|~|~|-|-|-|}
& & Root Mean Squared Error (RMSE) & $\sqrt{\frac{1}{n} \sum_{t=1}^{n} (y_t - \hat{y}_t)^2}$ & Square root of the average of squared errors, giving higher weight to larger errors. \\ \hhline{|~|-|-|-|-|}
& \multirow{2}{*}{\revision{Error-Based: Relative}} & Mean Absolute Percentage Error (MAPE) & $\frac{1}{n} \sum_{t=1}^{n} \left|\frac{y_t - \hat{y}_t}{y_t}\right| \times 100$ & Measures the size of the error in percentage terms. \\ \hhline{|~|~|-|-|-|}
& & Symmetric Mean Absolute Percentage Error (sMAPE) & $\frac{1}{n} \sum_{t=1}^{n} \frac{|y_t - \hat{y}_t|}{(|y_t| + |\hat{y}_t|)/2} \times 100$ & Measures the accuracy based on relative error. \\ \hhline{|~|-|-|-|-|}
& \multirow{2}{*}{\revision{Error-Based: Cumulative}} & Mean Forecast Error (MFE) & $\frac{1}{n} \sum_{t=1}^{n} (y_t - \hat{y}_t)$ & Average of forecast errors, indicating bias. \\ \hhline{|~|~|-|-|-|}
& & Cumulative Forecast Error (CFE) & $\sum_{t=1}^{n} (y_t - \hat{y}_t)$ & Sum of all forecast errors, measures total bias over the forecast horizon. \\ \hhline{|~|-|-|-|-|}

& \revision{Error-Based: Scaled} & Mean Absolute Scaled Error (MASE) & $\frac{\frac{1}{n} \sum_{t=1}^{n} |y_t - \hat{y}_t|}{\frac{1}{n-1} \sum_{t=2}^{n} |y_t - y_{t-1}|}$ & MAE scaled by the MAE of a naive forecast. \\ \hhline{|~|-|-|-|-|}

& \multirow{4}{*}{\centering Explained Variance Metrics} & Coefficient of Determination (R²) & $1 - \frac{\sum_{t=1}^{n} (y_t - \hat{y}_t)^2}{\sum_{t=1}^{n} (y_t - \bar{y})^2}$ & Proportion of variance explained by the model. \\ \hhline{|~|~|-|-|-|}
& & Adjusted Coefficient of Determination (Adjusted R²) & $1 - \frac{(1-R^2)(n-1)}{n-k-1}$ & R² adjusted for the number of predictors. \\ \hhline{|~|~|-|-|-|}
& & Explained Variance Score (EVS) & $1 - \frac{\text{Var}(y_t - \hat{y}_t)}{\text{Var}(y_t)}$ & Measures the proportion of variance explained by the model. \\ \hhline{|~|-|-|-|-|}

& \multirow{5}{*}{Model Selection Metrics} & Akaike Information Criterion (AIC) & $2k - 2\ln(\hat{L})$ & Trade-off between goodness of fit and model complexity. \\ \hhline{|~|~|-|-|-|}
& & Bayesian Information Criterion (BIC) & $k\ln(n) - 2\ln(\hat{L})$ & Similar to AIC with a stronger penalty for models with more parameters. \\ \hhline{|~|~|-|-|-|}
& & Hannan-Quinn Criterion (HQC) & $ \text{AICc} = 2k - 2\ln(\hat{L}) + \frac{2k(k+1)}{n-k-1} $ & Alternative to AIC and BIC with different penalty terms. \\ \hhline{|~|~|-|-|-|}
& & Corrected Akaike Information Criterion (AICc) & $AIC + \frac{2k(k+1)}{n-k-1}$ & AIC with correction for small sample sizes. \\ \hline

\multirow{9}{*}{Probabilistic} & \multirow{3}{*}{Error-Based Metrics} & Logarithmic Score (Log Score) & $ -\frac{1}{n} \sum_{t=1}^{n} \log(p_t)$ & Evaluates the difference between predicted probabilities and actual outcomes using a logarithmic function. \\ \hhline{|~|~|-|-|-|}
& & Continuous Ranked Probability Score (CRPS) & $\int_{-\infty}^{\infty} (\hat{F}(z) - \mathbf{1}(z \geq y_t))^2 dz$ & Evaluates the difference between the predicted probability distribution and the observed value using the cumulative distribution function. \\ \hhline{|~|-|-|-|-|}


& \multirow{3}{*}{Interval Metrics} & Prediction Interval Coverage Probability (PICP) & $ \frac{1}{n} \sum_{t=1}^{n} I\left(y_t \in [\hat{y}_{\text{lower},t}, \hat{y}_{\text{upper},t}]\right) $ & Measures the proportion of observed values that fall within the predicted intervals. \\ \hhline{|~|~|-|-|-|}
& & Prediction Interval Width (PIW) & $ \frac{1}{n} \sum_{t=1}^{n} \left( \hat{y}_{\text{upper},t} - \hat{y}_{\text{lower},t} \right)
$ & Evaluates precision by measuring the width of prediction intervals. \\ \hhline{|~|-|-|-|-|}

& Quantile Metrics & Quantile Loss (Pinball Loss) & $\frac{1}{n} \sum_{t=1}^{n} 
\begin{cases} 
\tau \cdot (y_t - \hat{y}_t) & \text{if } y_t \geq \hat{y}_t \\
(1 - \tau) \cdot (\hat{y}_t - y_t) & \text{if } y_t < \hat{y}_t
\end{cases}$ & Penalizes over- and under-predictions based on quantile $\tau$. \\ \hhline{|~|-|-|-|-|}

& Sharpness Metrics & Sharpness & $\frac{1}{n} \sum_{i=1}^{n} \text{Var}(\hat{y}_i)$ & Evaluates the concentration by the width of intervals or variance. \\

\arrayrulecolor{black}\specialrule{1.0pt}{0pt}{0pt} 
\end{tabular}

\end{sidewaystable}
\section{Historical TSF Models}\label{sec3}

\subsection{Conventional Methods (Before Deep Learning)}\label{sec3-1}

\subsubsection{Statistical Models}

Prior to machine learning and deep learning, traditional statistical models, which laid the foundation for analyzing sequential data, were commonly utilized for time series forecasting.
Exponential smoothing was introduced by \cite{brown1959statistical}, \cite{holt1957forecasting}, and \cite{winters1960forecasting} as a method to forecast future values using a weighted average of past data. This method operates by computing an exponential weight decay of historical data, assigning greater weight to more recent data.
The Autoregressive Integrated Moving Average (ARIMA) model was formalized through the work of George Box and Gwilym Jenkins in their book ``Time Series Analysis: Forecasting and Control'' \citep{box1970forecasting}. The ARIMA model predicts future values by leveraging the autocorrelation in data and is composed of three main components. AutoRegressive (AR) uses a linear combination of past values to predict the current value. Moving Average (MA) employs a linear combination of past error terms to predict the current value. Integrated (I) removes non-stationarity (the property where mean and variance change over time) by differencing the data to achieve stationarity.
While exponential smoothing models are advantageous for real-time data analysis due to their simplicity, ARIMA models are better suited for capturing complex patterns, making them ideal for long-term forecasting.
The SARIMA model extends ARIMA by incorporating seasonal differencing, along with seasonal autoregressive and moving average components, allowing it to effectively model and predict data with regular cyclical patterns. These statistical models are based on specific assumptions and are simple, intuitive, and useful for identifying basic patterns in data.

\subsubsection{Machine Learning Models}
While statistical methods struggle to capture complex patterns in time series data due to their reliance on predefined linear relationships, machine learning models excel in learning nonlinear patterns directly from the data without relying on such assumptions. To address the limitations of statistical methods, traditional machine learning models have been increasingly applied to time series forecasting.
Decision Trees \citep{quinlan1986induction} are machine learning models that use a tree structure for classification or prediction. The term ``classification and regression tree (CART) analysis'' was first introduced by \cite{barlin2013classification}, and in time series forecasting, regression decision trees are used to split data into a tree structure to predict continuous values. While they are intuitive and easy to interpret, they are prone to overfitting.
Support Vector Machines (SVM), introduced by Cortes and Vapnik \citep{cortes1995support}, are supervised learning models used for classification and regression analysis, characterized by finding the maximum margin. They handle high-dimensional data and non-linearities effectively, making them robust for classification and regression tasks. Support Vector Regression (SVR) applies SVM concepts to regression problems, predicting data points using an optimal hyperplane \citep{vapnik1996support}. This hyperplane is trained to minimize errors between data points, ensuring errors do not exceed a certain threshold.
Gradient Boosting Machines (GBM) were developed in 1999 with an explicit regression gradient boosting algorithm. This method uses ensemble learning to combine multiple weak models into a single, strong predictive model \citep{friedman2001greedy}. This method iteratively improves the model, offering high predictive performance. XGBoost, an extension of the Gradient Boosting algorithm \citep{xgboost}, incorporates various optimization techniques to enhance efficiency and performance. It gained significant attention for its outstanding performance in machine learning competitions, particularly in time series forecasting challenges.
These machine learning models outperform traditional statistical models in capturing data structures and patterns, demonstrating high predictive power on large datasets \citep{kontopoulou2023review}. Their ability to automatically learn and optimize various data features and relationships has led to their widespread use in time series forecasting.

\subsection{Fundamental Deep Learning Models}\label{sec3-2}

\subsubsection{MLPs: The Emergence and Constraints of Early Artificial Neural Networks}
The development of the \textbf{Multi-layer Perceptron (MLP)} \citep{MLP} and the backpropagation algorithm established the foundation for artificial neural networks. Early artificial neural network models utilizing these MLPs demonstrated strong performance in modeling nonlinear patterns, prompting numerous attempts to apply them to time series data processing. However, several key limitations, as outlined below, restricted the training of deep neural networks using MLPs.
\begin{itemize}
    \item \textbf{Limited in learning temporal dependencies}: MLPs are well-suited for processing fixed-length input vectors but struggle to adequately capture the temporal dependencies in time series data.
    \item \textbf{Vanishing gradient issue}: The vanishing gradient problem in deep networks made training difficult.
    \item \textbf{Lack of data and computing resources}: The scarcity of large-scale datasets and high-performance computing resources made it challenging to effectively train complex neural network models.
\end{itemize}

At that time, artificial neural network technology was still immature, and there was a lack of deep understanding and effective methodologies for dealing with time series data. Consequently, traditional statistical methods and machine learning models previously discussed continued to be widely used for time series analysis.

\subsubsection{RNNs: The first neural network capable of processing sequential data and modeling temporal dependencies}
\paragraph{The Emergence and Early Applications of RNNs}
\textbf{Recurrent Neural Networks (RNNs)} \citep{RNN} emerged, opening up new possibilities for processing time series data.
RNNs are specialized models designed to process sequential data, such as time series data, and were utilized to overcome the limitations of MLPs. The structure of an RNN consists of an input layer, a hidden layer, and an output layer, and it uses the output from previous time steps(hidden layer output) as the input for the current time step. This structural feature allows RNNs to naturally model the dependencies in data over time. The hidden state preserves past information by storing data from the previous step and passing it on to the next step, thereby maintaining historical context. Additionally, it is parametrically efficient because the same weights are used for calculations across all time steps, regardless of the sequence length.

Due to these features, RNNs have been utilized in various fields where sequential data modeling is crucial, such as time series data analysis, natural language processing (NLP), and so on. However, early RNNs had the following issues.
\begin{itemize}
    \item \textbf{Vanishing gradient problem}: When RNNs learn long sequences, the gradient gradually diminishes during the backpropagation process, causing information from earlier time steps to fail to reach later time steps. Thus, this made it difficult to learn long-term dependencies.
    \item \textbf{Exploding gradient problem}: In contrast to the vanishing gradient problem, the gradient values could become excessively large, making the training process unstable.
    \item \textbf{Difficulty in parallel processing}: Due to the sequential nature of RNNs, where the computation at each time step depends on the previous one, parallel processing becomes challenging. This leads to high computational costs and slow learning speeds.
\end{itemize}

Because of these critical drawbacks, RNNs were not widely used until the early 2000s.

\paragraph{Overcoming the Limitations of RNNs}
To address the aforementioned issues with vanilla RNNs, researchers began developing various models. Notably, the \textbf{Long Short-Term Memory (LSTM)} \citep{LSTM} model was developed to address the long-term dependency issues of vanilla RNNs. With the cell state and gate mechanisms (input, output, and forget gates), LSTMs preserve crucial information over extended periods and discard unnecessary information. The \textbf{Gated Recurrent Unit (GRU)} \citep{GRU} emerged as a simpler alternative to LSTM, offering similar performance while using a more straightforward structure(update and reset gates) to manage information. These two models effectively addressed the vanishing gradient problem and sparked a boom in RNN-based models for time series analysis. Many subsequent RNN-based models used these two as their backbone, making them powerful tools for handling time series data until the advent of Transformers.

\paragraph{Notable RNN Variants}
The \textbf{Dilated RNN} \citep{DilatedRNN} proposed Dilated Recurrent Skip Connections as a solution to various issues inherent in fundamental vanilla RNNs. This approach enhanced parameter efficiency in long sequences and enabled the network to effectively learn complex dependencies across diverse temporal scales.
The \textbf{DA-RNN} \citep{DA-RNN} attempted to effectively address long-term dependencies, which were challenging for existing methods, by employing a dual-stage attention mechanism. Utilizing an encoder with Input Attention and a decoder with Temporal Attention, it adaptively selected salient information from input features and temporal steps, thereby enhancing the accuracy and performance of time series predictions. Additionally, the \textbf{MQ-RNN} \citep{MQ-RNN} combined the Sequence-to-Sequence neural network framework with Quantile Regression to provide probabilistic forecasts, thereby reflecting uncertainties across various scenarios. This model simultaneously generated multiple forecast horizons at each time step and employed the Forking-Sequences training method to enhance the stability and performance of the RNN.

\subsubsection{CNNs: Extracting key patterns in time series data beyond just images}
\paragraph{The Emergence and Early Applications of CNNs}
The \textbf{Neocognitron} \citep{Neocognitron} was designed for visual pattern recognition and served as an early form of Convolutional Neural Networks (CNNs), introducing the fundamental concepts of CNNs.
The CNN architecture, which became more widely known with the development of \textbf{LeNet} \citep{LeNet}, learned spatial patterns in images through a combination of convolutional and pooling layers. These features made CNNs well-suited for handling image data, leading early CNN-based models to primarily focus on image data without being directly applied to time series data. However, over time, their potential for handling time series data has also been recognized.

\paragraph{Attempts to Apply CNNs to Time Series Data}
1D CNNs use one-dimensional convolutional filters to learn local patterns in time series data, allowing them to extract features that consider the temporal structure of the data. A prominent example is \textbf{WaveNet} \citep{WaveNet}, which utilized 1D convolutions and dilated convolutions to model speech signals, effectively capturing long-term dependencies in audio signals. WaveNet demonstrated the utility of CNN-based models in speech synthesis and time series prediction, leading to the widespread adoption of CNN-based models for time series analysis.
The development of the \textbf{Temporal Convolutional Networks (TCNs)} \citep{TCN} model further highlighted the potential of CNN-based models in the time series domain. TCN consists of multiple layers of 1D convolutional networks, using dilated convolutions at each layer to achieve a wide receptive field, enabling the construction of deeper networks. This allows TCNs to effectively learn from long sequences of data. TCNs have demonstrated excellent performance in various sequential domains, including time series prediction, signal processing, and natural language processing.

\paragraph{CNN and RNN \revision{Hybrid} Models}
Building on the popularity of fundamental time series forecasting models like RNNs and the emerging interest in CNN models, hybrid CNN-RNN models began to emerge, combining the strengths of both architectures. Models that combined CNNs and LSTMs were particularly advantageous for simultaneously learning local patterns and long-term dependencies in time series data. The structure involved using CNNs to extract complex features from the time series data and LSTMs to learn temporal dependencies.

\textbf{DCRNN} \citep{DCRNN} modeled traffic flow as a diffusion process on a directed graph, capturing both temporal and spatial dependencies. The Diffusion Convolution leveraged bidirectional random walks on the graph to capture spatial dependencies within the traffic network, while the Gated Recurrent Units (GRU) modeled temporal dependencies. The introduction of the Diffusion Convolutional Gated Recurrent Unit (DCGRU) enabled the effective processing of temporal information.
\textbf{TPA-LSTM} \citep{TPA-LSTM} combined CNN and LSTM with an added attention mechanism, introducing the Temporal Pattern Attention (TPA) methodology to learn important patterns in time series data. Local patterns were extracted using CNN, while the combination of LSTM and the attention mechanism captured significant temporal patterns.

\subsubsection{GNNs: Structurally modeling relationships between variables}
\paragraph{The Emergence and Slow Growth of GNN-Based Models}
\textbf{Graph Neural Networks (GNNs)} \citep{GNN} were primarily developed to process graph-structured data because they can effectively model the complex structural relationships between nodes and edges in graph data.
Over time, GNNs gradually began to be applied to time series data analysis, primarily because multivariate time-series data often encapsulates intricate structural relationships.

During their initial development, GNNs were less popular compared to widely used RNN or CNN-based models. The prevailing belief was that GNNs were not suitable for time series analysis compared to CNNs which are effective at extracting local patterns, and RNNs which are strong at learning long-term dependencies. However, as the range and complexity of time-series data increased, understanding their structural characteristics became more important. Consequently, GNNs began to gain prominence in fields such as traffic prediction and social networks. Research on applying GNNs to time-series data has primarily focused on capturing the dynamic characteristics of graph data, and the structural learning capabilities of GNNs have proven to be highly useful in capturing the complex patterns within time-series data.

\paragraph{Development and Expansion of GNN Applications}
Until the development of Transformers, GNNs consistently garnered interest, showing consistent yet slow growth compared to RNNs and CNNs.
The development of \textbf{Graph Convolutional Networks (GCNs)} \citep{GCN} marked another turning point for GNNs. GCNs learn node features through convolution operations on graph structures and are powerful tools for processing graph data. This advancement positioned GCNs as a versatile solution for addressing a wide range of graph-based challenges. 

GCNs were more effective at learning the relationships between variables in time series and temporal locality, and subsequently, GNNs began to be widely used for time-series data. \textbf{Spatial-Temporal Graph Convolutional Networks (ST-GCN)} \citep{ST-GCN} were developed as a model combining GCNs and RNNs to learn complex patterns in spatio-temporal data. By leveraging GCNs to learn spatial patterns and RNNs to capture temporal patterns effectively, ST-GCN demonstrated the potential of GNNs in time-series problems, such as traffic prediction.
\textbf{Graph Attention Networks (GATs)} \citep{GAT} were introduced, learning relationships between nodes using an attention mechanism. This allowed GATs to capture the importance of nodes, making them advantageous for identifying significant events in time-series data.
\textbf{Dynamic Graph Neural Networks (DyGNNs)} \citep{DyGNN} were developed to learn graph structures that change over time. Designed to reflect the dynamic characteristics of time-series data, DyGNNs could effectively capture complex temporal dependencies and structural relationships by allowing the graph structure to evolve over time.
\textbf{Temporal Graph Networks (TGNs)} \citep{TGN} were introduced as a model in which the nodes of a graph change over time, enabling the learning of temporal patterns in time-series data. By using a memory module to track the temporal state changes of nodes, TGNs learn the information of neighboring nodes whose importance varies over time, thus providing a basis for predicting future states.

\begin{figure}[H]
\centering
\includegraphics[width=1.0\textwidth, trim=0cm 4cm 0cm 4cm, clip]{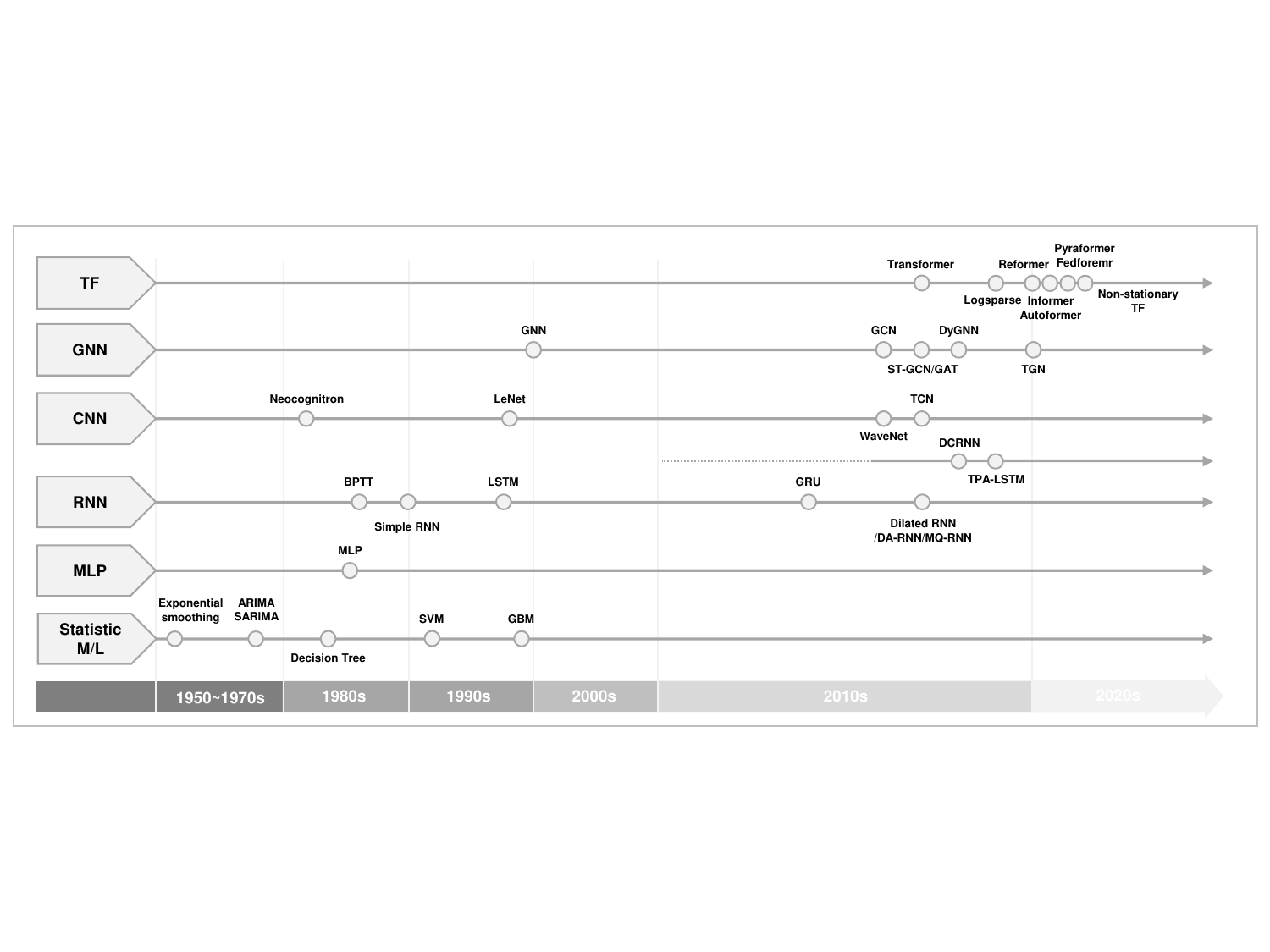}
\caption{Remarkable Historical TSF Models}\label{Remarkable Historical TSF Models}
\end{figure}

After the era of traditional statistical techniques and machine learning methodologies, fundamental deep learning approaches such as RNNs, CNNs, and GNNs were widely used for time-series analysis and prediction over a long period. However, the advent of Transformers brought a revolutionary change to the AI field. Upon their introduction, Transformers outperformed all existing methods across various domains and continue to dominate as the leading foundation model. In the time-series domain, a similar trend was observed. Beginning with the introduction of the \textbf{Logsparse Transformer} \citep{LogSparseTransformer}, numerous transformer-based models emerged, achieving state-of-the-art performance. As a result, fundamental deep learning methods experienced a period akin to the Ice Age.
The remarkable historical TSF models mentioned above are organized chronologically in Fig. \ref{Remarkable Historical TSF Models}.

\subsection{The Prominence of Transformer-based Models}\label{sec3-3}

\textbf{Transformer} \citep{Transformer}, is an innovative model designed to perform complex sequence-to-sequence tasks in natural language processing.
This model features an encoder-decoder structure and utilizes a self-attention mechanism to capture the relationships between tokens in the input sequence.

\revision{Modeling approaches in self-attention can be broadly categorized into unidirectional and bidirectional paradigms.
Unidirectional modeling restricts the self-attention mechanism to consider only the tokens preceding the current token in the sequence.
For a sequence $( x_1, x_2, \ldots, x_L )$, the attention for token $x_t$ is computed over the subsequence $( x_1, x_2, \ldots , x_t )$ to ensure causality, where $L$ denotes the length of the input sequence.
The unidirectional modeling is predominantly employed in tasks requiring sequential generation, such as language generation~\citep{GPT3, transformer-xl}. 
In contrast, bidirectional modeling allows the self-attention mechanism to utilize the entire sequence. 
For token $x_t$, attention is computed over $( x_1, x_2, \ldots , x_L )$, thereby leveraging both preceding and succeeding tokens. 
The bidirectional modeling is employed in tasks requiring holistic context, such as masked token prediction~\citep{BERT, ELECTRA}.}

As a result, Transformers have gained significant attention for replacing fundamental RNN-based models, offering parallel processing capabilities, and effectively addressing long-term dependency issues.
Transformers' success with sequential data naturally led to their extension into time series applications. 
\revision{The attention mechanism plays a significant role in TSF by considering the relationships across all time steps simultaneously, giving more weight to important moments, and effectively learning the trends and patterns in the data.}

This section explores various Transformer variants that address the limitations of the original Transformer for time series forecasting tasks.
It discusses the background of these variants, how they improve upon the original Transformer, and the reasons for their enhanced performance in time series prediction.
Additionally, it addresses the remaining limitations of these variant models.

\subsubsection{Transformer Variants}
In recent years, substantial research has focused on transformer-based time series analysis, particularly long-term time series forecasting (LTSF), which has led to the development of various models \citep{DLinear}.
The original Transformer model has several limitations when applied to long-term time series forecasting (TSF).
These limitations include quadratic time and memory complexity, as well as error accumulation caused by the auto-regressive decoder design.
\revision{Specifically, the time and memory complexity of self-attention increases quadratically with the length of the input sequence, denoted as \( O(L^2) \), where \( L \) represents the length of the input sequence.}

\begin{figure}[H]
    \centering
    \begin{subfigure}[b]{0.45\textwidth}
        \centering
        \includegraphics[width=0.85\textwidth, trim=0cm 0.5cm 0cm 0.5cm, clip]{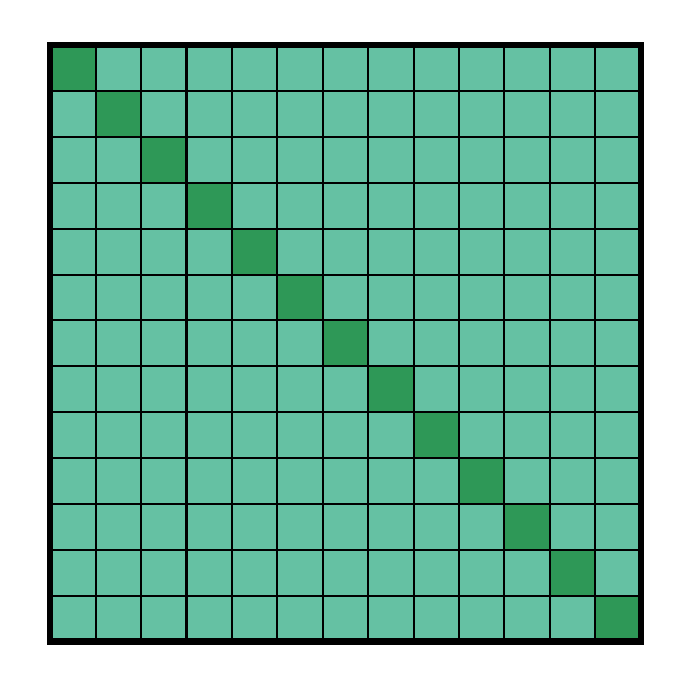}
        \caption{\revisionB{Full Self-Attention}}
        \label{fig:attention_full}
    \end{subfigure}
    \hfill
    \begin{subfigure}[b]{0.45\textwidth}
        \centering
        \includegraphics[width=0.85\textwidth, trim=0cm 0.5cm 0cm 0.5cm, clip]{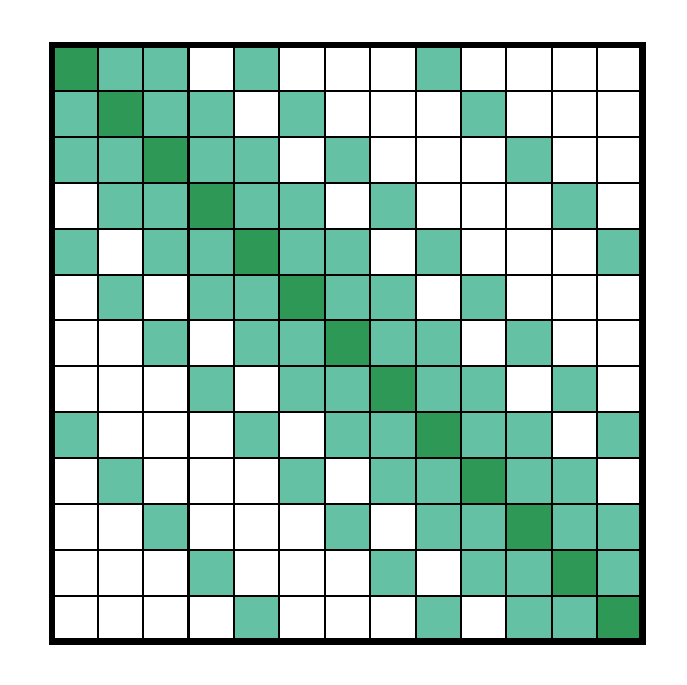}
        \caption{\revisionB{Sparse Self-Attention}}
        \label{fig:attention_sparse}
    \end{subfigure}
    \caption{\revisionB{Comparison of Full and Sparse Self-Attention Patterns}}
    \label{fig:comparison of full and sparse attention}
\end{figure}

This high complexity can become a computational bottleneck, especially for time series forecasting tasks that rely on long-term historical information.
To address these issues, Transformer variants primarily focus on reducing time and memory complexity.
\revisionB{An attention pattern determines which tokens a token attends to. Fig. \ref{fig:comparison of full and sparse attention} compares Full Self-Attention and Sparse Self-Attention patterns to illustrate the efficiency of Sparse Attention \citep{Beltagy2020Longformer}. In Fig. \ref{fig:attention_full}, all tokens attend to each other, sharing global information. In contrast, Fig. \ref{fig:attention_sparse} shows that only a subset of tokens is attended to, reducing computational cost while preserving essential information.}

\textbf{LogTrans} \citep{LogSparseTransformer} leverages the efficiency of Log Sparse attention mechanisms to reduce time and memory complexity to \( O(L log L) \), making it well-suited for long sequence forecasting tasks.
This model assumes recent data is more relevant than older data, utilizing positional sparsity to focus attention only on relevant data.
\textbf{Reformer} \citep{Reformer} enhances efficiency through locality-sensitive hashing and reversible residual layers, reducing memory and computational requirements to \( O(L log L) \) while maintaining performance on long sequences.
Locality-sensitive hashing (LSH) approximates attention by clustering similar items and performing attention partially on these clusters.
\textbf{Informer} \citep{Informer} improves computational and memory efficiency to \( O(L log L) \) by utilizing a ProbSparse self-attention mechanism and a generative-style decoder.
It effectively handles long sequences and captures dependencies across various time scales. Informer addresses the limitations of Reformer, which performs well only on extremely long sequences, and LogTrans, which uses heuristic positional-based attention.
ProbSparse self-attention features dynamic query selection based on query sparsity measurement.
\textbf{Autoformer} \citep{Autoformer} integrates time series decomposition into the internal blocks of the Transformer and utilizes an auto-correlation mechanism to aggregate similar sub-series.
Autoformer improves upon previous methods by addressing point-wise dependency and aggregation issues. It employs a decomposition architecture to handle complex temporal patterns by separating seasonal and trend components.
\textbf{Pyraformer} \citep{Pyraformer} introduces a novel pyramid attention mechanism designed to efficiently model long-term dependencies in time series data.
By using a hierarchical structure to progressively reduce sequence length, Pyraformer significantly lowers computational complexity compared to fundamental Transformer models.
\textbf{Fedformer} \citep{Fedformer} combines Fourier-enhanced decomposition blocks with frequency domain sparse representations, integrating seasonal-trend decomposition from Autoformer and offering a more effective representation of time series data.
The sparse representation basis can be Fourier-based or Wavelet-based, encompassing a subset of frequency components, including both high and low frequencies.
Additionally, attempts have been made to overcome the limitations of normalized Transformers, which often produce indistinguishable attention patterns and fail to capture significant temporal dependencies.
\textbf{Non-stationary Transformer} \citep{nonstationaryTransformer} identifies this issue as excessive normalization and addresses it with De-stationary Attention, which allows for the use of non-normalized data during QKV computations.

\subsubsection{Limitation of Transformer-based Models}
However, even the models designed specifically for time series forecasting based on the original Transformer still have the following limitations.

\paragraph{Efficiency Challenges}
Efforts to overcome the primary drawback of quadratic scaling in computational and memory complexity with window length have not fully resolved the issues.
Various approaches to reduce this complexity have been proposed, but the attention mechanism inherently incurs higher computational and memory costs compared to MLP-based or convolutional models with \( O(L) \) complexity.
Applying Transformers to the time domain does not easily mitigate this problem.
Models such as \textbf{Fedformer} \citep{Fedformer}, which uses Fourier or Wavelet transformations to operate in the frequency domain, have been made to address these issues.
Although research into more efficient attention variants is ongoing \citep{EfficientTransformers}, these solutions often sacrifice some of the effective characteristics of Transformers.
Sparse attention mechanisms, used to reduce Self-Attention complexity, may result in the omission of important information.
Models like LogTrans and Pyraformer introduce explicit sparsity biases in the attention mechanism but may suffer from significant performance degradation due to the loss of crucial information.
None of these variants have yet been proven effective across diverse domains \citep{Mamba}.

\paragraph{\revision{Context Window Limited to Current Input}}
\revision{In principle, Transformers can handle inputs of arbitrary length, provided there are sufficient computational resources. However, transformers must process the entire context at once in a single input, causing the memory and computation requirements to grow substantially with longer context length. By contrast, RNNs and State Space Models (SSMs) maintain hidden states over time, allowing them to preserve context beyond the immediate input. We highlight this practical limitation, which often arises when handling very long contexts.}

\paragraph{Ineffectiveness of Expanding Input Window Length}
Another significant issue is the minimal or no performance improvement observed when increasing the input window length.
Strong time series forecasting (TSF) models are generally expected to achieve better results with larger look-back window sizes due to their robust temporal relationship extraction capabilities.
However, research \citep{DLinear, Informer, TrasformersInTS} indicates that the performance of Transformers either deteriorates or remains stable as the look-back window increases, in contrast to improvements seen with linear-based methods \citep{DLinear}.
This suggests that transformer-based models tend to overfit noise rather than extract long-term temporal information when provided with longer sequences.

\vspace{1em}

\noindent
In conclusion, while transformer-based models have made significant advancements in time series forecasting, they still face limitations.
Therefore, future research will continue to address these limitations and will also serve as a catalyst for re-exploring various alternative architectures.

\subsection{Uprising of Non-Transformer-based Models}\label{sec3-4}


As previously discussed, transformer-based models demonstrate strong performance in processing time series data. However, they have several limitations when dealing with long-term time series data.
\begin{itemize}
    \item The point-wise operations of the self-attention mechanism have quadratic time complexity. Therefore, as sequence length increases, the computational load grows exponentially, making it challenging to handle long-term historical information.
    \item Because storing the relationship information for all input token pairs requires substantial memory usage, applying Transformers in environments with limited GPU memory becomes challenging.
    \item When the length of the look-back window exceeds the structural capacity, learning long-term dependencies becomes challenging.
    \item Due to the high complexity of the model, large-scale and high-quality datasets are required. In the absence of sufficient data, overfitting can occur leading to a drop in model performance.
\end{itemize}

Contemporary TSF tasks increasingly require the prediction of diverse multivariates and long sequences. Unlike earlier tasks that involved relatively short sequences, the limitations of transformer-based models have become more pronounced when dealing with long sequences. Researchers adhering to the Transformer's philosophy have begun focusing intensively on two aspects to address these issues: reducing computational complexity and enhancing long-term dependency learning. Conversely, some researchers have reverted to non-transformer methodologies, renewing interest in fundamental deep learning models such as RNNs, CNNs, GNNs, and MLPs. Non-transformer-based models have the potential to overcome the limitations of Transformers, and research has begun to leverage these strengths to move beyond the hegemony of the Transformer backbone.

\textbf{RNN-based} models sequentially process input data, retaining the previous state at each time step to predict the next state. This enables the natural modeling of long-term dependencies and helps overcome the limitations of Transformers by leveraging time-ordering learning. They require relatively less data and memory, making them less restrictive in their application. Additionally, their ability to handle real-time streaming data enhances their versatility across various applications.

Convolution operations in \textbf{CNN-based} models offer lower time complexity and reduced memory requirements compared to the self-attention mechanism in Transformers. Since time series data often exhibit periodic trends and seasonality, the ability of convolution to extract local patterns complements Transformers' specialization in learning global patterns, addressing their shortcomings.

\textbf{GNN-based} models can effectively represent and learn the complex relationships and structures of time series data through nodes and edges. They integrate localities efficiently through local operations, resulting in lower memory usage. Additionally, their ability to model both spatial and temporal relationships simultaneously makes them advantageous for handling multivariate datasets.

The primary advantage of \textbf{MLP-based} models lies in their simple structure, which facilitates efficient training, fast inference, and straightforward handling. Current AI models require lightweight designs suitable for mobile applications, and MLP-based models are well-suited for environments with many constraints.

Especially, the counterattack of simple linear models based on MLPs is one of the notable trend shifts. The emergence of \textbf{LTSF-Linear} \citep{DLinear} models which surpass transformer-based models using only simple linear layers has triggered this change. One of the key characteristics of time series data is that the time ordering of data points holds important meaning. Conversely, the permutation-invariant nature of the Transformer's attention mechanism, which emphasizes relationships between data points, is argued to be inherently unsuitable for time series data. Although positional embedding attempts to preserve time order, the multi-head attention process inevitably leads to information loss \citep{Informer}.

\textbf{LTSF-Linear} models highlighted the critical limitations of Transformers in handling long sequences, demonstrating that simple linear models could better preserve time-ordering information and extract features related to trends and seasonality.

This claim gained further traction through subsequent studies, leading to a more diversified interest in backbone models across various architectures. The emergence of this model sparked skepticism regarding the performance of Transformer-based models and served as a catalyst for the exploration of various alternative architectures. The era of Transformer prevalence has come to an end, ushering in a renaissance in time series research, characterized by a reevaluation of fundamental DNNs, the emergence of novel models like Mamba, and new approaches such as the utilization of LMMs.

\section{New Exploration of TSF Models}\label{sec4}

\subsection{Overcoming Limitations of Transformer}\label{sec4-1}

As previously mentioned in \ref{sec3-4}, the performance of simple linear models in \textbf{LTSF-Linear} \citep{DLinear} surpassing fundamental transformer-based models has raised doubts about the effectiveness of Transformers.
However, in the field of NLP, Transformers still demonstrate superior performance in handling long-term dependencies in sequential data compared to other models \citep{patwardhan2023transformers}.
This observation suggests that while Transformers have great potential, researchers have not fully leveraged their capabilities in time series analysis.

Therefore, various methods have emerged in time series forecasting to overcome the limitations of existing Transformer-based models.
This section categorizes and explains in detail the specific limitations of existing models and how these limitations are addressed.
The structure begins with an introduction to the patching technique, followed by the use of cross-dimension and exogenous variables, and then provides a detailed explanation of other approaches, concluding with a summary of key points in Table \ref{tab:RecentsTransformerModels}.

\subsubsection{Patching Technique}
Transformers were originally developed for natural language processing (NLP), and applying them to time-series analysis requires appropriate adjustments to fit the domain of time series.
Different from the rich semantic information carried by individual word tokens in NLP, individual data points in time series are often similar to their neighboring values, lacking substantial information \citep{PatchTST}.
Therefore, the point-wise attention mechanism of fundamental models fails to consider the broader context or patterns that may span multiple consecutive time steps and only calculates attention for individual time steps, which makes it difficult to capture the characteristics of time series data.
For this reason, considering the surrounding context along with a single time point provides more information in time series data.

\begin{figure}[!h]
\centering
\includegraphics[width=0.7\textwidth, trim=0cm 2cm 0cm 2cm, clip]{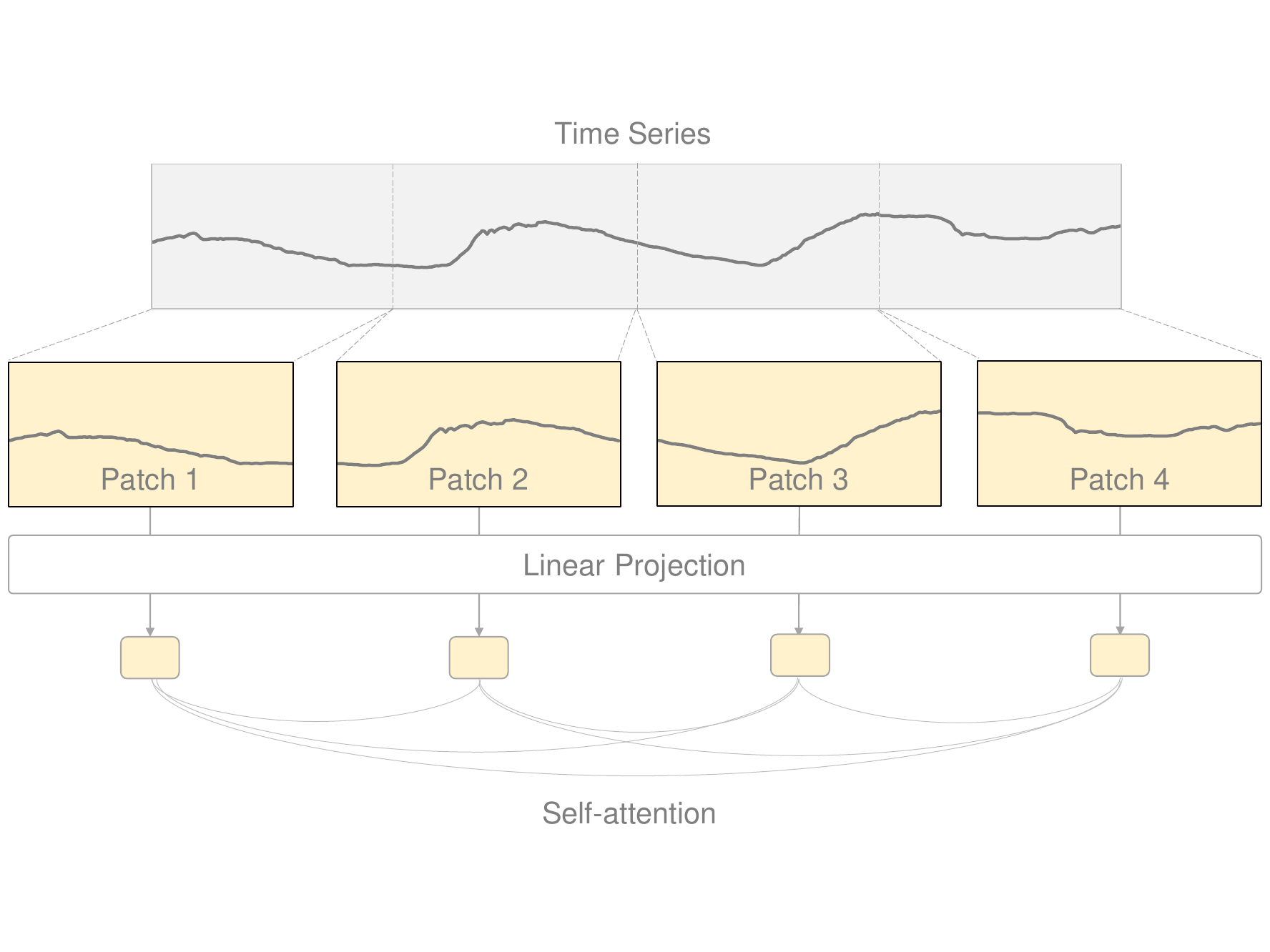}
\caption{Patching Technique in Self-Attention for Time Series Forecasting}\label{Patching Technique in Self-Attention for Time Series Forecasting}
\end{figure}

Patching refers to the technique of dividing input sequences into multiple patches, as illustrated in Fig. \ref{Patching Technique in Self-Attention for Time Series Forecasting}.
This method preserves the information within each patch, thereby enhancing locality.
By processing patches instead of individual points, the model processes fewer tokens, thereby reducing the computational complexity of the attention mechanism.
This approach helps overcome the issue of prediction performance degradation, which can occur when sparse attention is used to make self-attention more efficient, potentially missing critical information.

\textbf{PatchTST} \citep{PatchTST}, inspired by Vision Transformer's \citep{ViT} division of images into $16 \times 16$ patches for capturing local semantic information, divides time series into 64 patches. This approach allows for utilizing a longer look-back window by grouping data into patches. By processing time-series data in patch units, PatchTST maximizes the advantages of Transformer models for time-series applications, achieving better performance than LTSF-Linear. PatchTST employs only the Transformer encoder, flattens the encoder output, and uses a linear layer for the final predictions. Additionally, the study shows that channel independence (CI) yields better performance, highlighting the limitations in learning channel correlations despite the intuitive consideration of channel dependencies (CD) in multivariate scenarios.
\textbf{MTST} \citep{MTST} addresses the limitation of PatchTST in learning patterns across different scales present in time-series data. By adopting a multi-scale approach, MTST proposes an effective model utilizing both shorter and longer patches for locality and long-term trend analysis.
\textbf{PETformer} \citep{Petformer} critiques the flattening of Transformer encoder output in PatchTST, which results in a significant increase in parameters. PETformer introduces a placeholder-enhanced technique, modeling past and future data on the same time scale, thereby reducing the number of parameters by over 95\% compared to PatchTST. This reduction enhances generalization performance while using less memory and computational resources. The model also leverages rich context information by allowing direct interaction between past and future data, maintaining the continuity of time-series data.

By applying the patching technique, it was possible to refute the notion that traditional linear models are superior to Transformers. This led to the patching technique becoming a widely adopted approach.

\subsubsection{Cross-Dimension}
In multivariate time-series forecasting, understanding the relationships between variables is crucial for improving prediction accuracy.
Intuitively, higher temperatures lead to increased ice cream sales, indicating a relationship between variables.
Despite this apparent correlation, surprising results have shown that models treating channels independently, such as LTSF-Linear, PatchTST, and PETformer, often outperform those considering inter-channel correlations.
This result implies that current models fail to effectively capture the relationships between variables.
Time series analysis differs significantly from natural language processing (NLP) and computer vision (CV) in terms of channel correlation.
In NLP, there is no clear concept of channels.
In CV, although channels exist, their relationships are tightly intertwined and well-defined, as seen in the RGB representation of images.

Conversely, in time series analysis, channel relationships can be either independent or interdependent and often hidden, adding complexity to the task.
Therefore, time series analysis requires models capable of capturing these intricate correlations.
While earlier Transformer-based models primarily focused on temporal attention, recent models have increasingly focused on explicitly modeling the correlations between variables.

\begin{figure}[!h]
\centering
\includegraphics[width=0.8\textwidth, trim=0cm 2cm 0cm 2cm, clip]{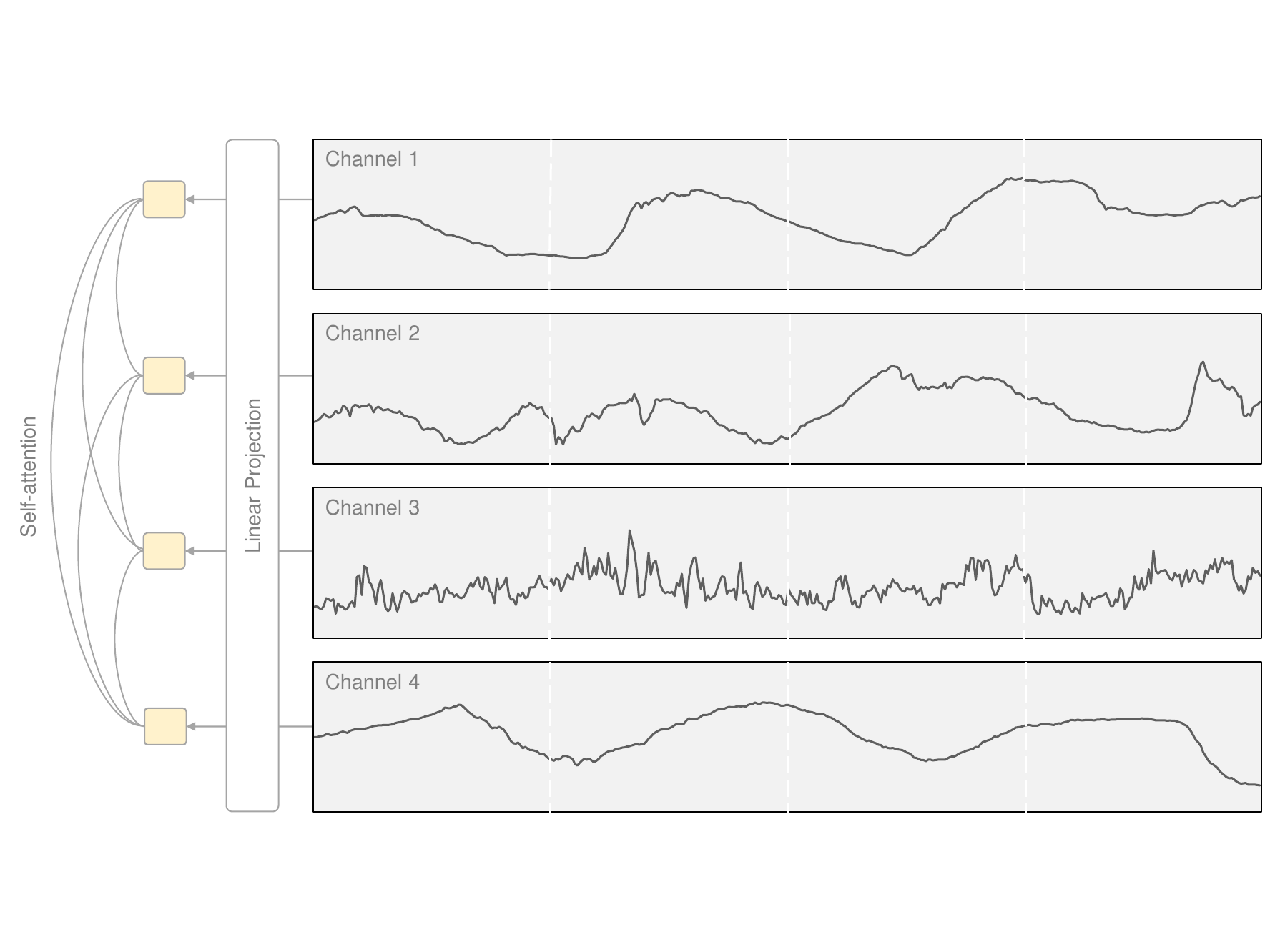}
\caption{Cross-Dimensional Self-Attention for Modeling Variable Relationships}\label{Cross-Dimensional Self-Attention for Modeling Variable Relationships}
\end{figure}

Fig. \ref{Cross-Dimensional Self-Attention for Modeling Variable Relationships} illustrates the explicit modeling of relationships between variables using the attention mechanism.

The overall progression of advancements is as follows. It begins with modeling that explicitly accounts for correlations between variables. This progresses to directly modeling the relationships between both temporal and variable aspects. Additionally, the model incorporates the possibility of time lags in the relationships between variables, allowing it to flexibly learn dependencies.

\textbf{Crossformer} \citep{Crossformer} breaks away from solely temporal attention by employing a Two-Stage Attention mechanism that sequentially processes Cross-Time and Cross-Dimension stages. It divides each dimension's time series into patches, embeds them into 2D vectors, and performs attention. Crossformer incorporates a router mechanism in the Cross-Dimension stage to handle the increased complexity of dual attention. This hierarchical structure across multiple scales enables Crossformer to effectively model both temporal dependencies and inter-dimensional correlations, significantly enhancing multivariate time series forecasting performance.
\textbf{DSformer} \citep{DSformer} criticizes Crossformer for emphasizing global interactions between variables in time series data, while overlooking the importance of local information, such as short-term variations and patterns. DSformer addresses this by using double sampling, obtaining global information via downsampling and local information through piecewise sampling. These samples undergo parallel temporal variable attention, allowing the model to integrate global, local, and inter-variable correlations in multivariate time series data through dual sampling and time-variable attention mechanisms.
\textbf{CARD} \citep{CARD} points out the structural complexity and high computational cost of Crossformer's hierarchical attention mechanism. Simplifying the structure, CARD employs only the encoder and uses a lightweight token blend module instead of explicitly generating token hierarchies. This module merges adjacent tokens to represent larger ranges, efficiently leveraging multi-scale information and enhancing prediction accuracy. A robust loss function is introduced to prevent overfitting, adjusting weights to favor near-future data over far-future data, which shows less temporal correlation.
\textbf{iTransformer} \citep{itransformer} innovatively reverses the time and variable dimensions in the standard Transformer model, learning inter-variable correlations. By embedding each variable into variate tokens via a Transformer encoder and performing temporal predictions through a feed-forward network, iTransformer surprisingly outperforms Crossformer, highlighting the limitations of merely adding explicit channel attention.
Its advantage lies in not needing to alter the original Transformer structure.
\textbf{VCformer} \citep{VCformer} addresses the issue of neglecting time lags in variable correlations. Its variable correlation attention module calculates cross-correlation scores between queries and keys across various lags using FFT (Fast Fourier Transform), efficiently exploring and adaptively integrating correlations for each lag.
\textbf{GridTST} \citep{GridTST} transforms time series data into a grid format, applying horizontal and vertical attention mechanisms to model time and variable correlations effectively. Like PatchTST and iTransformer, GridTST leverages the original Transformer, distinguishing itself from the more complex Crossformer. It experiments with three configurations: time-first, channel-first, and cross-application, finding that prioritizing channel attention yields the best performance, aligning with human intuitive analysis of inter-variable relationships.
\textbf{UniTST} \citep{uniTST} critiques the inability of cross-time and cross-dimension attention methods to directly and explicitly learn complex inter-variable and intra-variable dependencies. It proposes a unified attention mechanism by flattening patch tokens. To mitigate the increased computational cost, UniTST incorporates a dispatcher module to reduce memory complexity and enhance model feasibility.
\textbf{DeformTime} \citep{DeformTime} introduces deformable attention, dynamically adjusting to recognize and adapt to important data features instead of fixed attention areas. Parallel deformable attention mechanisms within the Transformer encoder capture correlations between variables and time, adaptively extracting critical information across various time intervals and variable combinations, significantly improving prediction accuracy and generalizability.

\subsubsection{Exogenous Variable}
Most existing research primarily utilizes endogenous variables for prediction. However, in real-world scenarios, relying solely on endogenous variables can be insufficient due to the complexity of various influencing factors. For instance, stock price predictions are significantly affected by external factors such as economic indicators, political changes, and technological advancements. Ignoring these external factors and relying only on past data of endogenous variables can lead to failures in accurately predicting market volatility. Therefore, incorporating exogenous variables as supplementary information has emerged as a method to improve prediction performance.

\textbf{TimeXer} \citep{TimeXer} proposes a method to integrate exogenous variables into the canonical Transformer model without structural changes. It operates by dividing the time series data of endogenous variables into patches, learning the temporal dependencies of each patch through self-attention. Then, it generates variate tokens summarizing the entire series of endogenous and exogenous variables and learns their interactions using a cross-attention mechanism. Through this process, TimeXer simultaneously considers the temporal patterns of endogenous variables and the impact of exogenous variables, enabling more precise and in-depth time-series predictions.
However, TimeXer requires manual identification and input of appropriate exogenous variables. If unsuitable data is provided, it can hinder the model's predictive accuracy.
\textbf{TGTSF} \citep{TGTSF} integrates text data from channel descriptions and news messages to enhance prediction accuracy. It embeds channel descriptions and news messages using a pre-trained text model, transforming them into a sequence of vectors over time. The cross-attention layer then calculates the relevance of the text to each channel. By incorporating text data into the time series prediction model, TGTSF not only improves prediction accuracy but also allows for direct comparison of the impact of textual information on predictive performance.

\subsubsection{Additional Approaches}
Beyond the topics discussed above, there are various other approaches to time series forecasting.

\paragraph{Generalization}
Research has been conducted to improve model generalization, avoid overfitting, and achieve consistent performance across diverse datasets.
\textbf{SAMformer} \citep{SAMformer} addresses the issue of self-attention mechanisms converging to sharp local minima, causing entropy collapse during training, and demonstrates that applying SAM (Sharpness-Aware Minimization) can significantly enhance performance. \textbf{Minusformer} \citep{minusformer} highlights the redundancy and overfitting caused by a large number of parameters in Transformers. To combat this, it employs a boosting ensemble method, where each subsequent model predicts the residuals of the previous model's outputs, thus reducing redundancy and improving generalization.

\paragraph{Multi-scale}
The multi-scale approach extracts more information from time series data across various scales, offering distinct advantages. \textbf{Scaleformer} \citep{scaleformer} proposes a general framework by stacking existing models across different scales, resulting in improved performance. \textbf{Pathformer} \citep{pathformer}, on the other hand, allows the model to learn adaptive scales independently, rather than relying on fixed scales.

\paragraph{Decoder-only}
Large-scale language models (LLMs) like LLaMA3 \citep{dubey2024llama} have been successfully implemented using only a decoder without the need for an encoder.
The decoder-only architecture is simpler and involves less complex computations, resulting in faster training and inference.
Additionally, the decoder-only structure helps avoid the temporal information loss often associated with the self-attention mechanism in encoders.
This has led to a research proposal aimed at improving performance in time series forecasting using a decoder-only structure.
\textbf{CATS} \citep{kim2024self} addresses the high time and memory complexity of self-attention in Transformer encoders and the loss of temporal order information. It proposes a structure using only cross-attention, focusing solely on the relationship between future and past data with a decoder-only architecture, which reduces parameter count and enhances efficiency. In contrast to the encoder-only structure of most models discussed CATS demonstrates the effectiveness of using only the decoder in achieving superior performance.

\paragraph{Feature Enhancement}
\textbf{Fredformer} \citep{fredformer} identifies the issue of frequency bias in time series data, where learning tends to focus disproportionately on either low or high frequencies. It addresses this by normalizing the frequencies to eliminate bias. \textbf{Basisformer} \citep{basisformer} proposes a method to construct flexible relationships with each time series by leveraging biases learned through contrastive learning.

\medskip

\begin{sidewaystable}

\centering
\captionsetup{justification=raggedright, singlelinecheck=false} 
\caption{Taxonomy and Methodologies of Transformer Models for Time Series Forecasting}
\label{tab:RecentsTransformerModels}

\renewcommand{\arraystretch}{1.9} 
\arrayrulecolor[gray]{0.8} 

\begin{tabular}{!{\color{black}\vrule width 1pt}>{\centering\arraybackslash}m{3.2cm}|>{\centering\arraybackslash}m{3cm}|>{\centering\arraybackslash}m{8cm}|>{\centering\arraybackslash}m{2cm}|>{\centering\arraybackslash}m{2cm}|>{\centering\arraybackslash}m{2cm}!{\color{black}\vrule width 1pt}}
\arrayrulecolor{black}\specialrule{1.0pt}{0pt}{0pt} 

\rowcolor[HTML]{808080}
{\color[HTML]{FFFFFF} \textbf{Main Improvement}} & 
{\color[HTML]{FFFFFF} \textbf{Model Name}} &
\multicolumn{1}{c}{\cellcolor[HTML]{808080}\centering {\color[HTML]{FFFFFF} \textbf{Main Methodology}}} & 
{\color[HTML]{FFFFFF} \textbf{Channel Correlation}} & 
{\color[HTML]{FFFFFF} \textbf{Enc/Dec Usage}} & 
\multicolumn{1}{c|}{\cellcolor[HTML]{808080}{\color[HTML]{FFFFFF} \textbf{Publication}}} \\
\arrayrulecolor[gray]{0.8} 

\multirow{3}{*}{Patching Technique}
 & PatchTST & \makecell[l]{· Patching \\ · Channel Independence} & CI & Enc & 2023 \\ \hhline{|~|-|-|-|-|-|}
 & MTST & \makecell[l]{· Multiple Patch-based Tokenizations} & CI & Enc & 2024 \\ \hhline{|~|-|-|-|-|-|}
 & PETformer & \makecell[l]{· Placeholder-enhanced Technique} & CI & Enc & 2022 \\ \hline

\multirow{9}{*}{Cross-Dimension} & Crossformer & \makecell[l]{· Dual Attention: Cross-time \& Cross-dimension} & CD & Enc \& Dec & 2023 \\ \hhline{|~|-|-|-|-|-|}
 & DSformer & \makecell[l]{· Dual Sampling \& Dual Attention} & CD & Enc & 2023 \\ \hhline{|~|-|-|-|-|-|}
 & CARD & \makecell[l]{· Dual Attention \\ · Token Blend Module for multi-scale} & CD & Enc & 2024 \\ \hhline{|~|-|-|-|-|-|}
 & iTransformer & \makecell[l]{· Attention on Inverted Dimension} & CD & Enc & 2024 \\ \hhline{|~|-|-|-|-|-|}
 & VCformer & \makecell[l]{· Variable Correlation Attention Considering Time Lag \\ · Koopman Temporal Detector for Non-stationarity} & CD & Enc & 2024 \\ \hhline{|~|-|-|-|-|-|}
 & GridTST & \makecell[l]{· Dual Attention with original Transformer} & CD & Enc & 2024 \\ \hhline{|~|-|-|-|-|-|}
 & UniTST & \makecell[l]{· Unified Attention by Flattening} & CD & Enc & 2024 \\ \hhline{|~|-|-|-|-|-|}
 & DeformTime & \makecell[l]{· Deformable Attention Blocks} & CD & Enc & 2024 \\ \hline

\multirow{2}{*}{Exogenous Variable} & TimeXer & \makecell[l]{· Integration of Endogenous and Exogenous Information} & CD & Enc & 2024 \\ \hhline{|~|-|-|-|-|-|}
 & TGTSF & \makecell[l]{· Exogenous Variable with Description, News} & CD & Enc \& Dec & 2024 \\ \hline

\multirow{2}{*}{Generalization} & SAMformer & \makecell[l]{· SAM (sharpness-aware minimization)} & CD & Enc & 2024 \\ \hhline{|~|-|-|-|-|-|}
 & Minusformer & \makecell[l]{· Dual-stream and Subtraction mechanism} & CD & Enc & 2024 \\ \hline

\multirow{2}{*}{Multi-scale} & Scaleformer & \makecell[l]{· Multi-scale framework} & Any & Enc \& Dec & 2023 \\ \hhline{|~|-|-|-|-|-|}
 & Pathformer & \makecell[l]{· Adaptive Multi-scale Blocks} & CD & Enc \& Dec & 2024 \\ \hline

\multirow{1}{*}{Decoder-only} & CATS & \makecell[l]{· Cross-Attention-Only Transformer} & CI & Dec & 2024 \\ \hline

\multirow{2}{*}{Feature Enhancement} & Fredformer & \makecell[l]{· Frequency Debias} & CD & Enc & 2024 \\ \hhline{|~|-|-|-|-|-|}
 & BasisFormer & \makecell[l]{· Automatic Learning of a Self-adjusting Basis} & CD & Dec & 2023 \\ 
\arrayrulecolor{black}\specialrule{1.0pt}{0pt}{0pt} 

\end{tabular}%
\end{sidewaystable}
\subsection{Growth of Fundamental Deep Learning Models}\label{sec4-2}

Since the advent of simple linear models, there has been a surge in research focused on non-transformer-based models. Attention has shifted to various architectures such as MLP, RNN, CNN, and GNN, with many models surpassing Transformers and achieving remarkable performance improvements. Although transformer-based models exhibit excellent performance across numerous fields, they have structural limitations in learning temporal order information, which is crucial for time series problems. While past tasks were simple and general enough to overlook these limitations, current real-world tasks involve many constraints and data-specific issues with diverse variables, necessitating approaches from various perspectives. Each architecture has its own strengths, and these characteristics provide valuable solutions for addressing diverse contemporary time series forecasting challenges.
In this section, we will investigate the latest major models for each architecture and analyze their technical features. The key characteristics of each backbone architecture have been briefly summarized in comparison to Transformers in Table \ref{Tab_ModelComparison}.

\medskip
\medskip

\begin{sidewaystable}
\centering
\footnotesize 

\captionsetup{justification=raggedright, singlelinecheck=false} 
\captionof{table}{Comparison of Other Deep Learning Models with Transformers in Terms of Criteria}
\label{Tab_ModelComparison}

\renewcommand{\arraystretch}{1.6} 
\arrayrulecolor[gray]{0.8} 



\begin{tabular}{!{\color{black}\vrule width 1pt} >{\columncolor[HTML]{E4E4E4}}>{\centering\arraybackslash}m{0.08\textheight}|>{\columncolor[HTML]{F2F2F2}}>{\centering\arraybackslash}m{0.16\textheight}|>{\centering\arraybackslash}m{0.16\textheight}|>{\centering\arraybackslash}m{0.16\textheight}|>{\centering\arraybackslash}m{0.16\textheight}|>{\centering\arraybackslash}m{0.16\textheight}!{\color{black}\vrule width 1pt}}

\arrayrulecolor{black}\specialrule{1.0pt}{0pt}{0pt} 

\rowcolor[HTML]{808080} 
{\color[HTML]{FFFFFF}\textbf{Criteria}} & 
{\color[HTML]{FFFFFF}\textbf{Transformer-based models}} & 
{\color[HTML]{FFFFFF}\textbf{MLP-based models}} & 
{\color[HTML]{FFFFFF}\textbf{CNN-based models}} & 
{\color[HTML]{FFFFFF}\textbf{RNN-based models}} & 
{\color[HTML]{FFFFFF}\textbf{GNN/GCN-based models}} \\

\arrayrulecolor[gray]{0.8}\hline

{\raggedright Structure} & Complex self-attention mechanism & Simple layer, relatively easy to implement and interpret & Utilizes convolutional layers, effectively capturing specific local patterns & Specialized in sequential data processing, effectively handling temporal dependencies but may struggle with long sequences & Learns relationships between nodes using graph structures, effectively capturing complex relationships \\ \hline
{\raggedright Data Requirements} & Requires large datasets & Can train on smaller datasets & Can train on smaller datasets & Can train on smaller datasets, suitable for quick training with limited data & Can achieve high performance with relatively small datasets \\ \hline
{\raggedright Training Time} & Relatively slow due to global attention mechanisms & Relatively fast & Generally faster due to localized computations & Trains effectively on smaller datasets but can be slow for long sequences & Varies depending on graph complexity \\ \hline
{\raggedright Model Size} & Comparatively larger and more parameter-intensive & Comparatively small, efficient use of resources & Typically smaller and more parameter-efficient, making it resource-efficient and scalable & Comparatively small & Depends on graph size and complexity, achieving high performance with fewer parameters in specific problems \\ \hline
{\raggedright Interpretability} & Difficult to interpret & Relatively high interpretability & More interpretable through visualizations of filters and feature maps & Moderately interpretable, easier to understand and explain model behavior & Moderately interpretable depending on graph structure and model complexity \\ \hline
{\raggedright Performance} & Suitable for learning long-term dependencies & Suitable for short-term predictions with sufficient performance in many cases & Excels at capturing local temporal dependencies, superior for problems with strong local patterns & Suitable for short-term and sequential dependencies, providing sufficient performance in specific time series problems & Excels at learning complex dependencies between nodes, offering high performance in learning specific relationship patterns \\ \hline
{\raggedright Flexibility} & Requires complex adjustments & Easily adjustable for specific problems & Easily customizable with various types of convolutions & Extensible with various RNN architectures & Can handle various graph structures and formats, adaptable to different data types and structures \\ \hline
{\raggedright Application} & Suitable for complex time series problems or NLP-related tasks & Versatile for various general forecasting problems & Well-suited for applications requiring spatial and temporal locality, effective for a wide range of time series problems & Effective for sequential data and time series forecasting, but struggles with long-term dependencies without modifications & Suitable for complex graph structures in tasks like social networks, recommendation systems, and time series graphs \\ \hline
{\raggedright Hardware Requirements} & High due to their complex structure and computationally intensive self-attention mechanisms & Lower due to their simpler structure and fewer computational demands & Lower computational and memory requirements & Low but inefficient on parallel hardware & Generally low but depends on graph size \\ \hline
{\raggedright Memory Usage} & Higher memory usage due to full sequence attention & Lower memory usage due to their simple structure and fewer parameters & Lower memory usage due to localized operations & Low but can increase with sequence length & Generally low but depends on graph size \\ \hline
{\raggedright Parallel Processing} & Highly parallelizable but requires synchronization due to attention mechanisms & Highly parallelizable due to independent computations & Highly parallelizable due to independent convolution operations & Difficult due to sequence dependencies & Difficult due to graph structure dependencies \\ 

\arrayrulecolor{black}\specialrule{1.0pt}{0pt}{0pt} 

\end{tabular}
\smallskip

\end{sidewaystable}

\subsubsection{MLP-Based Models}
MLP-based models have recently emerged as a key methodology for replacing Transformers in time series forecasting tasks. The simple structure of MLPs makes them easy to handle and interpret. Additionally, they perform well even in constrained environments with limited computational resources and limited data. Their lightweight architectures enable fast training and inference, making them increasingly important and widely used in contemporary industries. Previously, interest in MLPs diminished due to their structural limitations, such as the lack of sequential dependency, challenges with long-term dependency, difficulties in processing high-dimensional data, and limitations in capturing periodic patterns. However, recent advancements have enabled long-sequence learning and various technical enhancements, leading to remarkable performance improvements.

\revision{\textbf{N-BEATS} \citep{N-BEATS}, consisting of simple Fully Connected (FC) layers, demonstrated superior performance compared to traditional statistical models. This model utilizes a block-based architecture with repeatedly stacked backcast and forecast paths and enhances generalization performance through ensemble learning. By applying decomposition, it separates trend and seasonality, modeling them using Polynomial Basis and Fourier Basis, respectively, making it highly interpretable. 
\revisionB{Moreover, recent studies have improved the N-BEATS model to predict various quantiles, enabling probabilistic forecasting \citep{smyl2024any}.}
The improved \textbf{N-HiTS} \citep{N-HITS} model can handle multivariate data and overcomes the limitations of N-BEATS by employing techniques such as multi-rate signal sampling, non-linear regression, and hierarchical interpolation.}
\textbf{Koopa} \citep{Koopa} is a model designed to effectively handle non-stationary time series data by utilizing Koopman theory to predict the dynamic changes of components. It uses a Fourier filter to separate time-invariant and time-variant elements, which are then fed into respective Koopman predictors(KP). In KP, nonlinear time series data are linearly transformed, making them easier to manage. To effectively capture the characteristics of each component, the time-invariant KP uses a globally learned Koopman operator, while the time-variant KP computes a local operator for predictions.
\textbf{TSMixer} \citep{TSMixer} is a lightweight patch-based model that introduces the MLP-mixer from the computer vision field to the time-series domain. It efficiently learns long-term and short-term dependencies through inter-patch and intra-patch mixers respectively, and explicitly learns channel relationships through an inter-channel mixer. To enhance the understanding of inter-channel correlations, it adjusts the original forecast results using reconciliation heads and employs gated attention to filter important features. By introducing these techniques, TSMixer upgrades the simple MLP structure, resulting in improved performance that outperforms complex Transformer models.
\textbf{FreTS} \citep{FreTS} is a model that leverages two key characteristics of the frequency domain—global view and energy compaction—to directly train an MLP model in the frequency domain. While existing models primarily use frequency transformation to verify the periodicity of the model, this model distinguishes between channel and temporal dependencies in the frequency domain as real and imaginary parts, respectively, and learns them directly to better extract hidden features.
\textbf{TSP} \citep{TSP} utilizes a PrecMLP block with a precurrent mechanism to effectively model temporal locality and channel dependency. The precurrent mechanism combines previous information with current information to create hybrid features at each time point, serving as a computationally-free method to effectively recognize and process temporal locality. This lightweight MLP model comprises an encoder, consisting of tMLP operating in the time dimension and vMLP operating in the variable dimension, and a decoder using a linear model.
\textbf{FITS} \citep{FITS} is a very lightweight model with fewer than 10,000 parameters, yet it demonstrates competitive performance comparable to larger models. It performs interpolation through a complex linear layer that learns amplitude scaling and phase shifts in the frequency domain, thereby expanding the frequency representation of the input time series. By using a low-pass filter to remove high-frequency components, which are often noise, the model efficiently represents the data and reduces the number of learnable parameters.
\textbf{U-Mixer} \citep{U-Mixer} utilizes a hierarchical structure of U-Net's encoder-decoder composed of MLPs to extract and combine low-dimensional and high-dimensional features. Each MLP block processes temporal dependencies and channel interactions separately, enabling stable feature extraction. Additionally, the model employs a stationary correction method to limit the differences in data distribution before and after processing, thereby restoring the non-stationary information of the data. By calculating the transformation matrix and mean difference between the input and output data, the model adjusts the output to maintain temporal dependency, thereby enhancing its prediction performance.
\textbf{TTMs} \citep{TTMs} is a time series foundation model that uses TSMixer as its backbone and relies solely on time series data for rapid pre-training. This model is trained on the extensive collection of time series datasets with diverse channels and resolutions from the Monash archive (Table \ref{tab:dataset_summary}). Despite the data-specific characteristics of the time series datasets, TTMs demonstrate efficient transfer learning and high generalization performance.
\textbf{TimeMixer} \citep{TimeMixer} leverages the observation that time series data exhibit distinct patterns at different sampling scales. To extract important information from past data, multi-scale time series are generated through downsampling. In the Past-Decomposable-Mixing (PDM) block, the seasonal and trend components are decomposed and mixed separately. The Future-Multipredictor-Mixing (FMM) block then integrates predictions by ensembling multiple predictors, which utilize the past information extracted at various scales.
\textbf{CATS} \citep{CATS} enhances the performance of multivariate time series forecasting by generating Auxiliary Time Series (ATS) from the Original Time Series (OTS) and integrating them into the prediction process. The model proposes various types of ATS constructors, each with different roles and objectives, combining them to maximize predictive performance. To effectively generate and utilize ATS, the model structure and loss function incorporate three key principles: continuity, sparsity, and variability. Despite using a simple two-layer MLP predictor as the foundation, CATS effectively predicts multivariate information through the use of ATS.
\textbf{HDMixer} \citep{HDMixer} is a model that applies a Length-Extendable Patcher to overcome the limitations of fixed-length patching. It measures the point-wise similarity with fixed patches and compares the patch entropy loss, updating parameters to increase the complexity of the extendable patches. Through the Hierarchical Dependency Explorer mixer, it learns long-term, short-term dependencies and cross-variable dependencies.
\textbf{SOFTS} \citep{SOFTS} proposes the STar Aggregate-Redistribute (STAR) Module to effectively learn the relationships between channels. It converts the data from each channel into high-dimensional vectors and then aggregates these vectors to extract comprehensive information, referred to as the core. The generated core representation is then combined with the time series representation of each channel and converted back to the time series dimension, reducing complexity and enhancing robustness.
\textbf{SparseTSF} \citep{SparseTSF} is an extremely lightweight model that uses fewer than 1,000 parameters while demonstrating excellent generalization performance. It downsamples the data into multiple periods, performs predictions on each generated sub-sequence, and then upsamples to create the overall prediction sequence. While it shows limited performance only on data with clear periodic patterns, it efficiently handles complex temporal dependencies with minimal computational resources.
\textbf{TEFN} \citep{TEFN} proposes a novel backbone model from the perspective of information fusion to extract latent distributions from simple data structures. Based on evidence theory \citep{evidencetheory}, the Basic Probability Assignment Module maps each time and channel dimension of the time series data to a mass function in the event space. This mass function assigns probabilities to multiple possibilities for each data point, allowing for explicit modeling of uncertainty. The generated mass functions are fused using their expected values to derive the final predictions. This model achieves high predictive performance with few parameters and fast training times.
\textbf{PDMLP} \citep{PDMLP} surpasses the performance of complex transformer-based models using only simple linear layers and patching. It divides the data into patches of various scales and embeds each patch with a single linear layer. The embedding vectors are then decomposed into smooth components and residual components containing noise, allowing the model to process the sequences in two ways: intra-variable and inter-variable.
\textbf{AMD} \citep{AMD} introduces the AMS block, which assesses the importance of various temporal patterns and generates appropriate weights for each pattern. By simultaneously modeling the temporal and channel dependencies of the input information decomposed into multiple scales, it effectively captures interactions at different scales. Instead of merely integrating the results from each scale, it improves prediction performance by reflecting the importance of each pattern through the weights.

\subsubsection{CNN-Based Models}
CNNs were primarily developed for image recognition and processing tasks because they are highly effective at identifying 2D spatial patterns. Some studies have begun utilizing Temporal Convolutional Networks (TCNs), which are 1D CNNs that move along the time axis to recognize local patterns, effectively extracting key features of time series data. Additionally, CNNs can use filters of various sizes to capture patterns at multiple scales, allowing them to effectively learn both short-term and long-term patterns in time series data.
CNN's deep network technology and parallelism, which have been developed over a long period of time, have provided high performance and reliability for time series forecasting. However, the fixed filter size of CNNs lacks flexibility in learning complex patterns of long sequences and poses challenges in capturing long-term dependencies. Furthermore, as the network depth increases, there are limitations such as information loss, increased computational cost, and higher memory usage. For these reasons, the relatively flexible Transformer, which is advantageous for learning long-term dependencies, has gained more attention in time series data processing. However, due to their superior ability to capture diverse local patterns within long sequences, CNNs are once again receiving attention as a backbone architecture.

\revision{Recently, 3D convolution methodologies have been actively studied for handling complex spatio-temporal time series data \citep{chen2021multiple, feng2024learning}. 3D convolution processes input data in three dimensions (2D spatial axes + temporal axis) using convolution operations. This approach enables the extraction of richer patterns by processing not only the temporal axis but also spatial or multi-dimensional aspects of the data. Such methods have demonstrated high performance in region-based prediction tasks, and future research is expected to focus on developing more sophisticated data representations and learning structures.}

\textbf{TimesNet} \citep{Timesnet} introduces the TimesBlock, which embeds 1D time series into 2D tensors for analysis to effectively capture multi-periodicity. Extracting key frequencies via Fast Fourier Transformation (FFT) expands the data into a 2D tensor, representing intra-period variation in columns and inter-period variation in rows. It uses a parameter-efficient Inception module with 2D kernels to capture temporal variations across various scales. After converting back to a 1D tensor, it adaptively aggregates the data based on the importance of frequency amplitude values.
\textbf{PatchMixer} \citep{PatchMixer} introduces a patch-mixing design to efficiently replace the permutation invariant property of the Transformer's attention mechanism. It learns intra-patch correlations using single-scale depth-wise separable convolutions and inter-patch correlations using point-wise convolutions on the patched input sequence.
\textbf{ModernTCN} \citep{ModernTCN} enhances the fundamental TCNs to make them more suitable for time series analysis. Using larger kernels than conventional TCNs expands the effective receptive field, reducing the number of layers and effectively capturing long-term dependencies. It separates depth-wise convolution and point-wise convolution to independently learn temporal dependencies and feature dependencies, thereby enhancing computational efficiency. Additionally, it explicitly handles cross-variable dependency in multivariate time series data, resulting in improved performance.
\textbf{ConvTimeNet} \citep{ConvTimeNet} follows the framework of a Transformer encoder, replacing the role of self-attention with depthwise convolution and the role of multi-head attention with convolutions of various kernel sizes to increase computational efficiency. It uses Deformable Patch Embedding, which adaptively adjusts each patch's size and position based on the data. Depthwise convolutions capture temporal features for each channel, while point-wise convolutions model channel dependencies.
\textbf{ACNet} \citep{ACNet} aims to effectively model temporal dependencies and nonlinear features. It starts by removing unnecessary noise from the data using Wavelet Denoising. Then, it extracts temporal features through Multi-Resolution Dilated Convolution and Global Adaptive Average Pooling. To more accurately capture nonlinear features, Gated Deformable Convolution is employed, which adaptively adjusts the sampling positions. Additionally, ACNet employs dynamic prediction, retraining the model with new data if prediction performance declines.
\textbf{FTMixer} \citep{FTMixer} leverages the observation that the time domain effectively captures local dependencies while the frequency domain excels at learning global dependencies. It applies convolution in the frequency domain to capture global dependencies. Meanwhile, it converts multi-scale patches using Discrete Cosine Transformation (DCT), which is computationally simpler than Discrete Fourier Transformation (DFT), and then applies convolution to capture local dependencies.

\subsubsection{RNN-Based Models}
RNNs were fundamentally developed to process sequential data. Their recurrent structure can effectively model the temporal dependencies of time series data. Gated RNN models, such as LSTM and GRU, have partially alleviated the long-term dependency issues of vanilla RNNs. However, RNNs are difficult to parallelize because they process data sequentially. This slow speed is particularly detrimental when handling long sequences. RNNs also have inherent limitations in learning long-term dependencies and global information. Transformers have effectively addressed these issues through self-attention mechanisms. However, unlike Transformers, which rely on large datasets, RNN-based models can learn effectively with smaller amounts of data. They also use memory efficiently and are well-suited for real-time data processing due to their sequential nature, making them advantageous for online streaming applications. Additionally, because time ordering is naturally learned, RNNs still have inherent advantages over Transformers for time series processing.

\textbf{PA-RNN} \citep{PA-RNN} aims to address prediction Uncertainty Quantification and Model Compression issues for dependent time series data by proposing an optimization methodology for Sparse Deep Learning. It applies a Mixture of Gaussian Prior to introducing sparsity to the parameters and uses a Prior Annealing technique to gradually increase the strength of the prior distribution during the training process. Properties such as Posterior Consistency, Structure Selection Consistency, and Asymptotic Normality ensure the model's theoretical validity.
\textbf{WITRAN} \citep{WITRAN} proposes a new paradigm by implementing information transmission in bi-granular flows to capture long-term and short-term repetitive patterns. It models global and local correlations through the Horizontal Vertical Gated Selective Unit (HVGSU), which uses bidirectional Gated Selective Cells (GSC). To enhance the efficiency of information transmission through parallel processing, it applies the Recurrent Acceleration Network (RAN). This approach results in excellent prediction performance while reducing time complexity and maintaining memory complexity.
\textbf{SutraNets} \citep{SutraNets} transforms long sequences into multiple lower-frequency sub-series in order to effectively predict long sequence time series data. It performs autoregressive predictions among the sub-series, with each sub-series being conditionally predicted based on the values of other sub-series. This approach reduces the signal path length and allows for the generation of long sequences in fewer steps. Consequently, it mitigates the common RNN problem of error accumulation and models long-distance dependencies better.
\textbf{CrossWaveNet} \citep{CrossWaveNet} uses a dual-channel network to separate and process the seasonality and trend-cyclic components of data. The input data are initially decomposed into individual elements through series decomposition, and features are extracted using enriched LSTM and GRU structures. These extracted features undergo a cross-decomposition process to further refine the elements, which are then sent to each channel for integration. Series decomposition is performed at each RNN step, progressively filtering and reintegrating the seasonality and trend components to learn precise information.
\textbf{DAN} \citep{DAN} proposes a new model that utilizes Polar Representation learning to predict long-term time series data with high volatility. The Distance-weighted Auto-regularized Neural network (DAN) distinguishes and learns the polar representations of distant and nearby data, combining them to enhance prediction performance. It addresses the imbalance problem of extreme data, which occur infrequently but are critical for accurate predictions, using Kruskal-Wallis Sampling and employs a Multi-Loss Regularization Method to effectively learn both extreme and normal values simultaneously.
\textbf{RWKV-TS} \citep{RWKV-TS} proposes an efficient Recurrent Weighted Key-Value (RWKV) backbone characterized by linear time complexity and memory usage. It uses a Multi-head Weighted Key-Value (WKV) Operator, which is similar to the self-attention mechanism but maintains linear time and space complexity. Each head captures different temporal patterns, increasing the representational power of the information and enabling parallel computation, thus providing high computational efficiency. This allows RWKV-TS to capture longer sequence information more effectively than fundamental RNNs.
\textbf{CONTIME} \citep{CONTIME} introduces Continuous GRU to minimize the prediction delay issue in time series forecasting. This approach applies Neural ODE to the existing GRU, modeling data changes continuously over time. The bi-directional structure integrates forward and backward through the time series data, capturing more accurate temporal dependencies. By employing Time-Derivative Regularization, the model is guided to learn the rate of change in predictions over time, enhancing both the accuracy and speed of predictions.

\subsubsection{GNN-Based Models}
GNNs or GCNs are specialized in processing graph-structured data, being suitable for modeling complex interactions and learning local patterns in time series data. Multivariate time series data often involve interactions between variables, which can be effectively represented as relationships between nodes and edges. By assigning various attributes to nodes and edges, richer representations can be learned, and the structural characteristics of specific domains can be better reflected. Traditionally, various approaches such as GCN, ST-GCN, GAT, and TGN have been proposed, contributing to the advancement of the time series forecasting field. 

However, these models primarily focus on learning local information from adjacent nodes, leading to a lack of capability in learning global information and difficulty in capturing long-term dependencies with distant nodes. The sequential calculation of nodes and edges makes parallel processing challenging, and the models' specificity to particular graph structures reduces their generalization ability. These drawbacks have diminished interest in GNN-based models. However, with the recent advancements in social networks, much of current data naturally follows a graph structure. Various optimization techniques and hardware acceleration technologies have been developed, making it possible to process large-scale graph data. Additionally, the increased practical relevance due to the flexibility in dynamically handling time-varying data has once again drawn researchers' attention to this architecture.

\textbf{MSGNet} \citep{MSGNet} effectively learns the complex correlations of MTS data by combining frequency domain analysis with adaptive graph convolution. It uses Fast Fourier Transformation to identify key periods and transform time series data into various time scales. Then, it employs Mixhop graph convolution to learn inter-channel correlations and uses Multi-head Attention and Scale Aggregation to capture intra-series correlations.
\textbf{TMP-Nets} \citep{TMP-Nets} is a model proposed to learn features extracted from Temporal MultiPersistence (TMP) vectorization and capture key topological patterns. It combines two advanced topological data analysis techniques, Multipersistence and Zigzag Persistence, to effectively capture topological shape changes in the data. Spatial Graph Convolution is used to model dependencies between nodes in the graph, and the topological features extracted with TMP vectors are learned using CNN.
\textbf{HD-TTS} \citep{HD-TTS} is a model that demonstrates high performance even with missing data through masking and hierarchical downsampling. It incrementally reduces data in the temporal and spatial processing modules, learning various temporal and spatial dependencies. The model uses an attention mechanism to learn the patterns of missing values and combines them with weights to generate the final prediction.
\textbf{ForecastGrapher} \citep{ForecastGrapher} is based on the finding that the attention mechanism is more suitable for modeling inter-channel correlations than temporal correlations. Utilizing the similarity between GNN's neighborhood aggregation and attention mechanisms, it linearly embeds each channel as a node and learns the relationships between channels using GNN. Each GNN layer employs a self-learnable adjacency matrix to independently learn the interactions between nodes. It uses learnable scalers to divide node features into multiple groups and applies 1D convolution to each group.

\subsubsection{Model-Agnostic Frameworks}
Some studies propose model-agnostic methodologies that can be universally applied without being limited to a specific model backbone. These studies focus on the intrinsic characteristics of time series data or specific problems, improving model robustness by addressing these issues. Model-agnostic architectures, not being tied to a particular model, allow for flexible selection of the optimal model through various comparisons. They are also highly reusable and make it easier to interpret and understand the model's predictions, making them applicable to a wide range of domains and scenarios.

These models primarily address the distribution shift problem, which is one of the core issues in time series forecasting, and propose various plug-and-play alleviation methods. The details related to this will be discussed in Section 5, while here, we will focus on other methodologies, excluding distribution shift alleviation methods.

\textbf{RobustTSF} \citep{RobustTSF} introduces a selective sampling algorithm to train a resilient prediction model in Time Series Forecasting with Anomalies (TSFA). It calculates anomaly scores based on the difference between the trend and the actual data, selecting samples with scores below a certain threshold to train the model, thereby minimizing the impact of anomalies. By analyzing three types of anomalies (Constant, Missing, Gaussian), it identifies the most robust loss function as MAE. This approach addresses the discontinuity issues of the traditional detection-imputation-retraining pipeline, enabling the model to deliver more stable and consistent performance.
\textbf{PDLS} \citep{PDLS} proposes a constrained learning approach by setting an upper bound on the loss at each time step. It controls the loss by setting a constant upper bound (\(\epsilon\)) for all time steps and introduces a variable (\(\zeta\)) that relaxes the constraints, allowing automatic adjustment during training. This model overcomes the problem of existing methodologies that focus on averaging prediction performance, which often leads to uneven error distributions.
\textbf{Leddam} \citep{Leddam} proposes two independently usable modules, Learnable Decomposition and Dual Attention, to effectively capture the complex patterns of inter-series dependencies and intra-series variations in MTS. The Learnable Decomposition module replaces the traditional moving average kernel with a learnable 1D convolution kernel initialized with a Gaussian distribution for decomposition, making it more sensitive to critical information. The Dual Attention module simultaneously models inter-data dependencies and intra-series variability using channel-wise self-attention and auto-regressive self-attention.
\textbf{InfoTime} \citep{InfoTime} proposes a method to effectively utilize inter-channel and temporal correlations. For channel mixing, it employs Cross-variable Decorrelation Aware feature Modeling (CDAM) to minimize redundant information between channels and enhance useful mutual information. Additionally, it uses TAM(Temporal correlation Aware Modeling) to maximize the temporal correlations between the predicted future sequence and the target sequence.
\textbf{CCM} \citep{CCM} combines the strengths of the traditional channel-independent and channel-dependent strategies to overcome their limitations. It dynamically clusters channels based on their inherent similarities and learns representative patterns for each cluster to generate prototype embeddings. Separate FFNs are used for each cluster to perform predictions for the channels within the cluster, and the prototype embeddings enable zero-shot predictions.
\textbf{HCAN} \citep{HCAN} addresses the issue of excessive flattening caused by using the MSE loss function and proposes reconfiguring the prediction problem as a classification problem to better learn high-entropy features. It extracts important features from the backbone model, divides them into multiple layers and categories, and uses an Uncertainty-Aware Classifier at each layer to reduce uncertainty. It learns relative prediction values within each classified category and integrates the prediction results from multiple layers through the Hierarchy-Aware Attention module. This approach effectively captures the complex patterns and variability in time series data.
\textbf{TDT Loss} \citep{TDTLoss} points out the error accumulation issue in the traditional auto-regressive approach which recursively models Temporal Dependencies among Targets (TDT). To effectively capture changes between consecutive time points, it represents TDT using first-order differencing. The final TDT loss combines the target prediction loss, TDT values prediction loss, and an adaptive weight, allowing the model to dynamically balance between target prediction and TDT learning. Thus, TDT loss replaces the optimization objective of non-autoregressive models.

The different deep learning models and their methodologies discussed so far are summarized in Table \ref{tab:DLModelOverview}.

\begin{table}[!h]
\centering

\captionsetup{justification=raggedright,singlelinecheck=false}
\caption{Taxonomy and Methodologies of Fundamental Deep Learning Architectures for Time Series Forecasting}
\label{tab:DLModelOverview}

\tiny 
\renewcommand{\arraystretch}{2.1} 
\arrayrulecolor[gray]{0.8} 

\begin{tabular}{!{\color{black}\vrule width 1pt}>{\centering\arraybackslash}m{1.6cm}|>{\centering\arraybackslash}m{1.6cm}|>{\raggedright\arraybackslash}m{6.7cm}|>{\centering\arraybackslash}m{1.0cm}|>{\centering\arraybackslash}m{1.3cm}!{\color{black}\vrule width 1pt}}
\arrayrulecolor{black}\specialrule{1.0pt}{0pt}{0pt} 

\rowcolor[HTML]{808080}
{\color[HTML]{FFFFFF} \textbf{Model Name}} & 
{\color[HTML]{FFFFFF} \textbf{Base}} &
\multicolumn{1}{c}{\cellcolor[HTML]{808080}\raggedright {\color[HTML]{FFFFFF} \textbf{Main Methodology}}} & 
{\color[HTML]{FFFFFF} \textbf{Channel Correlation}} & 
{\color[HTML]{FFFFFF} \textbf{Publication}} \\
\arrayrulecolor[gray]{0.8} 

\revision{N-BEATS} & \multirow{20}{*}{MLP} & \makecell[l]{\revision{· Neural Basis Expansion (Polynomial Basis \& Fourier Basis)}} & \revision{CI} & \revision{2020} \\ \hhline{|-|~|-|-|-|}
\revision{N-HiTS} &  & \makecell[l]{\revision{· Neural Hierarchical Interpolation} \\ \revision{· Multi-Rate Data Sampling}} & \revision{CI} & \revision{2023} \\ \hhline{|-|~|-|-|-|}
Koopa &  & \makecell[l]{· Koopa Block with Koopman Predictor(KP)} & CD & 2023 \\ \hhline{|-|~|-|-|-|}
TSMixer &  & \makecell[l]{· MLP Mixer layer architecture \\ · Gated attention (GA) block \\ · Online hierarchical patch reconciliation head} & CI/CD & 2023 \\ \hhline{|-|~|-|-|-|}
FreTS &  & \makecell[l]{· Frequency-domain MLPs} & CD & 2023 \\ \hhline{|-|~|-|-|-|}
TSP &  & \makecell[l]{· PrecMLP block with precurrent mechanism} & CD & 2024 \\ \hhline{|-|~|-|-|-|}
FITS &  & \makecell[l]{· Complex Frequency Linear Interpolation \\ · Low Pass Filter(LPF)} & CI & 2024 \\ \hhline{|-|~|-|-|-|}
U-Mixer &  & \makecell[l]{· Unet Encoder-decoder with MLPs \\ · Stationarity Correction} & CD & 2024 \\ \hhline{|-|~|-|-|-|}
TTMs &  & \makecell[l]{· Multi-Resolution Pre-training via TTM Backbone\\ \ \ (TSMixer blocks) \\ · Exogenous mixer} & CD & 2024 \\ \hhline{|-|~|-|-|-|}
TimeMixer &  & \makecell[l]{· Past-Decomposable-Mixing (PDM) block \\ · Future-Multipredictor-Mixing (FMM) block} & CD & 2024 \\ \hhline{|-|~|-|-|-|}
CATS &  & \makecell[l]{· Auxiliary Time Series(ATS) Constructor} & CD & 2024 \\ \hhline{|-|~|-|-|-|}
HDMixer &  & \makecell[l]{· Length-Extendable Patcher \\ · Patch Entropy Loss \\ · Hierarchical Dependency Explorer} & CD & 2024 \\ \hhline{|-|~|-|-|-|}
SOFTS &  & \makecell[l]{· STar Aggregate-Redistribute Module} & CD & 2024 \\ \hhline{|-|~|-|-|-|}
SparseTSF &  & \makecell[l]{· Cross-Period Sparse Forecasting} & CI & 2024 \\ \hhline{|-|~|-|-|-|}
TEFN &  & \makecell[l]{· Basic Probability Assignment(BPA) Module} & CD & 2024 \\ \hhline{|-|~|-|-|-|}
PDMLP &  & \makecell[l]{· Multi-Scale Patch Embedding \& Feature Decomposition \\ · Intra-, Inter-Variable MLP} & CD & 2024 \\ \hhline{|-|~|-|-|-|}
AMD &  & \makecell[l]{· Multi-Scale Decomposable Mixing(MDM) Block \\ · Dual Dependency Interaction(DDI) Block \\ · Adaptive Multi-predictor Synthesis(AMS) Block} & CD & 2024 \\ \hhline{|-|-|-|-|-|}

\arrayrulecolor{black}\specialrule{1.0pt}{0pt}{0pt} 
\arrayrulecolor[gray]{0.8} 

TimesNet & \multirow{7}{*}{\centering CNN} & \makecell[l]{· Transform 1D-variatios into 2D-variations \\ · Timesblock} & CD & 2023 \\ \hhline{|-|~|-|-|-|}
PatchMixer &  & \makecell[l]{· Patch Embedding \\ · PatchMixer layer with Patch-mixing Design} & CI & 2023 \\ \hhline{|-|~|-|-|-|}
ModernTCN &  & \makecell[l]{· ModernTCN block with DWConv \& ConvFFN} & CD & 2024 \\ \hhline{|-|~|-|-|-|}
ConvTimeNet &  & \makecell[l]{· Deformable Patch Embedding \\ · Fully Convolutional Blocks} & CD & 2024 \\ \hhline{|-|~|-|-|-|}
ACNet &  & \makecell[l]{· Temporal Feature Extraction Module \\ · Nonlinear Feature Adaptive Extraction Module} & CD & 2024 \\ \hhline{|-|~|-|-|-|}
FTMixer &  & \makecell[l]{· Frequency Channel Convolution \\ · Windowing Frequency Convolution} & CD & 2024 \\ \hhline{|-|-|-|-|-|}

\arrayrulecolor{black}\specialrule{1.0pt}{0pt}{0pt} 
\arrayrulecolor[gray]{0.8} 

PA-RNN & \multirow{8}{*}{\centering RNN} & \makecell[l]{· Mixture Gaussian Prior \\ · Prior Annealing Algorithm} & CI & 2023 \\ \hhline{|-|~|-|-|-|}
WITRAN &  & \makecell[l]{· Horizontal Vertical Gated Selective Unit \\ · Recurrent Acceleration Network} & CI & 2023 \\ \hhline{|-|~|-|-|-|}
SutraNets &  & \makecell[l]{· Sub-series autoregressive networks} & CI & 2023 \\ \hhline{|-|~|-|-|-|}
CrossWaveNet &  & \makecell[l]{· Deep cross-decomposition \\ · Dual-channel network} & CD & 2024 \\ \hhline{|-|~|-|-|-|}
DAN &  & \makecell[l]{· RepGen \& RepMerg with Polar Representation Learning \\ · Distance-weighted Multi-loss Mechanism \\ · Kruskal-Wallis Sampling} & CI & 2024 \\ \hhline{|-|~|-|-|-|}
RWKV-TS &  & \makecell[l]{· RWKV Blocks with Multi-head WKV Operator} & CD & 2024 \\ \hhline{|-|~|-|-|-|}
CONTIME &  & \makecell[l]{· Bi-directional Continuous GRU with Neural ODE} & CI & 2024 \\ \hhline{|-|-|-|-|-|}

\arrayrulecolor{black}\specialrule{1.0pt}{0pt}{0pt} 
\arrayrulecolor[gray]{0.8} 

MSGNet & \multirow{5}{*}{\centering GNN/GCN} & \makecell[l]{· ScaleGraph block with Scale Identification \\· Multi-scale Adaptive Graph Convolution. \\ · Multi-head Attention and Scale Aggregation} & CD & 2024 \\ \hhline{|-|~|-|-|-|}
TMP-Nets &  & \makecell[l]{· Temporal MultiPersistence(TMP) Vectorization} & CD & 2024 \\ \hhline{|-|~|-|-|-|}
HD-TTS &  & \makecell[l]{· Temporal processing module with Temporal message passing \\ · Spatial processing module with Spatial message passing} & CD & 2024 \\ \hhline{|-|~|-|-|-|}
ForecastGrapher &  & \makecell[l]{· Group Feature Convolution GNN (GFC-GNN)} & CD & 2024 \\ \hhline{|-|-|-|-|-|}

\arrayrulecolor{black}\specialrule{1.0pt}{0pt}{0pt} 
\arrayrulecolor[gray]{0.8} 



RobustTSF & \multirow{9}{*}{\makecell[c]{Model-agnostic}} & \makecell[l]{· RobustTSF Algorithm} & - & 2024 \\ \hhline{|-|~|-|-|-|}
PDLS &  & \makecell[l]{· Loss Shaping Constraints \\ · Empirical Dual Resilient and Constrained Learning} & - & 2024 \\ \hhline{|-|~|-|-|-|}
Leddam &  & \makecell[l]{· Learnable Decomposition Module \\ · Dual Attention Module} & CD & 2024 \\ \hhline{|-|~|-|-|-|}
InfoTime &  & \makecell[l]{· Cross-Variable Decorrelation Aware Feature Modeling \\ \ \ (CDAM) \\ · Temporal Aware Modeling (TAM)} & CD & 2024 \\ \hhline{|-|~|-|-|-|}
CCM &  & \makecell[l]{· Channel Clustering \& Cluster Loss \\ · Cluster-aware Feed Forward} & CD & 2024 \\ \hhline{|-|~|-|-|-|}
HCAN &  & \makecell[l]{· Uncertainty-Aware Classifier(UAC) \\ · Hierarchical Consistency Loss(HCL) \\ · Hierarchy-Aware Attention(HAA)} & - & 2024 \\ \hhline{|-|~|-|-|-|}
TDT Loss &  & \makecell[l]{· Temporal Dependencies among Targets(TDT) Loss} & - & 2024 \\ 
\arrayrulecolor{black}\specialrule{1.0pt}{0pt}{0pt} 

\end{tabular}%
\captionsetup{justification=raggedright, singlelinecheck=false}
\caption*{\raggedright \footnotesize  - : Indicates that the feature is model-agnostic and depends on which backbone model is applied.}


\end{table}

\subsection{Emergence of Foundation Models}\label{sec4-3}

In recent years, foundation models have demonstrated remarkable performance and strong zero-shot capabilities across various tasks and domains, leading to increased focus~\citep{Llama2, Llava, BLIP2, GPT4, gallifant2024peer, CLIP}. While significant progress has been made in domains like vision and language, developing foundation models for time series data has faced several challenges. Firstly, time series data exhibit distinct characteristics depending on the dataset. For instance, electrocardiogram (ECG), weather, and sensor data from industrial processes have unique properties in terms of variables, frequency, periodicity, and noise, often requiring domain-specific knowledge for effective modeling. 
Additionally, unlike the vision and language domains, where large-scale pre-training corpora can be relatively easily constructed from publicly available sources like the web, collecting time series data is more difficult due to high acquisition costs and security concerns. Despite these obstacles, research on time series foundation models is essential for improving model scalability and generality.
This necessity has led to active exploration in several key directions, particularly in the field of time series forecasting.

\subsubsection{Sequence Modeling with LLMs}
One prominent approach leverages the sequence modeling capabilities of LLMs. LLMs have revolutionized deep learning with their groundbreaking generalizable sequence modeling. Given the sequential nature of both text and time series data, extending the sequential capabilities of language models to time series is a natural progression.
Early research includes \textbf{GPT4TS}~\citep{OneFitsAll}, which demonstrates that by freezing most of the language model’s backbone and fine-tuning only the layer normalization parameters and positional embeddings on time series data, GPT4TS can serve as a general time series task solver for forecasting, anomaly detection, and classification.
Additionally, \textbf{PromptCast}~\citep{Promptcast} introduces a framework where numerical time series sequences are input as text prompts to a large language model. Then the model outputs the forecasting results in a question-answering format. This approach integrates domain-specific knowledge through text and provides forecasting results in a human-friendly format.

Both GPT4TS and PromptCast rely on fine-tuning to achieve their results. In contrast, without fine-tuning, \textbf{LLMTime}~\citep{LLMTime} has demonstrated impressive zero-shot forecasting capabilities. The main idea of LLMTime is encoding time series data as a string of numerical digits, which emphasizes the importance of preprocessing and its inherent dependencies. To align the time series modality with the language modality, \textbf{Time-LLM}~\citep{TimeLLM} introduces the concept of patch reprogramming. This concept aims to mitigate the challenges of a large reprogramming space and attempts to connect time series local characteristics with language semantics, such as ``short up''. While patch reprogramming offers more flexibility by reducing the reliance on large-scale time series corpora, it also presents the challenge of adapting time series data to fit the characteristics of large language models.

\subsubsection{Pre-training}
Another approach that focuses on building large-scale time series corpora to pre-train time series foundation models from scratch has emerged. An example is \textbf{Lag-LLaMA}~\citep{LagLlama}, which uses a decoder-only Transformer to generate forecasting results, enabling probabilistic forecasting. Lag-LLaMA also consolidates 27 publicly available forecasting datasets from various domains to create a comprehensive pre-training corpus. On the other hand, \textbf{TimesFM}~\citep{TimesFM} extends beyond publicly available forecasting datasets by incorporating additional pre-training corpora based on Google Trends and Wiki Pageview statistics. It also utilizes synthetic datasets generated from piece-wise linear trends, autoregressive processes, and sine and cosine functions to capture universal characteristics. The entire pre-training corpus spans hundreds of billions of time steps. While most of the time series foundation models rely on temporal embedding, \textbf{CHRONOS}~\citep{CHRONOS} takes a different idea by learning a patch-based tokenizer, similar to conventional language models, to capture the intrinsic ``language'' of time series data. 

Conventional foundation models often overlook the relationships between variables in multivariate time series forecasting, typically extending to multivariate forecasting by independently combining univariate forecasts based on channel independence. \textbf{Uni2TS}~\citep{Uni2TS} addresses this limitation by explicitly considering the expansion to arbitrary multivariate time series using variate IDs. Additionally, it leverages a large-scale time series dataset called LOTSA, which accounts for multi-domain and multi-frequency characteristics.

\begin{sidewaystable}
\centering

\captionsetup{justification=raggedright, singlelinecheck=false} 
\caption{Taxonomy and Methodologies of Foundation Models for Time Series Forecasting}
\label{tab:RecentTSFLLMModels}

\renewcommand{\arraystretch}{2.5} 
\arrayrulecolor[gray]{0.8} 

\begin{tabular}{!{\color{black}\vrule width 1pt}>{\centering\arraybackslash}m{4cm}|>{\centering\arraybackslash}m{2.5cm}|>{\centering\arraybackslash}m{12cm}|>{\centering\arraybackslash}m{2cm}!{\color{black}\vrule width 1pt}}
\arrayrulecolor{black}\specialrule{1.0pt}{0pt}{0pt} 

\rowcolor[HTML]{808080}
{\color[HTML]{FFFFFF} \textbf{Approach}} & 
{\color[HTML]{FFFFFF} \textbf{Model Name}} &
\multicolumn{1}{c}{\cellcolor[HTML]{808080}\centering {\color[HTML]{FFFFFF} \textbf{Main Improvement \& Methodology}}} & 
{\color[HTML]{FFFFFF} \textbf{Publication}} \\
\arrayrulecolor[gray]{0.8} 

\multirow{4}{*}{\makecell[c]{Sequential modeling \\ with LLM}} 
 & GPT4TS & \makecell[l]{· Demonstrate the effectiveness of LLM for time series modeling \\ · Fine-tune the layer normalization and positional embedding parameters} & 2023 \\ \hhline{|~|-|-|-|}
 & PromptCast & \makecell[l]{· Enable text-level domain-specific knowledge for TSF \\ · Cast TSF problem into question and answering format} & 2023 \\ \hhline{|~|-|-|-|}
 & LLMTime & \makecell[l]{· Zero-shot TSF with pre-trained LLMs \\ · Covert time series input into a string of digits} & 2023 \\ \hhline{|~|-|-|-|}
 & Time-LLM & \makecell[l]{· Align time series modality into text modality \\ · Introduce patch reprogramming to mitigate a large reprogramming space} & 2024 \\ \hline

\multirow{4}{*}{Pre-training} 
 & Lag-Llama & \makecell[l]{· First pre-training based time series foundation model \\ · Pre-train a decoder-only model with autoregressive loss} & 2024 \\ \hhline{|~|-|-|-|}
 & TimesFM & \makecell[l]{· Pre-trained with hundreds of billions time steps \\ · Autoregressive decoding with arbitrary forecasting length} & 2024 \\ \hhline{|~|-|-|-|}
 & CHRONOS & \makecell[l]{· Learning the language of time series \\ · Utilize tokenizer to capture the intrinsic language of time series} & 2024 \\ \hhline{|~|-|-|-|}
 & UniTS & \makecell[l]{· Explicit consideration of multivariate TSF \\ · Provide variate IDs to directly consider multiple variables} & 2024 \\ 

\arrayrulecolor{black}\specialrule{1.0pt}{0pt}{0pt} 

\end{tabular}%

\end{sidewaystable}

\subsection{Advance of Diffusion Models}\label{sec4-4}


Diffusion models are renowned generative models that have gained prominence for their ability to produce high-quality images, as demonstrated by \textbf{DALL-E2} \citep{DALL-E2}, \textbf{Stable Diffusion} \citep{StableDiffusion}, and \textbf{Imagen} \citep{Imagen}. In addition to their success in image generation, diffusion models have shown excellent performance in various fields, such as audio generation, natural language processing, and video generation \citep{cao2024survey}. Consequently, there has been a growing number of research papers exploring their application in time series forecasting.

\revision{Diffusion models, theoretically proposed by \cite{sohl2015deep}, are inspired by non-equilibrium statistical physics.
They learn by progressively adding noise to the data and then gradually reversing the process to recover the original data.
However, this initial research lacked concrete methods for training, making practical implementation challenging.
This issue was addressed by Denoising Diffusion Probabilistic Models (DDPM) \citep{DDPM} and Noise Conditional Score Networks (NCSN) \citep{song2019generative}.
DDPM models the forward and reverse processes explicitly based on a probabilistic approach, enabling efficient learning.
NCSN, on the other hand, employs score matching techniques to refine the noise removal process, resulting in the generation of higher-quality samples.}

A diffusion model operates through two primary processes: the forward process and the reverse process. In the forward process, noise is gradually added to the data until it transforms into complete noise. Conversely, the reverse process reconstructs meaningful data from random noise. By training a denoising network, the model generates data by injecting random values and processing them through the reverse process.

Diffusion models have a significant advantage in modeling uncertainty through the forward and reverse processes.
By providing multiple possible prediction outcomes instead of a single prediction, it can reflect the uncertainty of the real world, making it highly beneficial for TSF.
Therefore, it can offer prediction confidence intervals or enable probabilistic forecasting, thereby contributing to decision-making.

\begin{figure}[!h]
\centering
\includegraphics[width=0.95\textwidth]{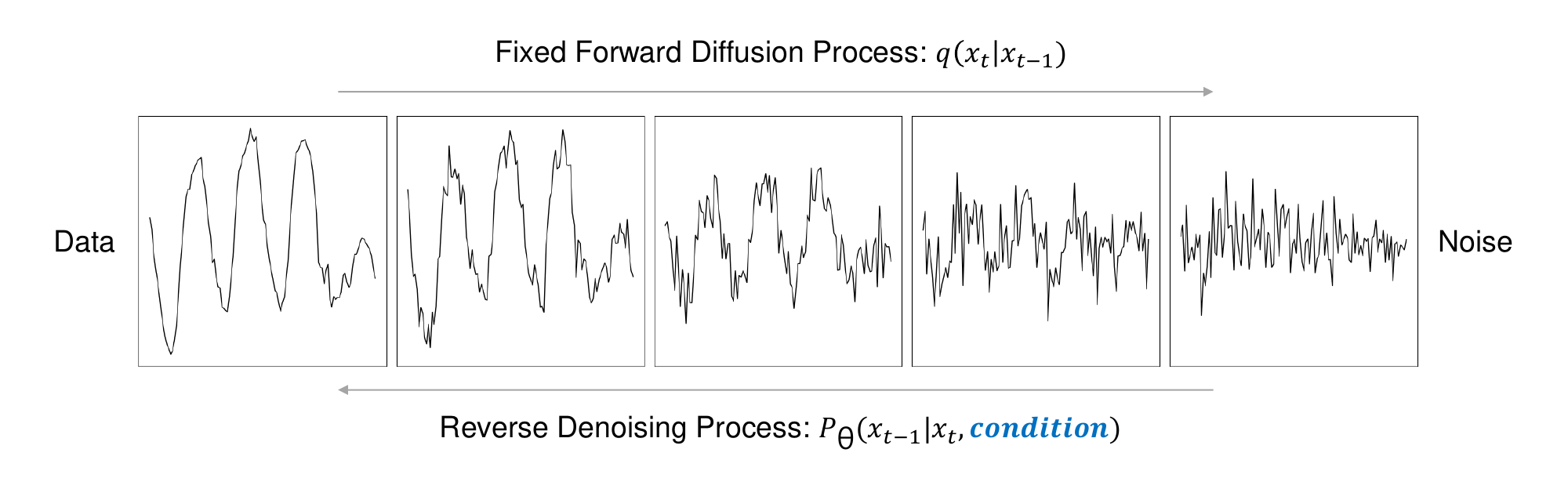}
\caption{Conditional Diffusion Process for Time Series Data}\label{Fig_4.4}
\end{figure}

For forecasting tasks, it is crucial to employ conditional generation, where conditions are incorporated into the model to produce data that aligns with the given conditions. In this context, historical data is used as a condition and injected into the model to enable it to learn and predict future values. Fig. \ref{Fig_4.4} illustrates the diffusion model process, visualizing the transition from data to noise and back to data. Here, the addition of conditions explains the concept of the conditional diffusion model.

Diffusion-based models for time series forecasting have achieved significant performance improvements through the introduction of effective conditional embedding, the integration of time series feature extraction, and advancements in the diffusion models themselves.
The following sections explain the key characteristics of diffusion-based models according to these classification criteria, and Table \ref{tab:DiffusionModels} provides a clear summary of this information.

\subsubsection{Effective Conditional Embedding}
Early diffusion-based models for time series forecasting focused on effective conditional embedding to guide the reverse process \citep{TMDM}. Typically, for forecasting tasks, conditional diffusion models use past data as a condition to predict the future.
Therefore, the meticulous construction of the condition is paramount for the denoising model to effectively learn the data and enhance prediction performance. This highlights the importance of data preparation in the forecasting process.
This preparation allows the model to effectively utilize past information, thereby enhancing the accuracy of future time-series predictions and improving the model's ability to learn from historical data.

\textbf{TimeGrad} \citep{TimeGrad} is the first notable diffusion model that operates using an autoregressive Denoising Diffusion Probabilistic Model (DDPM) \citep{DDPM} approach. It encodes past data with an RNN, using the hidden state in a conditional diffusion model for forecasting. The hidden state encapsulates information from past sequential data, capturing temporal dependencies effectively. The denoising network employs dilated ConvNets with residual connections, adopted from WaveNet \citep{WaveNet} and DiffWave \citep{kong2020diffwave}, which are designed for audio generation. Unlike the commonly used U-net for image synthesis in diffusion models, TimeGrad utilizes a broader receptive field suitable for time series data, similar to audio.
\textbf{CSDI} \citep{CSDI} captures temporal and feature dependencies of time series using a two-dimensional attention mechanism. Designed for both forecasting and imputation tasks, CSDI replaces RNNs with Transformers for imputation since RNNs are not suitable. The attention mechanism learns the relationships across all positions in the input sequence, capturing long-term dependencies effectively. Therefore, for long-term time series forecasting, using attention is more advantageous than RNNs and dilated convolution, as seen in TimeGrad.
\textbf{SSSD} \citep{SSSD} demonstrates that replacing the Transformer used as the denoising network in CSDI with the S4 model \citep{gu2021efficiently} yields superior performance. This is because S4 is more efficient and better at capturing long-term dependencies compared to the high computational complexity required for Transformer attention.
\textbf{TimeDiff} \citep{TimeDiff} enhances prediction accuracy by using future mixups during training, mixing past data with actual future values to generate conditional signals. Including some actual future values helps the model create effective conditional signals for more accurate predictions. Additionally, it addresses the boundary disharmony issue in non-autoregressive models like CSDI and SSSD. TimeDiff uses a simple statistical model, the Linear Autoregressive model, to provide an initial guess, alleviating the boundary disharmony problem.
\textbf{TMDM} \citep{TMDM} offers an extreme form of effective conditional embedding. It states that the best condition is the prediction itself, using prediction values from transformer-based models like Informer \citep{Informer} or Autoformer \citep{Autoformer}, which have shown good performance in time series forecasting tasks, as the condition. This allows the Transformer to handle the estimated mean while the diffusion model focuses on uncertainty estimation. Unlike previous researches that utilize conditional embeddings only in the reverse process, TMDM uses conditional information as prior knowledge for both the forward and reverse processes.

\subsubsection{Time Series Feature Extraction}
Time series data hides various features, and there are techniques to extract these unique characteristics effectively. Thus, many works have emerged that combine time series feature extraction methods with diffusion models to understand the complex patterns in time series data and improve prediction performance.

\paragraph{Decomposition}
Decomposition techniques involve breaking down time series data into components such as trend, seasonality, and irregularity, analyzing the unique patterns of each component. \textbf{Diffusion-TS} \citep{Diffusion-ts} points out that conventional methods cannot properly learn each component because the forward process causes the components to collapse. Therefore, it models the decomposed time series data individually during the diffusion process, learning each component, such as trend, seasonality, and residuals, independently and then recombining them to restore the entire time series data.

\paragraph{Frequency Domain}
Fourier analysis, a type of decomposition technique, converts time series data into the frequency domain to analyze periodic components. This method helps identify periodic patterns and remove noise.
\cite{DiffusionInfrequencyDomain} explore the idea that the representation of time series data in the frequency domain can provide useful inductive biases for score-based diffusion models. This paper demonstrates that the components of time series data are more clearly represented in the frequency domain, and diffusion models in the frequency domain outperform those in the time domain.

\paragraph{Multi-Scale}
Multi-scale techniques analyze time series data at various time scales, effectively extracting long-term trends and diverse features. This approach plays a crucial role in understanding the complex patterns of time series data to improve prediction performance.
\textbf{MG-TSD} \citep{MG-TSD} observes that the forward process of diffusion models aligns with gradually smoothing fine data into coarser representations. It suggests that coarse-granularity data can serve as effective guides in the intermediate stages of diffusion models. In other words, the initial stages of the reverse process involve coarse-granularity data, guiding the process to intermediate-stage targets. This multi-granularity level approach helps learn various levels of information, enhancing prediction stability and accuracy.
\textbf{mr-Diff} \citep{mr-Diff} constructs time series data at multiple resolutions, performing sequential predictions at each resolution. In the initial stages, it predicts coarse data, and in subsequent stages, it uses these predictions as conditions to generate finer data gradually. This structure incrementally adds finer patterns and noise at each stage, ultimately reconstructing high-resolution time series data. This allows for effective prediction of both long-term trends and short-term fluctuations in time series data.

\subsubsection{Additional Approaches}
Beyond the two main approaches mentioned earlier, there are numerous examples where techniques evolving from the diffusion framework itself are applied to models for time series forecasting.

\paragraph{Score-Based Generative Modeling through SDEs}
 \textbf{Score-Based Generative Modeling through Stochastic Differential Equations (SDEs)} \citep{SDE} serves as continuous energy-based generative models, handling data in a continuous time domain, thus reflecting natural and precise changes and noise in the real world. The previously discussed methods are based on \textbf{DDPM} \citep{DDPM}, which add and remove noise incrementally using fixed-time steps. However, the DDPM approach faces challenges such as specific functional constraints and sensitivity to hyperparameters. Some research applying SDEs overcomes these limitations and explores diverse approaches.
\textbf{ScoreGrad} \citep{Scoregrad} is the first work to apply SDE, overcoming the constraints of DDPM-based models and offering a more flexible and powerful generative model.
\textbf{D\textsuperscript{3}M} \citep{D3M} addresses the limitations of conventional SDEs, such as the complexity of determining drift and diffusion coefficients and slow generation speed, by utilizing a decomposable noise-removal diffusion model based on explicit solutions. This method reduces computational complexity through clear SDE formulations and separates the signal decay and noise injection processes in model design. As a result, it enhances model efficiency and accuracy while accelerating generation speed.
\cite{DiffusionInfrequencyDomain} improve denoising score matching in the frequency domain by using mirrored Brownian motions instead of standard Brownian motion, emphasizing the symmetry between components when applying SDEs.

\paragraph{Latent Diffusion Model}
\textbf{Latent Diffusion Model} \citep{StableDiffusion} is a generative model that operates not on the original data directly but within a latent space where the diffusion process takes place.	
Transforming data into the latent space reduces complexity and stabilizes the training process, resulting in high-quality outputs.
\textbf{LDT} \citep{LDT} applies the concept of the latent diffusion model to time series data, addressing the non-stationarity issues that often arise during the compression into the latent space.
Dynamically updating statistical properties during the autoencoder training process effectively overcomes these challenges and enhances model performance.

\paragraph{Guidance}
Some works use guidance instead of explicitly feeding conditions into the denoising network for forecasting.
\textbf{Diffusion-TS} \citep{Diffusion-ts} employs classifier guidance, using a separate classifier to guide the sampling process through the gradients of the classifier. This method maintains the basic unconditional diffusion model while performing conditional generation tasks through various classifiers. It generates samples that are better aligned with specific conditions, resulting in higher-quality outputs. However, classifier guidance requires a classifier for each time step, which necessitates training a new classifier instead of using pre-trained ones.
\textbf{LDT} \citep{LDT} uses classifier-free guidance, learning both conditional and unconditional models within a single model to perform conditional sampling. This work eliminates the need for an additional classifier and implicitly obtains a classifier, making implementation simpler and more efficient.
\textbf{TSDiff} \citep{TSDiff} proposes observation self-guidance, allowing the use of an unconditional diffusion model for conditional generation without separate networks or modifications to the training process.

\begin{sidewaystable}
\centering

\captionsetup{justification=raggedright, singlelinecheck=false}
\caption{Taxonomy and Methodologies of Diffusion Models for Time Series Forecasting}
\label{tab:DiffusionModels}

\footnotesize 
\renewcommand{\arraystretch}{2} 

\arrayrulecolor[gray]{0.8} 
\begin{tabular}{!{\color{black}\vrule width 1pt}>{\centering\arraybackslash}m{3cm}|>{\centering\arraybackslash}m{2.5cm}|>{\centering\arraybackslash}m{10.2cm}|>{\centering\arraybackslash}m{1.5cm}|>{\centering\arraybackslash}m{1.5cm}|>{\centering\arraybackslash}m{1.5cm}!{\color{black}\vrule width 1pt}}

\arrayrulecolor{black}\specialrule{1.0pt}{0pt}{0pt} 

\rowcolor[HTML]{808080}
{\color[HTML]{FFFFFF} \textbf{Main Improvement}} & 
{\color[HTML]{FFFFFF} \textbf{Model Name}} &
\multicolumn{1}{c}{\cellcolor[HTML]{808080}\centering {\color[HTML]{FFFFFF} \textbf{Main Methodology}}} & 
{\color[HTML]{FFFFFF} \textbf{Diffusion Type}} & 
{\color[HTML]{FFFFFF} \textbf{Conditional Type}} & 
{\color[HTML]{FFFFFF} \textbf{Publication}} \\

\arrayrulecolor[gray]{0.8} 

\multirow{5}{*}{\makecell[c]{Effective \\ Conditional \\ Embedding}}
 & TimeGrad & \makecell[l]{· Autoregressive DDPM using RNN \& Dilated Convolution} & DDPM & Explicit & 2021 \\ \hhline{|~|-|-|-|-|-|}
 & CSDI & \makecell[l]{· 2D Attention for Temporal \& Feature Dependency \\ · Self-supervised Training for Imputation} & DDPM & Explicit & 2021 \\ \hhline{|~|-|-|-|-|-|}
 & SSSD & \makecell[l]{· Combination of S4 model} & DDPM & Explicit & 2023 \\ \hhline{|~|-|-|-|-|-|}
 & TimeDiff & \makecell[l]{· Future Mixup \\ · Autoregressive Initialization} & DDPM & Explicit & 2023 \\ \hhline{|~|-|-|-|-|-|}
 & TMDM & \makecell[l]{· Integration of Diffusion and Transformer-based Models} & DDPM & Explicit & 2024 \\ \hline

\multirow{5}{*}{\makecell[c]{Time-series \\ Feature Extraction}}
 & Diffusion-TS & \makecell[l]{· Decomposition techniques \\ · Instance-aware Guidance Strategy} & DDPM & Guidance & 2024 \\ \hhline{|~|-|-|-|-|-|}
 & Diffusion in Frequency & \makecell[l]{· Diffusing in the Frequency Domain} & SDE & Explicit & 2024 \\ \hhline{|~|-|-|-|-|-|}
 & MG-TSD & \makecell[l]{· Multi-granularity Data Generator \\ · Temporal Process Module \\ · Guided Diffusion Process Module} & DDPM & Explicit & 2024 \\ \hhline{|~|-|-|-|-|-|}
 & mr-Diff & \makecell[l]{· Integration of Decomposition and Multiple Temporal Resolutions} & DDPM & Explicit & 2024 \\ \hline

\multirow{2}{*}{SDE}
 & ScoreGrad & \makecell[l]{· Continuous Energy-based Generative Model} & SDE & Explicit & 2021 \\ \hhline{|~|-|-|-|-|-|}
 & D\textasciicircum{}3 M & \makecell[l]{· Decomposable Denoising Diffusion Model based on Explicit Solutions} & SDE & Explicit & 2024 \\ \hline

\multirow{1}{*}{Latent Diffusion Model}
 & LDT & \makecell[l]{· Symmetric Time Series Compression \\ · Latent Diffusion Transformer} & DDPM & Guidance & 2024 \\ \hline

\multirow{1}{*}{Guidance}
 & TSDiff & \makecell[l]{· Observation Self-guidance} & DDPM & Guidance & 2023 \\ \hline

\arrayrulecolor{black}\specialrule{1.0pt}{0pt}{0pt} 

\end{tabular}%

\end{sidewaystable}

\subsection{Debut of the Mamba}\label{sec4-5}

\subsubsection{History of the State Space Models (SSMs)}
One of the notable recent developments is the emergence of a new architecture called \textbf{Mamba} \citep{Mamba}. In an atmosphere previously influenced by Transformers, Mamba has garnered significant interest from researchers and is rapidly establishing its own ecosystem.

RNNs, which were once the mainstream in sequence modeling, lost their dominance after the advent of Transformers. This was due to the limitations in information encapsulation of a ``context'' (single vector) in RNN-based encoder-decoder models and the slow training speed inherent in their recurrent nature. In contrast, the parallelism of the attention mechanism and its ability to focus on all individual pieces of information overcame the limitations of RNNs and demonstrated superior performance. However, new challenges have emerged with Transformers, such as the quadratic computational complexity, which limits the window length, and the increased memory requirements for processing long sequences. Subsequently, many efforts have been made to overcome the limitations of both approaches while preserving their advantages. In this context, some research that continues the philosophy of RNNs has turned its attention to state space models (SSMs) \citep{statespacemodel}.

\begin{figure}[H]
\centering
\includegraphics[width=1.0\textwidth, trim=0cm 5cm 0cm 5cm, clip]{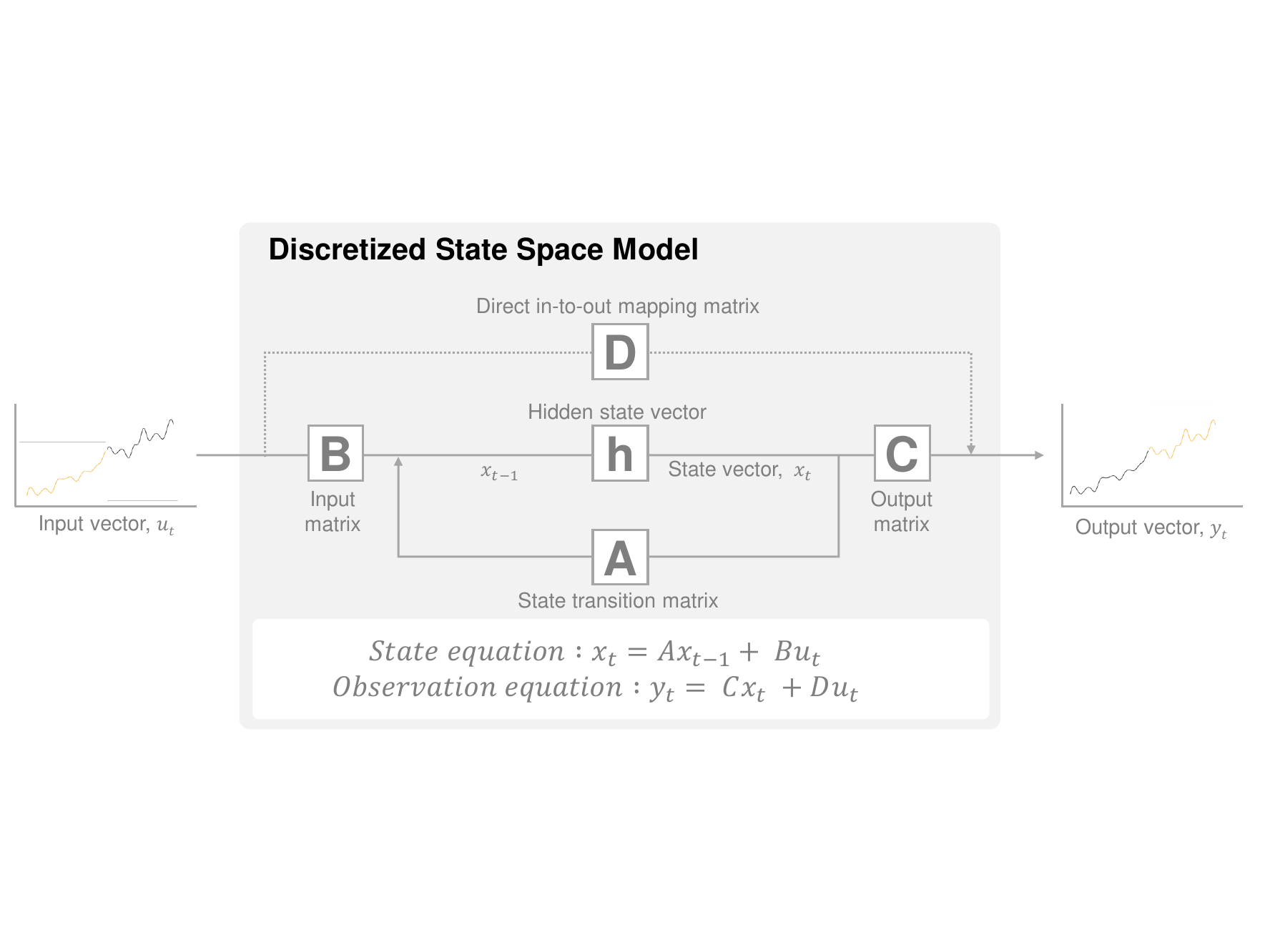}
\caption{Diagram of Discretized State Space Model}\label{Diagram of Discretized State Space Model}
\end{figure}

SSM is a mathematical framework used to model the dynamic state of a system that changes over time, compressing only the essential information for effective sequential modeling. SSM describes the relationship between the internal state of the system and external observations and is used in various fields such as control theory, signal processing, and time series analysis. It comprises a `State equation' that explains how the internal state changes over time and an `Observation equation' that explains how the internal state is transformed into external observations. Although it is a continuous model that performs linear transformations on the current hidden state and input, it can handle discrete sequences like time series through discretization (Fig. \ref{Diagram of Discretized State Space Model}). It closely resembles RNNs in that it combines observation data and hidden state data.

However, SSMs have different characteristics from RNNs in that they are linear and time-invariant (LTI). This means that the operations performed on each token do not vary, allowing for the pre-computation of global kernels. In other words, the parameter matrices of the system do not change over time and operate consistently across time. Therefore, the ability to precompute globally applicable kernels enables parallel processing, which can overcome the limitations of RNNs.

\begin{figure}[H]
\centering
\includegraphics[width=0.9 \textwidth, trim=0cm 5cm 0cm 5cm, clip]{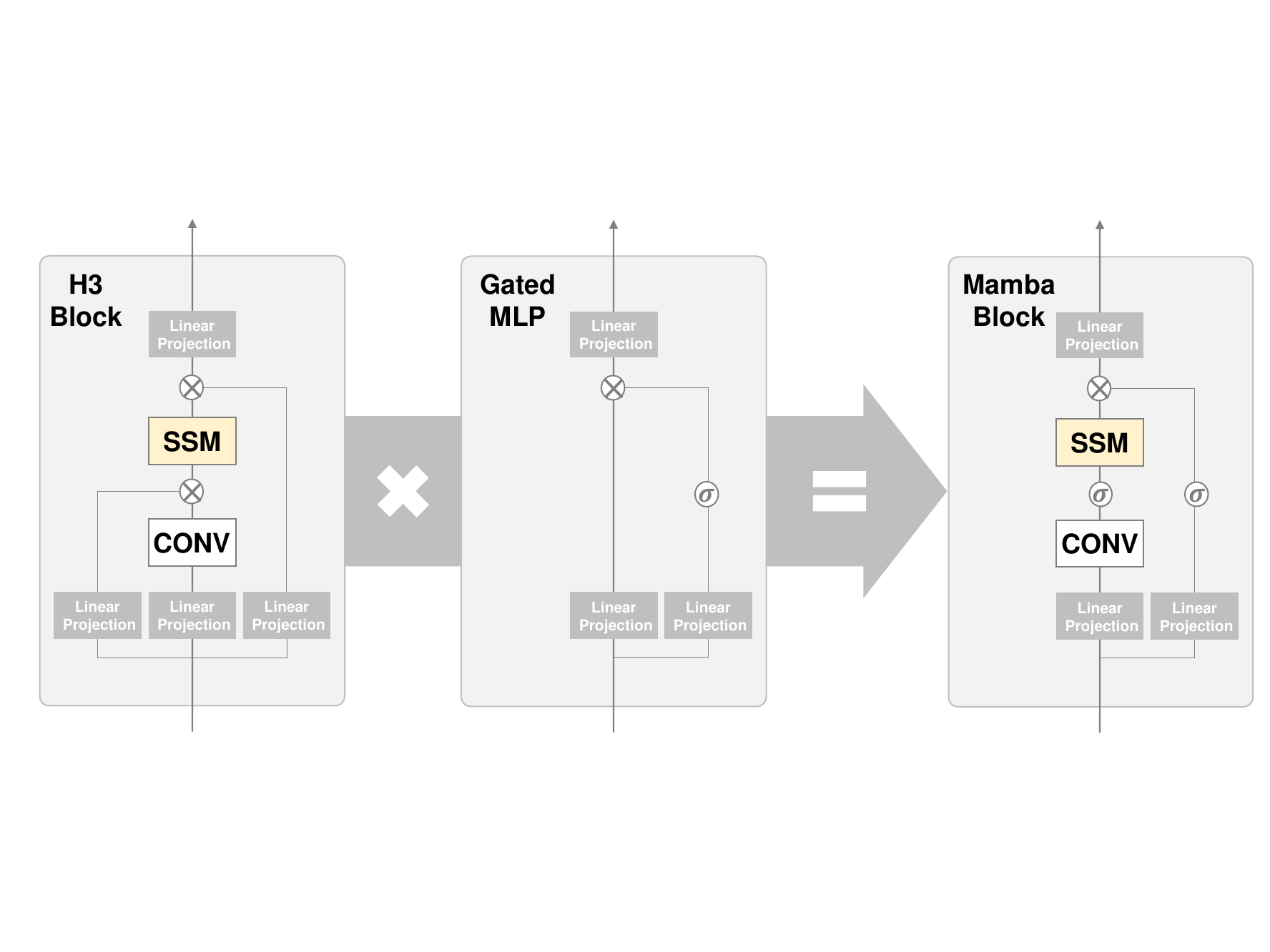}
\caption{Structure of the Mamba Block}\label{Structure of the Mamba block}
\end{figure}

Early applications of SSMs, such as the S4 \citep{S4} model, utilized diagonalization of the transition matrix to effectively model long-term dependencies in long and complex sequence data. These models were able to achieve high performance when combined with deep learning architectures like Transformers, RNNs, and CNNs. Subsequently, based on S4, advanced blocks like H3 \citep{H3} were developed, which hierarchically structured convolution, gating, and MLP to provide more efficient and powerful sequence modeling (Fig. \ref{Structure of the Mamba block}).

\subsubsection{Introduction of the Mamba}
SSMs have fixed state transition and observation equations, which limits their ability to flexibly handle input data. Furthermore, due to their inherent continuous system origins, they are weaker in information-dense and discrete domains such as text.

Mamba addresses these limitations of SSMs by introducing an advanced deep structured SSM model with selective SSM. It is designed so that the parameters of the SSM dynamically change depending on the input, allowing for selective memory or disregard of specific parts of the sequence, enabling efficient data processing. However, this approach, while enhancing the system's flexibility to better learn complex patterns, sacrifices the parallel processing advantages of SSMs. To compensate, traditional techniques such as kernel fusion, parallel scan, and recomputation are applied to efficiently implement selective SSMs on GPUs. Mamba adopts a simplified architecture centered around selective SSM, replacing the first multiplicative gate in the traditional H3 block with an activation function and incorporating an SSM into the gated MLP block. Due to the absence of attention mechanisms or separate MLP blocks, computational complexity is reduced, resulting in efficient training, fast inference, and high scalability and versatility.

\subsubsection{Applications of the Mamba}
In the field of time-series forecasting, there is a growing trend to apply Mamba. A lot of variants based on Mamba have been proposed to address TSF tasks, and these models are reported to exceed the performance of Transformer SOTA models.

Most of these models typically share the following common features:
\begin{itemize}
    \item They aim to ensure stability in large networks. Mamba, based on SSM, heavily relies on the eigenvalues of the system's dynamic matrix. Since the general solution of the system is expressed as an exponential function of the eigenvalues, if the real part of the eigenvalues is positive, the system cannot converge and becomes unstable. Therefore, a key feature is the ideation aimed at improving this instability.
    \item They explicitly incorporate the learning of channel correlations. Recent research has demonstrated the superiority of channel dependence (CD) learning strategies, and the adoption of the CD strategy by the iTransformer, which shows SOTA performance, has further heightened this trend. This is a common feature observed in other foundational models as well.
    \item They incorporate various techniques from transformer-based models. Mamba-based models, which integrate various techniques from Transformers such as patching, frequency domain application, bidirectional processing, and FFN incorporation, are demonstrating significant performance improvements. The diverse techniques developed over many years of research on Transformer models provide valuable ideation for Mamba variants.
\end{itemize}

The following are representative examples of Mamba-based models for handling TSF tasks and their key technical features.

\paragraph{Embedding and Multi-Scale Learning}
In this section, we introduce methodologies for embedding and learning from data at various scales. These approaches primarily focus on richly extracting information to capture long-term dependencies and context.

\textbf{TimeMachine} \citep{TimeMachine} aims to capture long-term dependencies and context at different scales through two-stage embeddings. It is divided into internal and external embeddings based on embedding dimensions, and each section consists of two parallel Mamba modules that explore temporal dependencies and channel dependencies, respectively. The internal Mamba module operates at a low resolution to capture both global and local contexts, while the external Mamba module operates at a high resolution to capture the global context.

\paragraph{Channel Correlation Learning}
Next, we examine models that focus explicitly on learning the inter-channel correlations in time series data. These models are centered around techniques that effectively integrate information from each channel and model their interdependencies.

\textbf{S-Mamba} \citep{S-Mamba} features bi-directional Mamba blocks to consider both past and future information, enabling the learning of inter-channel correlations. The role of learning temporal dependencies is assigned to the Feed Forward Network (FFN). By clearly separating the roles of Mamba blocks and FFN, computational efficiency is improved, ensuring the stability of the Mamba architecture in large networks. 
In \textbf{SiMBA} \citep{SiMBA}, channel modeling is achieved through Einstein FFT (EinFFT), while sequence modeling is handled by Mamba modules. After applying the Fast Fourier Transform (FFT), the real and imaginary parts are separated, and their respective weights are learned. Channel mixing is performed through Einstein Matrix Multiplication (EMM), creating new data that reflects the relational information between channels, thus internalizing channel relationships. Additionally, to ensure system stability, eigenvalues of the state matrix are adjusted to be negative through Fourier transformation and nonlinear layers. 
\textbf{MambaTS} \citep{MambaTS} reconstructs sequences by integrating past information of all variables to learn channel correlations. Since variable information is integrated in advance, unnecessary convolution in Mamba blocks is removed to enhance computational efficiency, and dropout is applied to Temporal Mamba Blocks (TMB) to reduce overfitting. Additionally, Variable Permutation Training (VPT) is introduced to dynamically determine the optimal order of integrated variable information, enabling predictions based on the optimal sequence of variables. 
\textbf{C-Mamba} \citep{C-Mamba} generates new virtual sample data by linearly combining different channels (channel mixup), which is expected to enhance generalization performance. It uses a main block that combines patched Mamba modules with attention modules for learning channel relationships.

\paragraph{Sequence Information and Dependency Learning}
This category emphasizes methods for learning the sequential information and dependencies in time series data. It proposes various techniques for modeling both long-term and short-term dependencies within the series.

\textbf{Mambaformer} \citep{Mambaformer} is a hybrid model that combines Mamba with a Transformer decoder framework. Since the Mamba block naturally internalizes the sequence order information, positional embedding is unnecessary. Long-term dependencies are learned through the Mamba block, while short-term dependencies are captured through the self-attention layer, effectively capturing the overall dependencies. This approach overcomes the computational efficiency limitations of attention mechanisms. 
\textbf{Bi-Mamba+} \citep{Bi-Mamba+} integrates patching techniques to finely learn inter-dependencies in the data. To preserve long-term information, it introduces Mamba+ blocks, which add a forget gate to the Mamba block. It also employs a Bi-Mamba+ encoder to process input sequences bidirectionally. Using the Spearman correlation coefficient, the Series-Relation-Aware (SRA) decider is designed to automatically select channel tokenization strategies (CI or CD).
\textbf{DTMamba} \citep{DTMamba} is composed of Dual Twin Mamba blocks, effectively learning long-term dependencies in time series data channel independently. Each Twin Mamba block consists of two parallel Mamba structures that process the same input data to capture different patterns effectively. One Mamba structure learns detailed patterns and short-term variations, while the other learns overall patterns and long-term trends.

\paragraph{Theoretical Frameworks and Efficient Modeling}
Lastly, we explore models that introduce new theoretical frameworks or propose efficient modeling techniques. These models focus on effectively capturing the dynamic characteristics of time series data and employ theoretical approaches and methodologies that enhance computational efficiency.

\textbf{Time-SSM} \citep{Time-SSM} proposes a theoretical framework called the Dynamic Spectral Operator, which extends the Generalized Orthogonal Basis Projection (GOBP) theory for efficient use of SSM. The Dynamic Spectral Operator explores the changing spectral space over time to effectively capture the dynamic characteristics of time series data. To implement this, a novel variant of the SSM basis called Hippo-LegP is proposed, enabling more precise modeling of time series data and achieving optimal performance through S4D-real initialization. This allows it to demonstrate excellent performance with only about one-seventh of the parameters required by Mamba models.
\textbf{Chimera} \citep{Chimera} features the use of 2D SSM to capture dependencies independently along the time and variable axes. By updating states in parallel along both axes, it achieves efficient computation.

The emergence of a new deep learning architecture, Mamba, is causing a shift in the long-standing hegemony of deep learning architectures. In current TSF tasks, which increasingly deal with long-term sequences, Mamba's strengths—efficient sequence processing, selective information retention, simplified architecture, and hardware optimization—prove to be highly valuable. These features allow Mamba-based models to overcome the limitations of Transformers, showing rapid performance improvements in a short period and suggesting a new direction for deep learning modeling. The growth trajectory of Mamba raises attention to whether it will become the new dominance in this field.

\begin{sidewaystable}
\centering
\captionsetup{justification=raggedright, singlelinecheck=false} 
\caption{Taxonomy and Methodologies of Mamba Models for Time Series Forecasting}
\label{tab:MambaModels}

\footnotesize 
\renewcommand{\arraystretch}{2.2} 
\arrayrulecolor[gray]{0.8} 

\begin{tabular}{!{\color{black}\vrule width 1pt}>{\centering\arraybackslash}m{6.5cm}|>{\centering\arraybackslash}m{2cm}|>{\raggedright\arraybackslash}m{7cm}|>{\centering\arraybackslash}m{1.5cm}|>{\centering\arraybackslash}m{1.5cm}|>{\centering\arraybackslash}m{1.5cm}!{\color{black}\vrule width 1pt}}

\arrayrulecolor{black}\specialrule{1.0pt}{0pt}{0pt} 

\rowcolor[HTML]{808080}
{\color[HTML]{FFFFFF} \textbf{Main Improvement}} & 
{\color[HTML]{FFFFFF} \textbf{Model Name}} &
\multicolumn{1}{c}{\cellcolor[HTML]{808080}\raggedright {\color[HTML]{FFFFFF} \textbf{Main Methodology}}} & 
{\color[HTML]{FFFFFF} \textbf{Channel Correlation}} & 
{\color[HTML]{FFFFFF} \textbf{Base}} & 
\multicolumn{1}{c|}{\cellcolor[HTML]{808080}{\color[HTML]{FFFFFF} \textbf{Publication}}} \\
\arrayrulecolor[gray]{0.8} 

\multirow{1}{*}{Embedding and Multi-Scale Learning} & TimeMachine & \makecell[l]{· Integrated Quadruple Mambas} & CD & Mamba & 2024 \\ \hhline{|-|-|-|-|-|-|}
\multirow{4}{*}{Channel Correlation Learning} & S-Mamba & \makecell[l]{· Channel Mixing: Mamba VC Encoding Layer \\ · Sequence Modeling: FFN TD Encoding Layer} & CD & \makecell[c]{Mamba \\ MLP} & 2024 \\ \hhline{|~|-|-|-|-|-|}
& SiMBA & \makecell[l]{· Channel Mixing: Einstein FFT (EinFFT) \\ · Sequence Modeling: Mamba} & CD & Mamba & 2024 \\ \hhline{|~|-|-|-|-|-|}
& MambaTS & \makecell[l]{· Temporal Mamba Block (TMB) \\ · Variable Permutation Training (VPT)} & CD & Mamba & 2024 \\ \hhline{|~|-|-|-|-|-|}
& C-Mamba & \makecell[l]{· Channel Mixup \\ · C-Mamba Block (PatchMamba + Channel Attention)} & CD & Mamba & 2024 \\ \hline

\multirow{3}{*}{Sequence Information and Dependency Learning} & Mambaformer & \makecell[l]{· Mambaformer (Attention + Mamba) Layer} & CI & \makecell[c]{Mamba \\ Transformer} & 2024 \\ \hhline{|~|-|-|-|-|-|}
& Bi-Mamba+ & \makecell[l]{· Series-Relation-Aware (SRA) Decider \\ · Mamba+ Block \\ · Bidirectional Mamba+ Encoder} & CI/CD & Mamba & 2024 \\ \hhline{|~|-|-|-|-|-|}
& DTMamba & \makecell[l]{· Dual Twin Mamba Blocks} & CI & Mamba & 2024 \\ \hline

\multirow{2}{*}{Theoretical Frameworks and Efficient Modeling} & Time-SSM & \makecell[l]{· Dynamic Spectral Operator with Hippo-LegP} & CD & Mamba & 2024 \\ \hhline{|~|-|-|-|-|-|}
& Chimera & \makecell[l]{· 2-Dimensional State Space Model} & CD & Mamba & 2024 \\ \hline

\arrayrulecolor{black}\specialrule{1.0pt}{0pt}{0pt} 

\end{tabular}%
\end{sidewaystable}

\section{TSF Latest Open Challenges \& Handling Methods}\label{sec5}

\subsection{Channel Dependency Comprehension}\label{sec5-1}

\paragraph{Spread of Channel Independent Strategy}
Multi-variate time series (MTS) forecasting primarily hinges on how well it can learn the short-term and long-term temporal dependencies. However, many recent real-world datasets predominantly deal with multivariate data, where the relationships between variables carry significant semantic information. As the relationships between variables become more complex, models can provide more information and thus improve predictive performance by leveraging this complexity.

Traditionally, it has been vaguely assumed that understanding the relationships between variables in time series forecasting problems would make better performance. Despite being obtained from different instruments, the data observe the same phenomenon, which intuitively suggests that they offer rich interpretations from various perspectives. However, with the PatchTST \citep{PatchTST} model adopting a channel-independent (CI) strategy and achieving state-of-the-art (SOTA) performance, research has begun to question the previously assumed channel-dependent (CD) strategy. These studies have shown that high performance can be attained without learning the interactions between variables.

The CI strategy simplifies the model by excluding inter-variable modeling, allowing it to focus solely on learning the temporal dependencies of each channel. This approach reduces model complexity and enables faster inference, while also mitigating the risk of overfitting due to noise from inter-variable interactions. Additionally, since the addition of new channels does not affect the model, it can adapt flexibly to changes in data. These advantages have led many studies to adopt the CI strategy, resulting in improved performance. However, the CI strategy did not consistently show superior performance across all studies. 
The CD strategy still demonstrated high performance in many studies, such as iTransformer \citep{itransformer}. Additionally, both strategies showed inconsistent performance depending on the datasets used.
Consequently, without clear justification, both CI and CD strategies were employed according to the researchers' preferences for a period of time.

\paragraph{Importance of Learning Channel Correlations}
Learning the correlations between variables remains important. In multivariate time series data, each variable does not change independently but is interdependently related to others. Even if these relationships sometimes introduce noise or fail to provide critical information, the relationships themselves are not devoid of meaning. The information from multivariate variables often intertwines to create complex patterns that cannot be captured by a single variable. Understanding the correlations between multiple variables helps interpret these complex patterns.

Modeling the correlations between variables is also crucial for improving prediction accuracy. Especially when dealing with long sequence patterns, it is essential to understand the numerous local patterns within them. During this process, important causal relationships are often derived from other variables or exogenous factors. In the case of noisy data, learning the correlations between variables can help extract key information effectively. By comprehensively observing the information from multiple variables, it is possible to emphasize important features and complement missing information.

\paragraph{What Makes CI Look Better?}
If the CD strategy is important and has the potential for significant performance improvements, why do many studies show that the CI strategy performs better? To answer this question, some research has been conducted, and the following summary can be provided based on the \textbf{“The capacity and robustness trade-off: Revisiting the channel independent strategy for multivariate time series forecasting”} 
 \citep{CapacityRobustnessTrade-off} paper.

According to the study, an examination of the Auto-correlation Function (ACF) values between the training set and test set of the benchmark data, which have been used in various experiments, revealed a distribution shift. The ACF can identify the correlation between data values at specific time lags in time series data.

A distribution shift refers to the difference in statistical properties between the training data and test data extracted from the same dataset. Distribution shifts can occur for various reasons, often due to anomalies in the training series or variations in the trend, but sometimes the exact cause cannot be clearly defined. An important finding from the study is that the CI strategy demonstrates relatively greater robustness to distribution shifts. Since the CI strategy relies on the average ACF of all channels, it is less sensitive to distribution shifts compared to the CD strategy, which depends on the ACF variations of individual channels. Additionally, by excluding inter-variable relationships, the CI strategy simplifies the model, reducing the likelihood of overfitting and enhancing robustness. Therefore, the CI strategy has been able to demonstrate good performance on benchmark datasets where distribution shifts are present.

\begin{figure}[H]
\centering
\includegraphics[width=1.0\textwidth, trim=0cm 0.5cm 0cm 0.5cm, clip]{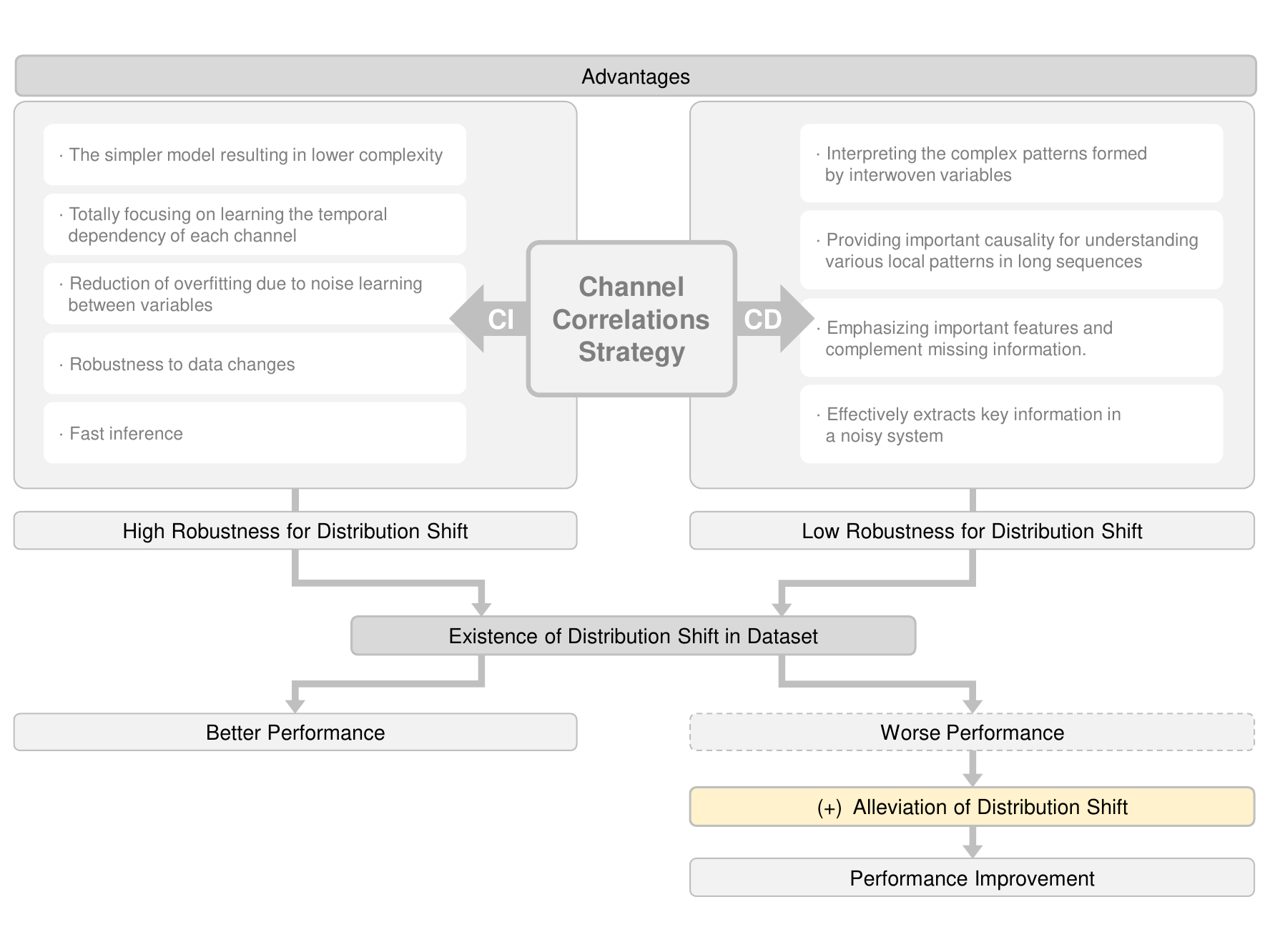}
\caption{Comparison of CI and CD Strategies in Channel Correlations}
\label{Comparison of CI and CD Strategies in Channel Correlations}
\end{figure}

However, this does not imply that the CI strategy is superior to the CD strategy. It merely suggests that, in some datasets, the positive effects resulting from the robustness of the CI strategy outweigh the advantages of learning inter-variable relationships. Conversely, if distribution shifts can be effectively alleviated, the CD strategy can provide much more useful information than the CI strategy. To this end, even the application of simple regularization methods, such as Predict Residual with Regularization (PRReg), adapting low-rank layers, and using the MAE loss function, can reverse the performance gap. Furthermore, various distribution shift alleviation methodologies have been researched, and these can be applied in a model-agnostic manner. The details of these methodologies will be discussed in the next section, and the previous content has been illustrated in Fig. \ref{Comparison of CI and CD Strategies in Channel Correlations}

\paragraph{Recent Approaches}
As the effectiveness of learning channel correlation has been validated, recent multivariate time series forecasting problems have predominantly adopted the CD strategy. In contrast, the CI strategy is increasingly applied in a limited way, primarily in univariate models focused on temporal dependency.

Approaches to addressing recent MTS problems can be broadly categorized into three types. The first approach involves explicitly integrating modules for channel mixing into the backbone model. Most models adopt this strategy, aiming to maximize effectiveness by explicitly incorporating channel correlation. However, as discussed earlier, distribution shifts in the dataset can potentially degrade the performance of the CD strategy, necessitating the introduction of appropriate alleviation methods. By employing this combination, the strengths of learning channel correlations can be applied more reliably, leading to enhanced model performance.
The distribution shift alleviation methods will be discussed in detail in Section \ref{sec5-2}.

The second approach implicitly incorporates channel correlations into the input. Instead of explicitly integrating channel learning modules into the backbone model, this method takes a preprocessing approach by mixing channel information in advance to create a new input, which is then fed into the prediction model. This new input combines the mixed channel information with the original individual channel data before being processed by the prediction model. By inherently reflecting the relationships between channels, this method allows even a simple CI backbone to achieve the effectiveness of a more complex CD model. Models like \textbf{SOFTS} \citep{SOFTS} and \textbf{C-LoRA} \citep{C-LoRA} fall into this category.

\begin{figure}[H]
\centering
\includegraphics[width=1.0\textwidth, trim=0cm 3cm 0cm 3cm, clip]{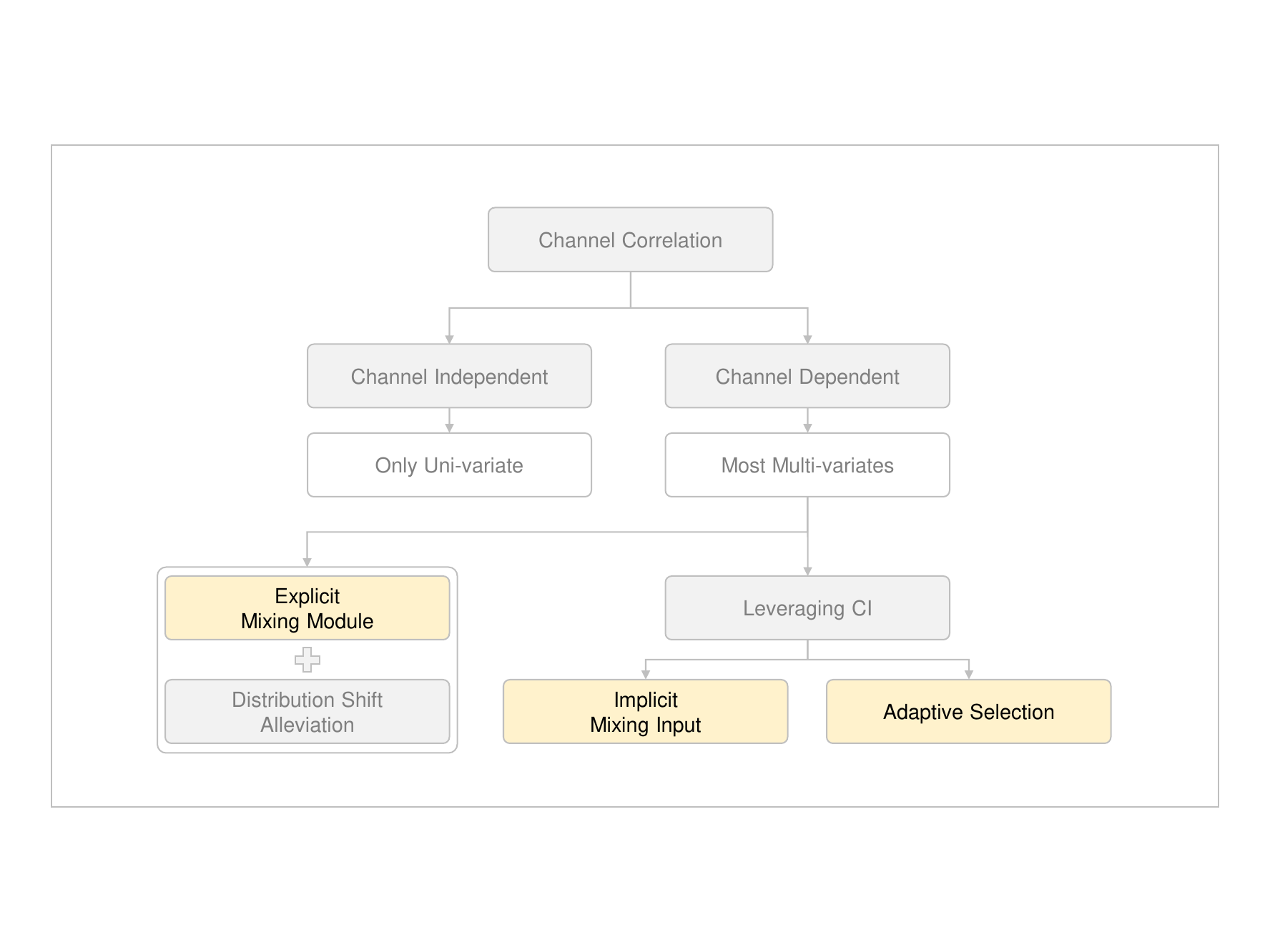}
\caption{Recent Approaches to Channel Strategies}
\label{Recent Approaches to Channel Strategies}
\end{figure}

The third approach adaptively selects between the CI and CD strategies. The model is designed to operate in both directions, allowing it to choose the most effective strategy based on the characteristics of the dataset being used. In some cases, like \textbf{TSMixer} \citep{TSMixer}, the user can manually select the appropriate strategy, while in others, such as \textbf{Bi-Mamba+} \citep{Bi-Mamba+}, the model automatically determines the optimal strategy. The second and third approaches both utilize the CI strategy alongside the CD strategy, aiming to integrate the advantages of both into the model. A summary of the criteria for selecting these strategies is presented in Fig. \ref{Recent Approaches to Channel Strategies}

In the past, the criteria for selecting channel strategies were ambiguous, but recent research has provided clear guidelines. Based on these advancements, users can now choose the appropriate channel strategy according to the characteristics of their dataset, effectively improving model performance through various approaches.

\subsection{Alleviation of Distribution Shift}\label{sec5-2}
Many real-world time series data exhibit non-stationarity, where their distribution gradually changes. For instance, trends such as increasing electricity consumption and shifts in consumer preferences can lead to changes in data distribution. This non-stationarity introduces challenges for the generalization of time series forecasting (TSF) models by creating distribution discrepancies within the training data and between the training and test data. To mitigate the issue of distribution shift caused by non-stationarity in time series forecasting, various research efforts have been proposed. 

\revision{Representative approaches for addressing distribution shifts include Domain Adaptation, Transfer Learning, and Robustness Techniques.
\textbf{Domain Adaptation} aims to reduce the distribution differences between the source and target domains to improve generalization performance across domains.
This approach is divided into supervised domain adaptation, which uses some labeled data from the target domain, and unsupervised domain adaptation which aligns feature distributions using unlabeled datasets.
Unsupervised domain adaptation began with the introduction of \textbf{Maximum Mean Discrepancy (MMD)}~\citep{MMD} -based methods and advanced with the emergence of \textbf{Generative Adversarial Networks (GANs)}~\citep{GAN}, which led to the development of adversarial domain adaptation techniques.
\textbf{Transfer Learning}~\citep{TransferLearning} refers to applying a model trained in one domain or task to a new dataset.
This includes Feature Transfer, which reuses or retrains specific layers of a pre-trained network for a new task, and Fine-Tuning which adjusts the entire pre-trained model to fit the new dataset.
Transfer learning originated from pre-trained embeddings in NLP in the late 1990s, expanded through CNNs, and eventually evolved with LLMs.
\textbf{Robustness Techniques} enhance a model’s ability to maintain performance despite data uncertainties, such as noise, outliers, and data scarcity.
Methods such as \textbf{Dropout}~\citep{Dropout} and \textbf{Data Augmentation}~\citep{DataAugmentation} help models handle noise, while robust loss functions like the \textbf{Huber loss}~\citep{HuberLoss} reduce sensitivity to outliers.
These techniques have been widely adopted in time series forecasting models since 2020.
}

This survey focuses on one prominent approach that introduces normalization and denormalization modules within a normalization-TSF model-denormalization framework. 
\revision{This framework has evolved as a specialized approach for mitigating non-stationarity within the same domain of time series data.}
Here, normalization is applied to the look-back window before feeding it into the TSF model, aiming to remove input statistics. As a result, the TSF model processes inputs from a less time-variant distribution. Denormalization is then applied to the forecasting window obtained from the TSF model to restore the original statistics. The denormalized forecasting window is ultimately used as the final forecast. However, calculating the appropriate statistics for normalization and denormalization in non-stationary scenarios is a non-trivial task, and numerous studies have been proposed within this framework to address this challenge.

\textbf{DAIN}~\citep{DAIN} introduces a Deep Adaptive Input Normalization layer that learns how much to shift and scale each observation end-to-end. Considering the changing distribution characteristics of time series data, this approach outperforms widely used normalization methods like batch normalization~\citep{BatchNorm} and instance normalization~\citep{InstanceNorm} across various domains, highlighting the importance of normalization techniques in TSF. However, DAIN omits a denormalization step, meaning it does not account for restoring statistics in the forecasting results. 
In contrast, \textbf{RevIN}~\citep{RevIN}, which extends instance normalization to be reversible, adopts a normalization-denormalization framework. RevIN performs instance normalization on each look-back window, followed by variable-wise multiplication of a learnable scale factor and addition of a bias factor. During de-normalization, the same parameters are applied in reverse. \textbf{NST}~\citep{nonstationaryTransformer} normalizes the look-back window in a non-parametric manner without a learnable affine transformation and then de-normalizes the prediction window using the mean and standard deviation of the look-back window. \textbf{Dish-TS}~\citep{Dish-TS} learns a module that predicts the statistics of the prediction window, considering the distribution shift between the look-back window and subsequent prediction windows, and performs normalization and de-normalization accordingly. \textbf{SAN}~\citep{SAN} considers the distribution shift within both the look-back window and prediction windows, performing statistical prediction on smaller slices.

\begin{table}[!h]
\centering

\captionsetup{justification=raggedright,singlelinecheck=false}
\caption{Normalization-Denormalization-based Approaches to Alleviate Distribution Shifts in Time Series Forecasting}
\label{tab:RecentNormalizationTSFModels}

\footnotesize 
\renewcommand{\arraystretch}{2.3} 
\arrayrulecolor[gray]{0.8} 

\begin{tabular}{!{\color{black}\vrule width 1pt}>{\centering\arraybackslash}m{0.09\textwidth}|>{\centering\arraybackslash}m{0.73\textwidth}|>{\centering\arraybackslash}m{0.12\textwidth}!{\color{black}\vrule width 1pt}}
\arrayrulecolor{black}\specialrule{1.0pt}{0pt}{0pt} 

\rowcolor[HTML]{808080}
{\color[HTML]{FFFFFF} \textbf{Model Name}} & 
\multicolumn{1}{c}{\cellcolor[HTML]{808080}\centering {\color[HTML]{FFFFFF} \textbf{Main Improvement \& Methodology}}} & 
\multicolumn{1}{c|}{\cellcolor[HTML]{808080}{\color[HTML]{FFFFFF} \textbf{Publication}}} \\

\arrayrulecolor[gray]{0.8} 

DAIN & \makecell[l]{· Introduce adaptive normalization \\ · Adaptive scaling, shift, and gating layers to normalize look-back window} & 2019 \\ \hline
RevIN & \makecell[l]{· Introduce normalization-denormalization framework \\ · Denormalize prediction with reversible affine transformation parameters} & 2022 \\ \hline
NST & \makecell[l]{· Non-parametric normalization-denormalization \\ · Normalize and denormalize using mean and std of look-back window without learnable \\ \ \ parameters} & 2022 \\ \hline
Dish-TS & \makecell[l]{· Divides distribution shifts: \\ \hspace{0.2cm} 1) within look-back window and 2) between look-back window and forecasting window \\ · Introduce Dual-CONET modules for statistics prediction} & 2023 \\ \hline
SAN & \makecell[l]{· Predict statistics in slice-level \\ · Slice forecasting windows and predict mean and std for each slice} & 2023 \\

\arrayrulecolor{black}\specialrule{1.0pt}{0pt}{0pt} 

\end{tabular}%

\end{table}
\subsection{Enhancing Causality}\label{sec5-3}

\paragraph{Why Causal Analysis is Essential for Accurate Time Series Forecasting}
Causal analysis is crucial in achieving accurate time series forecasting by better understanding the underlying factors driving data patterns. Time series data often exhibit correlations that can be misleading without a proper understanding of causality. For instance, two variables may show a strong correlation simply due to coincidence or because they are both influenced by a third, unobserved variable. Without distinguishing between correlation and causation, forecasting models risk attributing changes in the data to irrelevant or spurious factors, leading to inaccurate predictions. Causal analysis helps overcome this by identifying the true cause-and-effect relationships, ensuring the model is grounded in reality rather than coincidental patterns.

Moreover, causal analysis enhances the interpretability and practical application of forecasting models. By explaining how different variables influence the outcome, causal models \revision{uncover the underlying mechanisms} driving the forecast, which is crucial for decision-making. Businesses and policymakers can use these insights to predict the impact of specific actions, such as policy changes or marketing strategies, and make informed decisions based on the likely effects. This ability to simulate interventions and conduct counterfactual scenarios makes causal analysis an indispensable tool for accurate and actionable time series forecasting.

\paragraph{Research on TSF with Causality}
As mentioned earlier, utilizing causality in time series forecasting not only improves prediction accuracy but also enhances model interpretability. For these reasons, research applying causal discovery information to time series forecasting models is actively ongoing. In particular, various methodologies have been proposed in fields such as healthcare, environmental science, and social sciences, where causal discovery is actively researched.

\revision{In TSF, various causal inference methods are employed to identify cause-and-effect relationships within the data. The \textbf{Granger Causality Test} \citep{Granger} examines whether the historical information of one variable helps predict the future of another variable. This method detects the direction of causal influence based on regression analysis. However, it has limitations in fully excluding indirect correlations caused by the presence of a third variable. \textbf{Structural Causal Models (SCM)} \citep{SCM} utilize causal graphs and structural equations to model the relationships between variables. These models enable intervention simulations and counterfactual analyses, providing a visual representation of causal relationships to aid interpretation and integrate interactions across multiple variables. \textbf{Do-Calculus} \citep{Do-Calculus} is an intervention analysis technique for quantitatively analyzing the effects of interventions by computing the impact that changes in specific variables have on others, thereby supporting causal predictions. \textbf{Propensity Score Matching (PSM)} \citep{PSM} is a method for performing causal inference by matching groups with similar characteristics. This approach minimizes the influence of confounders and more accurately evaluates the effects of interventions. \textbf{Directed Acyclic Graphs (DAGs)} \citep{DAGs} are directional, non-cyclic graphs that visually represent the relationships between causes and effects. DAGs help clearly identify causal structures and understand complex interactions among variables. Various attempts are being made to enhance the causality of TSF by utilizing these diverse causal inference methods \citep{runge2019detecting, scholkopf2021toward}.
}

\cite{qian2023causality} introduce a model to predict Kuroshio Volume Transport (KVT). It employs multivariate causal analysis to discover causal relationships and selects only the variables with causal relationships to make predictions using an LSTM model. This approach captures meaningful information from causally related variables and prevents the model from being confused by unrelated variables.
\cite{mu2023nao} propose a model for predicting the North Atlantic Oscillation. It uses information obtained through causal discovery not only for variable selection but also by directly applying it to a GCN (Graph Convolutional Network). The overall structure utilizes a symmetrical encoder and decoder of ConvLSTM, with the GCN acting as a coupler in between.
\cite{sharma2022incorporating} focuses on a model for energy consumption. It applies the Granger causality test to determine the causal relationship between weather conditions and energy consumption and then integrates this causal information into a Bi-LSTM to improve energy consumption prediction accuracy.
\textbf{CausalGNN} \citep{wang2022causalgnn} is a model aimed at epidemic forecasting. It extracts causal features using the SIRD model and incorporates them into an Attention-Based Dynamic GNN module to learn spatio-temporal patterns.
\textbf{Caformer} \citep{zhang2024caformer} criticizes existing methods for failing to learn causal relationships effectively due to false correlations caused by environmental factors. To address this, it explicitly models environmental influences and removes them to obtain reliable causal relationships.

\subsection{Time Series Feature Extraction}\label{sec5-4}
Time Series Feature Extraction is the process of extracting useful information from time series data to enhance model performance. In other words, it involves identifying and quantifying time series data into key patterns, trends, seasonality, outliers, periodicity, and statistical characteristics to transform it into a form that is suitable for model training.
This process clarifies the unique patterns or characteristics of time series data, allowing models to learn them more effectively.
This enhances the understanding of data, improves the predictive performance of models, and increases efficiency through data compression. The reasons why Time Series Feature Extraction is necessary are as follows:
\begin{itemize}
    \item \textbf{Understanding the characteristics of data}\\ Time series data, which is ordered sequentially over time, possesses unique characteristics such as various patterns, seasonality, and periodicity.
    However, these data are complex and high-dimensional, making these features not easily apparent.
    The continuous values in time series data tend to be similar to adjacent values, which makes it challenging to accurately distinguish the context and meaning of each data point, often resulting in a lack of semantic information.
    Therefore, the process of transforming time series data into an analyzable format and extracting key features is crucial for understanding the inherent behaviors and interactions within the data.
    \item \textbf{Explainability of data}\\ Unlike traditional machine learning methods, deep learning models automatically learn useful features from data without requiring manual feature extraction. However, due to the nonlinear structure of deep learning models, it is often difficult for humans to interpret the learned features. The extracted features can better explain the structural characteristics of the data, helping to interpret and understand its meaning. This process provides critical insights for making data-driven decisions and contributes to improving explainability in complex model architectures.
    \item \textbf{Enhancing Model Performance}\\ By summarizing complex time series data and removing noise, it emphasizes the essential signals of the data, reducing the computational burden on models and improving predictive performance. Focusing on critical information allows the model to avoid overfitting to the training data and develop strong generalization capabilities.
\end{itemize}

\begin{figure}[H]
\centering
\includegraphics[width=1.0\textwidth, trim=0cm 1cm 0cm 1cm, clip]{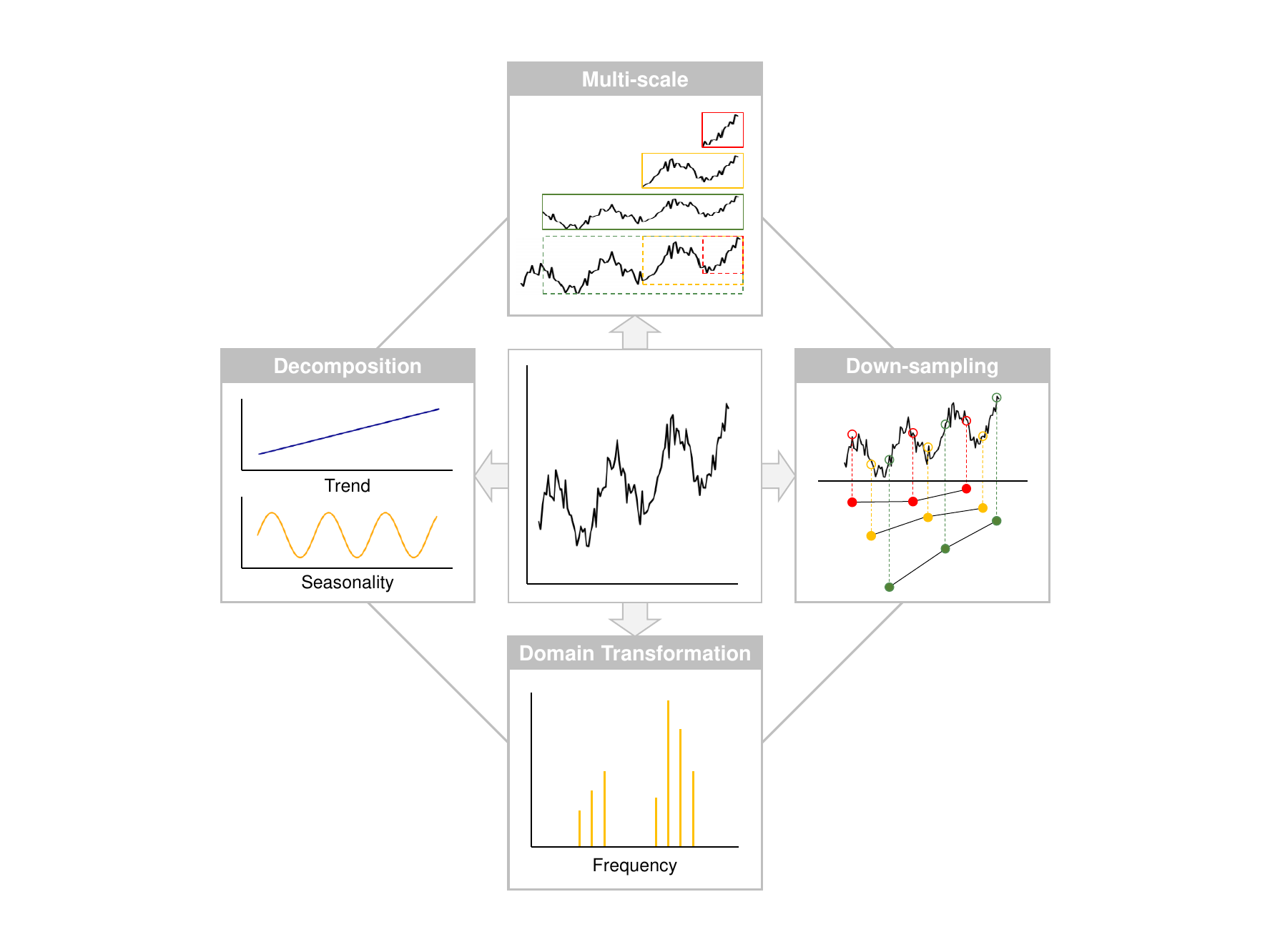}
\caption{Key Techniques in Time Series Feature Extraction}
\label{Fig_FeatureExtraction}
\end{figure}

Decomposing or transforming time series data into different forms before feeding it into the model is a widely used and researched approach for feature extraction.
These approaches help the model focus on learning the critical information in the data, thereby reducing sensitivity to noise and overfitting.
These methods are not limited to specific architectures and can be applied across various models.
Fig. \ref{Fig_FeatureExtraction} presents the key techniques for time series feature extraction, followed by an explanation of related application models for these approaches.

\subsubsection{Decomposition}\label{sec5-4-1}
Time series decomposition has long been used as a fundamental time series feature extraction technique, which involves separating time series data into components such as trend, seasonality, periodicity, and residual.
The advantage of decomposition is that it simplifies complex time series data by breaking it down into understandable, independent components, enabling more accurate model predictions and easier analysis and interpretation.

\paragraph{Moving Average Kernel}
Many models apply a Moving Average kernel to the input sequence to separate trend and seasonality components.
In this process, high-frequency noise or short-term fluctuations are removed, allowing long-term trends to be more clearly captured and overall patterns of increase and decrease to be emphasized.
The method of using a Moving Average Kernel is widely adopted due to its computational simplicity and efficiency. However, it has limitations when dealing with complex nonlinear patterns.

\textbf{Autoformer} \citep{Autoformer} goes beyond using decomposition techniques merely as preprocessing for forecasting tasks by progressively decomposing time series data throughout the prediction process within the model itself.
Similarly, \textbf{CrossWaveNet} \citep{CrossWaveNet} employs a dual-channel network to perform gradual deep cross-decomposition, enabling it to capture complex temporal patterns effectively.
Likewise, models such as \textbf{FEDformer} \citep{Fedformer}, \textbf{LTSF-Linear} \citep{DLinear}, and \textbf{PDMLP} \citep{PDMLP} utilize a moving average kernel to individually model each component of the time series data, which are then recombined to make effective predictions.
However, the Moving Average method lacks robustness because it is not learnable by the model.
Additionally, since it assigns equal weights to each data point within the sliding window, it has limited ability to identify specific patterns.
To address these limitations, \textbf{Leddam} \citep{Leddam}  use learnable 1D convolutional kernels, which can better capture nonlinear structures and dynamic variations.
Meanwhile, diffusion-based models face challenges as components like trend and seasonality can easily collapse during the diffusion process.
\textbf{Diffusion-TS} \citep{Diffusion-ts}  overcomes this by applying the diffusion process after decomposition, thereby preserving the characteristics of each component more effectively.
Through this approach, the traditional limitations of time series decomposition are overcome, allowing for a more effective separation of trend and seasonal components.

\paragraph{Downsampling}
The technique of using downsampling for decomposition is also frequently employed. 
Originating from signal processing, downsampling involves reducing the number of samples from the original signal, effectively decreasing the data by a specific ratio.
Time series forecasting is typically achieved through pooling methods, where representative values are extracted from specific segments of the time series data, thereby suppressing high-frequency components and emphasizing low-frequency components.
By adjusting the pooling size, various downsampling levels can be achieved, allowing for the exploration of multiple patterns within the data.

\textbf{SparseTSF} \citep{SparseTSF} and \textbf{SutraNets} \citep{SutraNets} use downsampling to separate data into trend and periodic components, predicting each subseries independently.
This allows the model to learn each component separately and better understand the influence of each on the complex time series data, enhancing prediction accuracy.
By dividing the complex time series data into various sub-series, it can be decomposed into a simpler form, thereby reducing the overall complexity of the model.
Therefore, this approach enhances the model's generalization ability and helps prevent overfitting.

\paragraph{\revision{Non-linear Methods}}
\revision{
While Moving Average Kernel and Downsampling are linear methods, \textbf{Empirical Mode Decomposition (EMD)} \citep{EMD} is a non-linear and non-stationary signal analysis technique that decomposes time series data into a number of Intrinsic Mode Functions (IMFs).
EMD involves finding local extrema in the signal and using linear interpolation for the upper and lower envelopes, and by subtracting their average, it extracts high-frequency components as IMFs iteratively.
The final residual signal, also known as the residue, reveals the long-term trend.
EMD is advantageous for processing non-linear and non-stationary time series data due to its adaptive nature, effectively separating and analyzing complex patterns and changes in the time series.
\textbf{EMD-BI-LSTM} \citep{EMD-BI-LSTM} proposes a method that combines EMD with a conventional Bi-LSTM model for forecasting electric load time series data.
By applying EMD, the model effectively captures complex non-linear relationships, thereby enhancing forecasting performance.
}

\revision{
However, EMD has a mode mixing problem, where signals of different frequencies are mixed within the same IMF \citep{EmdModeMixing}.
In the signal decomposition process of EMD, IMF is generated based on local extrema, which causes difficulties in independently separating components for signals with sudden changes, such as intermittent signals.
This leads to a failure in frequency separation, which not only reduces prediction performance but also impairs interpretability.
\textbf{Ensemble Empirical Mode Decomposition (EEMD)} \citep{EEMD} addresses this mode mixing problem by adding noise to the original signal and repeating the decomposition process, averaging the results.
Through this process, signals with different scales are properly aligned into the correct IMFs.
Through successive iterations, the influence of the noise is reduced, and the components of the original signal are emphasized, providing more physically meaningful results and enabling more accurate and stable time series decomposition.
\cite{EEMD_app} proposes models applying EEMD to LSTM, linear regression, and Bayesian ridge regression for forecasting.}

\revision{\textbf{Variational Mode Decomposition (VMD)} \citep{VMD} is a method that uses variational optimization to separate signals into modes with specific frequency bands.
Each mode’s frequency center and bandwidth are determined through variational optimization.
As the frequency ranges do not overlap, the separation is clearer and more stable.
However, compared to EMD, which requires relatively simple calculations, VMD is computationally more complex due to its use of variational optimization.
\cite{VMD_app} proposes a new model for crude oil price forecasting that applies VMD.
By decomposing the time series data with VMD, transforming it into images, and extracting features with CNN, this method improves forecasting performance for oil prices.
It effectively utilizes the advantages of VMD by making predictions with Bidirectional Gated Recurrent Unit (BGRU).
}

\revision{Various decomposition methods are widely used in time series forecasting.
However, decomposition methods also have some drawbacks.
When the data is divided into separate components based on different rules, the model predicts each component independently.
As a result, important interactions or relationships may not be considered, which can lead to underutilization of necessary information for prediction and degrade performance.
Nevertheless, considering the characteristics of time series, such as periodicity, trends, and noise, decomposition remains a powerful technique.
Moreover, since time series data is more difficult to intuitively interpret compared to other data types, decomposition provides interpretability, which helps increase the model’s trustworthiness.}

\subsubsection{Multi-scale}
The multi-scale approach analyzes data at various time scales, capturing patterns and trends at multiple levels simultaneously. This method is beneficial not only for time series data but also in fields like computer vision and natural language processing \citep{fan2021multiscale, nawrot2021hierarchical}. By integrating information from longer time frames, the multi-scale approach plays a significant role in improving performance in time series forecasting.

\textbf{MTST} \citep{MTST}, \textbf{PDMLP} \citep{PDMLP}, and \textbf{FTMixer} \citep{FTMixer} extend the patching application to a multi-scale framework, where shorter patches effectively learn high-frequency patterns and longer patches capture long-term trends.
\textbf{TimeMixer} \citep{TimeMixer} and \textbf{AMD} \citep{AMD} leverage multi-scale decomposition to capitalize on the strengths of each scale, enabling the model to make more accurate predictions.
\textbf{HD-TTS} \citep{HD-TTS} uses spatiotemporal downsampling to decompose data into various scales across both time and space.
\textbf{Scaleformer} \citep{scaleformer} and \textbf{Pathformer} \citep{pathformer} apply the multi-scale concept to the overall model architecture, while HD-TTS hierarchically implements temporal and spatial downsampling.
\textbf{MG-TSD} \citep{MG-TSD} and \textbf{mr-Diff} \citep{mr-Diff} successfully integrate multi-scale by incorporating inductive biases that prioritize generating coarse data in the early stages of the diffusion reverse process.

\subsubsection{Domain transformation}
A common method for enhancing the expression of latent periodic features in time series data is to transform the data into the frequency domain.
Techniques such as Fourier transformation, Wavelet transformation, and Cosine transformation are commonly used for this purpose.
Generally, time series data exhibits periodicity, which refers to patterns that repeat at regular intervals.
Frequency transformations effectively explore this periodicity.
In time series analysis, frequency transformations are used with two main approaches: extracting periodic components and direct learning in the frequency domain.

\paragraph{Periodicity Extraction}
Periodicity in time series data serves as critical information for predictive models, where features that may be hidden in the time domain can be more easily uncovered in the frequency domain.
Frequency transformations are advantageous because they can remove high-frequency noise while retaining important low-frequency components, thereby improving data quality.
By selectively extracting key frequency components and feeding them into the model, the process helps in learning essential patterns, enhances computational efficiency, and reduces model complexity.
Some models leverage these characteristics to identify and extract various periodic patterns in the data.
These adaptively extracted patterns are used either as inputs for the model or integrated into the learning process, thereby enhancing the model's predictive performance.

\textbf{Autoformer} \citep{Autoformer} utilizes periodicity extracted through auto-correlation within the Attention mechanism.
\textbf{TimesNet} \citep{Timesnet} transforms periodic components into a 2D format to train the time series with CNNs.
\textbf{MSGNet} \citep{MSGNet} uses periodic components to allow the model to determine appropriate scale levels for multi-scale analysis autonomously.
However, since the periodicity extracted in this manner is based on selective sampling, it does not encompass all the latent information, and if critical information is not selected, performance degradation becomes inevitable.

\paragraph{Training in the Frequency Domain}
The approach of directly training models in the frequency domain is widely studied as it overcomes these limitations by evenly learning all latent frequency components, ensuring that critical information is not missed.
While time series data can exhibit complex and varied patterns in the time domain, these patterns can often be succinctly represented by a few dominant frequency components when transformed into the frequency domain \citep{Fedformer}.
This simplification makes the learning process more straightforward, as most of the information can be captured with a minimal set of frequency components.
This method allows for an easy transition to the frequency domain while preserving the original information, thanks to various transformation and inverse transformation techniques.
When time series data exhibit nonlinear characteristics, learning in the frequency domain can better capture these complex patterns than in the time domain.
The frequency domain approach doesn't require consideration of time-axis variations, enabling stable performance even with non-stationary data.
Furthermore, this method allows for efficient learning by compressing the data without losing important characteristics.

In the frequency domain, each frequency component is represented as a complex number, consisting of a real part and an imaginary part, each conveying different information.
The real part is related to the magnitude (amplitude) of the time series data, indicating how prominent the periodic components are.
The imaginary part, on the other hand, relates to the phase information of the data, determining the temporal positioning of the frequency components.
By simultaneously understanding the amplitude and phase information in the frequency domain, models can accurately capture and predict complex periodic patterns in the data.
The research initially focused on simply changing the domain, but more recently, it has been moving towards expanding the expressiveness of the frequency domain.

\textbf{FreTS} \citep{FreTS} and \cite{DiffusionInfrequencyDomain} transform data into the frequency domain before feeding it into the model and then convert the predicted values back into the time domain.
Simply converting data to the frequency domain has limitations in achieving good model performance, leading to the development of models that overcome these challenges. Research continues to expand the representation in the frequency domain to fully utilize its rich features. \textbf{FEDformer} \citep{Fedformer} applies season-trend decomposition and then performs attention in the frequency domain, facilitating independent modeling of key information components. \textbf{Fredformer} \citep{fredformer} not only transforms data into the frequency domain but also investigates frequency bias through experiments, addressing issues in the frequency domain using frequency normalization techniques.
\textbf{FITS} \citep{FITS} utilizes distinct complex-valued linear layers in the frequency domain to learn amplitude scaling and phase changes, thereby enhancing the frequency representation of input time series through interpolation learning. \textbf{DERITS} \citep{DERITS} effectively handles the non-stationarity of time series data by differentiating frequency components and representing them in a static form that is easier to predict. \textbf{SiMBA} \citep{SiMBA} transforms time series data into the frequency domain using Fourier transforms, learning real and imaginary components separately, thus allowing more precise analysis of data in the frequency domain and improving model prediction performance. While frequency bands capture global dependencies well, they often struggle with local dependency capture. To overcome this, \textbf{WaveForM} \citep{WaveForM} uses discrete wavelet transforms (DWT) to decompose time series data into various frequency bands while preserving the time information of each band. This approach captures frequency changes within specific time intervals, simultaneously capturing features from both time and frequency domains. \textbf{FTMixer} \citep{FTMixer} proposes a method that directly utilizes data from both domains to leverage the strengths of global dependency extraction in the frequency domain and local dependency extraction in the time domain.

\subsubsection{Additional Approach}
In addition to the methods previously described, various other approaches have been explored.
A notable example is the use of High-Dimensional Embedding, which generates high-dimensional representations to better capture the essential information of the data.
This approach extracts complex, multidimensional information from the original time series data and integrates it to enhance the model's predictive capabilities.
For instance, \textbf{CATS} \citep{CATS} creates a new Auxiliary Time Series (ATS) by combining variables from the input data and then utilizes these for prediction.
\textbf{SOFTS} \citep{SOFTS} focuses on extracting common features (Core) from the variables and incorporating them into the learning process.
\revision{Furthermore, \textbf{BSA} \citep{2024BSA} enhances feature representation by leveraging temporal correlations between continuously streaming input samples, rather than depending solely on a single current input.
The method accumulates multiple scales of feature transformations over time via an exponential moving average (EMA), then applies Spectral Attention to these multi-scale momentums to capture long-range dependencies and enrich the resulting features.}

Efforts to effectively represent the complex features of time series data for better model learning are being explored in various forms.
These efforts are essential research topics for improving performance in the challenging field of time series forecasting.
Additionally, integrating these feature extraction techniques appropriately within the basic structure of existing deep learning models can create synergy, enhancing the model's predictive accuracy.

\subsubsection{\revision{Automated Feature Engineering and Self-Supervised Learning}}
\revision{In TSF, extracting meaningful features has traditionally required domain expertise and extensive manual work.
However, such approaches are often time-consuming and costly, and they may miss critical patterns when dealing with complex data.
To address these challenges, automated feature engineering has become increasingly important for automatically identifying key patterns in the data. 
One of the core techniques for automated feature engineering is self-supervised learning \citep{zhang2024self}.
}

\revision{Self-supervised learning is a pretraining approach that enables models to automatically learn important features from unlabeled datasets by identifying patterns directly from the data itself.
This allows the model to capture structural patterns without relying on manual feature extraction, reducing the dependency on labeled data.
Key approaches in self-supervised learning for TSF are as follows.
\textbf{Transformation-based approach} learns key features by comparing patterns before and after data transformations, such as time shifts and scaling.
For example, \textbf{TS-TCC} \citep{TS-TCC} enhances features by applying various transformations and learning the pattern similarities between the transformed data.
\textbf{Time Correlation Learning} extracts temporal dependencies in time series data by learning the relationships between past and future time points.
 \textbf{Contrastive Predictive Coding (CPC)} \citep{CPC} divides the data into multiple time segments and learns the correlations between these segments.
\textbf{Mask-based Learning} involves masking specific time points in the time series data and predicting these masked points based on the remaining data to learn key patterns \citep{TimeBERT}.
The application of such self-supervised learning approaches can enhance the performance of TSF \citep{SignalMix, Timesnet}.
}

\revision{Self-supervised learning enables models to automatically recognize and extract key patterns and nonlinear relationships from data without labeled examples.
It can effectively learn from small datasets, addressing the issue of label scarcity and contributing to model lightweight and algorithm optimization.
In particular, multimodal self-supervised learning techniques, which integrate time series data with other data types such as images and text, are expected to advance, enabling the model to learn richer patterns.
Such multimodal learning can significantly contribute to solving complex decision-making problems, such as healthcare diagnostics and disaster prediction.
Additionally, research on converting the learned features into interpretable forms will become increasingly important.
This will enhance the interpretability and reliability of the model and improve trust in data-driven decision-making processes.
}

\revision{In conclusion, self-supervised learning-based automated feature engineering is a pivotal research area for effectively understanding complex patterns in time series data and enhancing predictive performance.
Research combining this approach with data efficiency improvements, multimodal learning, and explainable AI (XAI) is expected to further increase the practicality and reliability of TSF.
}

\subsection{\revision{Model Combination Techniques}}\label{sec5-5}
\revision{The instability of single models in time series forecasting has been a persistent challenge from the past to the present. Due to the complexity, volatility, and distinguishing characteristics of time series data, the use of a single model can lead to overfitting and unstable forecasting performance. Therefore, methods for combining models offer a simple yet effective approach to overcome these issues. Model combination techniques reduce forecasting uncertainty and enable more robust predictions.
\cite{makridakis2020m4} experimentally demonstrates that model combination, as applied in the M4 Competition task, improves forecast accuracy over single models or model selection. However, Model combination techniques can face issues such as model redundancy, increased computational costs, and reduced interpretability due to the complexity of the combined models. To maximize the combination effect, it is necessary to properly leverage model diversity or use sophisticated combination methods.}

\subsubsection{\revision{Ensemble Models}}
\begin{figure}[H]
\centering
\includegraphics[width=1.0\textwidth, trim=0cm 0.5cm 0cm 0.5cm, clip]{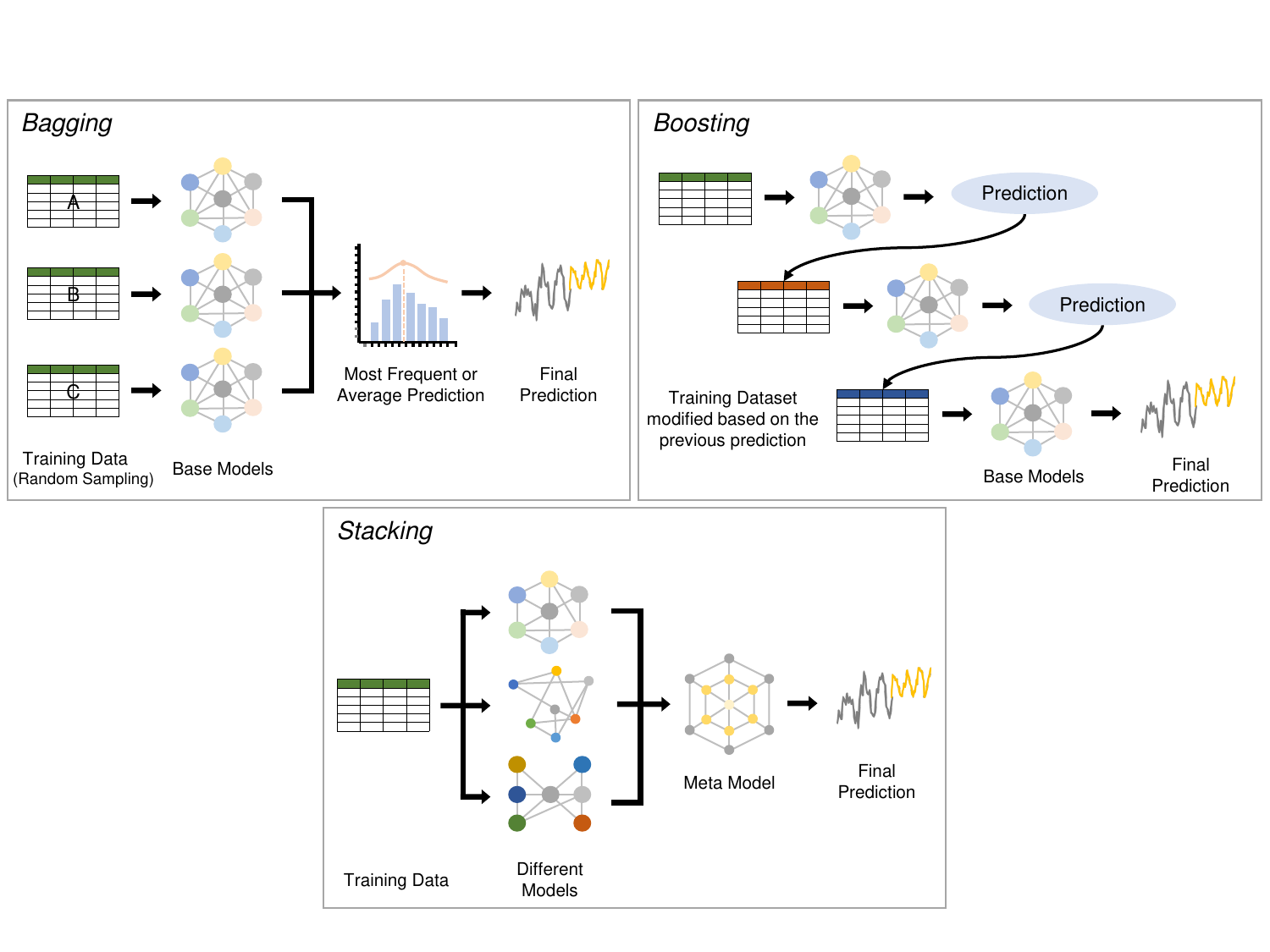}
\caption{\revisionB{Key Ensemble Techniques}}
\label{Fig_Ensemble}
\end{figure}

\revision{Ensemble models combine the independent predictions of several models to complement the weaknesses of individual models and generate a final forecast.
\cite{EnsembleDL} summarizes the theoretical aspects of ensemble learning's success based on existing research, arguing that combining predictions from multiple models is an effective way to improve model performance. This section explores key ensemble methods, including bagging, boosting, and stacking, as illustrated in Fig. \ref{Fig_Ensemble}.}

\revision{\textbf{Bootstrap Aggregating (Bagging)} involves randomly sampling the data to train several independent models and then combining their results through averaging or voting \citep{BaggingPredictors}. \cite{whyBaggingForTS} explains why bagging is effective for time series forecasting by addressing data uncertainty, model uncertainty, and parameter uncertainty, improving forecasting performance. \cite{kim2022bagging} combines Wavelet transforms with bagging techniques in MLP to improve univariate time series forecasting performance. Using the Maximum Overlap Discrete Wavelet Transform (MODWT) \citep{nason2000wavelet}, the signal is decomposed, and bagging is applied only to the detail part, excluding trends. This creates data diversity by preserving the main trend while altering the detailed fluctuations.}

\revision{\textbf{Boosting} overcomes the issue of model diversity by sequential training and combining weak learners \citep{freund1997decision}. This approach addresses the diversity problem that can arise in bagging when each model is trained in a very similar manner. Each model places more weight on the data that was incorrectly predicted by the previous models, aiming to reduce bias. This creates a strong learner by combining weak learners and ensures added diversity. Gradient boosting is an algorithm that optimizes models using the gradient of the loss function within the boosting framework \citep{friedman2001greedy}. XGBoost (eXtreme Gradient Boosting) executes the gradient boosting algorithm in parallel, making it faster and more effective in handling sparse data while improving learning speed and predictive performance \citep{xgboost}. In the 2015 Kaggle \href{https://www.kaggle.com}{(https://www.kaggle.com)} machine learning competition, 17 out of 29 winning solutions used XGBoost \citep{xgboost}. As deep learning advanced, various works combining DL with XGBoost emerged. LightGBM \citep{ke2017lightgbm} is a boosting-based algorithm with a leaf-wise tree splitting method, and \cite{ensemble_stock} uses an ensemble approach that combines traditional and advanced methods for stock prediction. Specifically, the integration of LSTM, GRU, LR, and LightGBM models enhances forecasting performance.
However, boosting is generally not applied to deep learning models, as boosting typically uses simple weak learners that are repeatedly trained to improve performance. Deep learning models like MLP already learn complex patterns effectively with large amounts of data, so the need for iterative error correction is minimal. Thus, simpler machine learning methods are typically used to construct weak learners instead of deep neural networks.
\cite{ltboost} proposes predicting trends using linear models and then applying boosting to the residuals with tree-based models for forecasting. This model combines the simple trend estimation capability of linear models with the nonlinear pattern learning ability of tree-based models, achieving more precise and effective predictions in LTSF.
However, there is also a case where boosting principles are applied to deep learning models, such as \cite{minusformer}, which integrates boosting's progressive residual learning approach into deep learning to enhance time series forecasting.}

\revision{\textbf{Stacking} combines the predictions of several models to generate a final forecast \citep{wolpert1992stacked}. Stacked Generalization involves training several base learners, and then feeding their prediction results into a meta-model to generate the final prediction. Unlike bagging and boosting, stacking does not directly combine the predictions of the models; instead, it inputs the prediction results into a new model to make the final forecast. Stacking is distinctive in that it incorporates model weightings and analysis, enabling more refined and comprehensive predictions \citep{stackedEnsemble}.
\cite{LGBM-XGB-MLP} suggests stacking LightGBM and eXtreme Gradient Boosting (XGB) to tackle stochastic variations in short-term load forecasting, where the meta data generated by both models is input to an MLP for the final prediction.
Furthermore, the use of MLP as a meta-model instead of traditional ML models like linear regression enables the learning of more complex, nonlinear relationships, improving prediction performance. The results demonstrate the model's generalization ability and superior performance across various datasets.
\revisionB{Stacking can be applied not only to deterministic forecasting but also to probabilistic forecasting, and some studies have shown that stacking is effective in handling uncertainty \citep{probstacking1, probstacking2}.}
\cite{theoreticalEnsemble} explores the theoretical aspects of stacked generalization. Through cross-validation, it provides theoretical guarantees that the performance of a stacked generalization model selected through cross-validation does not fall significantly behind that of an oracle model, the best performing individual model. The paper also explains that extending the basic models used in stacking to learnable models, rather than constant models, provides better theoretical guarantees.}

\revision{Ensemble models require significant training resources since each model must be trained independently. Nevertheless, ensemble techniques combine diverse, high-performance models to complement each other and can be implemented in a relatively simple manner. Due to these characteristics, ensemble learning can enhance prediction stability and generalization capabilities in application fields, penetrating nearly all industries, from grain product distribution, semiconductor manufacturing, and sensor design to commercial software development and testing services \citep{ensemble_app}.}

\subsubsection{\revision{Hybrid Models}}
\revision{
Hybrid models combine different types of models or techniques to leverage the strengths of individual models while compensating for their weaknesses.
These models employ a more sophisticated structural integration approach compared to ensembles, encompassing various strategies such as combining statistical models, machine learning models, and deep learning models.
}

\revision{
\textbf{ESRNN} \citep{ESRNN}, a hybrid model that combines statistical models with deep learning architectures, integrates exponential smoothing (ES) and LSTM to leverage the interpretability of statistical methods alongside the powerful nonlinear learning capabilities of deep learning. Through ES, the model effectively captures key components of individual time series, such as level, trend, and seasonality, while LSTM learns the correlations and nonlinear patterns across multiple time series. Additionally, the use of dynamic computation graphs and a hierarchical design provides strong performance and interpretability in real-world applications. This model highlights the potential of hybrid models for time series forecasting with its superior performance. 
\revisionB{Moreover, AQ-ESRNN \citep{smyl2024any} model, an extension beyond deterministic forecasting to probabilistic forecasting, enables distributional forecasting by quantifying prediction uncertainty.}
Subsequently, hybrid models combining deep learning architectures have advanced, and this section focuses on explaining such combinations of deep learning architectures.
}

Particularly, time series data often involve not only temporal dependencies but also important interactions between variables. Therefore, many models explicitly learn the interdependencies between variables in addition to temporal dependencies. However, incorporating all this information into a single architecture can lead to overfitting, which may limit the learning process. Recent models often combine two or more architectures to efficiently learn by distinguishing the roles based on the nature of the information.

\textbf{WaveForM} \citep{WaveForM} improves prediction performance by transforming time series data into the wavelet domain and combining it with GNN. The Discrete Wavelet Transformation (DWT) uses High-Pass and Low-Pass Filters to decompose the time series data into small wavelets. This process removes noise and allows CNN to efficiently capture temporal and frequency characteristics. The GNN is then used to model the interactions between these components.
\textbf{TSLANet} \citep{TSLANet} is a model that replaces the computationally expensive attention mechanism with CNN in the Transformer framework. It incorporates frequency analysis and a learnable threshold to selectively attenuate high-frequency noise, effectively learning long-term and short-term interactions. Additionally, it uses parallel convolution layers with different kernel sizes to capture both local patterns and long-term dependencies. This approach effectively compensates for the typical CNN's limitations in modeling long-term dependencies.
\textbf{DERITS} \citep{DERITS} enhances the prediction performance of non-stationary time series data by combining convolution operations in the frequency domain with MLP in the time domain. The Frequency Derivative Transformation (FDT) converts time series signals into the frequency domain and then differentiates the frequency components to represent them in a static form that is easier to predict. The Order-adaptive Fourier Convolution Network (OFCN) is responsible for frequency filtering and learning dependencies in the frequency domain. To improve prediction performance, it processes and fuses derivative information of various orders in parallel, utilizing a Parallel-stacked Architecture.
\textbf{BiTGraph} \citep{BiTGraph} focuses on improving performance in situations with missing data by capturing both temporal dependencies and spatial structures. The Multi-Scale Instance Partial TCN module learns short-term and long-term dependencies through kernels of various sizes and can also compensate for missing values through partial TCN. The Biased GCN module for inter-channel relationship learning represents the relationships between data points as a graph structure and adjusts the strength of information propagation between nodes, considering the missing patterns.
\revision{
\cite{zhang2021hybrid} proposes a hybrid model for wind speed forecasting.
By performing decomposition using Singular Spectrum Analysis (SSA) and Multivariate Empirical Mode Decomposition (MEMD), the model removes noise and separates the trend and various frequency components of the time series data, enabling more accurate forecasting.
Then, by combining CNN with attention mechanism and Bidirectional Long Short-Term Memory, the model extracts spatial correlations and learns temporal dependencies, optimizing the wind speed forecasting performance.
\cite{zhang2021novel} also uses both ensemble and hybrid models for TSF.
The ensemble deep learning model, which combines various deep learning models such as CNN, MLP, and LSTM, demonstrates excellent performance in forecasting real-world time series data, including PM$_{2.5}$ concentration, wind speed, and electricity prices.
The model utilizes Extended AdaBoost \citep{freund1997decision} to generate diverse base predictors, followed by applying a stacking method to use Kernel Ridge Regression as a meta-predictor to produce the final forecast.
}

\revision{
In hybrid models, optimization processes such as hyperparameter tuning, regularization techniques, and cross-validation can help reduce the risk of overfitting caused by the interdependence and complexity between multiple models.
Additionally, evolutionary algorithms can be used to explore and optimize model structures \citep{wei2022npenas}.
Hybrid models may increase training time and resource consumption, and they can have limitations in interpretability and maintenance.
Therefore, techniques like model compression or feature selection can be applied, and model interpretability issues can be addressed using interpretation methods.
}

\subsection{\revision{Interpretability \& Explainability}}\label{sec5-6}
\revision{
Although AI models exhibit outstanding performance, their black-box nature often undermines trust in decision-making processes \citep{xua2024interpretability}.
In mission-critical domains, particularly those dealing with human life and safety, understanding the reasons behind model decisions is crucial, which has led to restrictions on the use of deep learning models \citep{papapetrou2024interpretable}.
The European Union, through the General Data Protection Regulation (GDPR), mandates the explanation of automated decisions, ensuring that individuals can understand the rationale behind decisions made by automated systems like AI models \citep{hamon2022bridging}.
As data-driven decision-making becomes more widespread, the need to understand how a model arrives at specific predictions and why such outcomes occur is growing \citep{TSFwithDL}.
}

\revision{
Interpretability and explainability aim to ensure model understanding and transparency, making it possible to know how and why a model produces certain results.
These two concepts can be applied in various ways.
For example, consumers might demand explanations when they notice anomalies in model outcomes, while AI researchers could use these concepts to improve model performance.
Ultimately, interpretability and explainability promote not only research on AI models but also their expansion into various fields, enhancing performance and usability, which could drive fundamental changes in industries and societal paradigms \citep{han2023impact, jung2020icaps, kim2020interpretation}.
}

\revision{
These concepts are particularly important in time series data, where interpretability helps in understanding how a model recognizes and predicts periodic and repetitive patterns.
Furthermore, explainability is necessary for anomaly detection models to clarify how anomalies are detected and which features are abnormal, thereby aiding decision-making.
}

\revision{
The machine learning community has noted that there is no universally agreed-upon definition for the terms interpretability and explainability \citep{lipton2018mythos}.
Consequently, in various studies, these terms are often used interchangeably, even though they carry different meanings and are used differently across domains \citep{graziani2023global}.
To avoid confusion for our readers, we define and explain these terms as follows.
\textbf{Interpretability} focuses on understanding the internal workings of a model, providing transparency about how the model makes predictions \citep{csahin2024unlocking}.
Interpretability requires an intuitive understanding of the model's structure, feature selection, and prediction process.
\textbf{Explainability}, on the other hand, is concerned with providing reasons for a model's outputs in a way that humans can comprehend.
It is a technical approach that helps make the decision-making process of the model more understandable.
In summary, interpretability is about understanding how the model works, while explainability is about providing understandable reasons for the model's predictions.
}


\subsubsection{\revision{Interpretability}}
\revision{Interpretability can be achieved through the following representative models.
\textbf{Temporal Fusion Transformer (TFT)} \citep{lim2021temporal} is a deep learning model for multi-horizon forecasting proposed by the Google Cloud AI team, which utilizes the attention mechanism to effectively learn the interactions between static and time-dependent input data.
This model incorporates a static feature encoder, a variable selection gating mechanism, and interpretable weights, allowing the model to better capture the characteristics of time series data and significantly improve interpretability compared to a standard Transformer.
The gating mechanism suppresses unnecessary inputs, enhancing model efficiency, while the static feature encoder converts static data into contextual vectors that are incorporated into predictions.
The following two models enhance the interpretability of the TFT model for wind speed prediction by applying a two-stage decomposition approach.
Given the high volatility and multi-resolution modes of wind speed data, the decomposition technique described in Section \ref{sec5-4-1} allows for the independent analysis of the impact of each component on the prediction results, thus improving interpretability.
\textbf{EMD-EEMD-JADE-TFT} decomposes the data using Empirical Mode Decomposition (EMD) and Ensemble Empirical Mode Decomposition (EEMD) and optimizes the TFT’s hyperparameters using Adaptive Differential Evolution with optional external archive (JADE) \citep{wu2024two}.
\textbf{IEEMD-EWT-JADE-TFT} model, on the other hand, uses Improved Complete Ensemble Empirical Mode Decomposition with Adaptive Noise (IEEMD) and Empirical Wavelet Transform (EWT) for decomposition, and similarly optimizes TFT hyperparameters with JADE \citep{wu2024interpretable}.
Additionally, meteorological feature engineering combines statistical features to enhance both performance and interpretability.}

\subsubsection{\revision{Explainability}}
\revision{Different methodologies exist for models focused on explainability.
Simple machine learning models, such as Linear Regression or Decision Trees, are transparent and provide an intuitive understanding of predictions by themselves.
These models have \textbf{intrinsic explainability}, which falls under ante-hoc explainability, meaning their predictions can be easily explained from the design stage.
On the other hand, deep learning models are highly complex and learn high-dimensional features, making it difficult to achieve intrinsic explainability through simple aggregation methods.
Therefore, \textbf{post-hoc explainability} techniques are primarily used for deep learning models.
Post-hoc explainability refers to approaches that interpret and explain the prediction or decision-making process after the model has already been trained.
Post-hoc interpretability can be divided into several subcategories.}

 \revision{\textbf{Surrogate models} are used to explain the predictions of complex models by training a simpler, interpretable model to explain the predictions.
Local Interpretable Model-Agnostic Explanations (LIME) is a representative surrogate model that provides local explanations for individual predictions \citep{ribeiro2016should}.
LIME works by generating perturbated sample data around a data point and obtaining predictions from the black-box model.
Based on this, a simple linear model is trained to identify the important features in that prediction.
\cite{yang2023investigating} proposes an improved LIME algorithm to address the interpretability issues of black-box models in wind power forecasting and applies it to a wind power prediction model.
This model introduces a trust index to quantify how reliable the features used by the prediction model are and analyzes how the black-box model handles important features in the prediction.
Additionally, this model proposes global trust modeling and interpretable feature selection methods, demonstrating that these approaches can enhance the model's reliability and reduce prediction errors.}

 \revision{\textbf{Feature attribution} is a method that calculates how much each input feature contributed to the model’s prediction and evaluates the contribution of input features for a specific prediction.
SHapley Additive exPlanations (SHAP) is a representative feature attribution technique based on the Shapley value concept from game theory \citep{lundberg2017unified}.
SHAP fairly measures the contribution of each feature to model predictions.
SHAP visualizes the importance of features from both global and local perspectives and helps identify interactions between model predictions and features.
The key idea behind SHAP is to compute the average contribution of each feature across all possible feature combinations, which allows the decomposition of predictions into feature contributions, providing intuitive and interpretable results.
\cite{garcia2020shapley} uses an LSTM model to predict NO$_2$ concentrations and presents a method for interpreting the prediction results based on SHAP.
While SHAP values are typically applied to linear models, this study extends SHAP by applying Deep SHAP to accommodate the nonlinear characteristics of deep learning models, making it easier to understand the complex decision-making process of deep learning models \citep{chen2022explaining}.}

 \revision{\textbf{Counterfactual explanations} describe what inputs need to be altered to change a specific prediction \citep{wachter2017counterfactual}.
Counterfactual explanations generate counterfactual inputs that would alter the prediction.
In time series research, counterfactual explanations have been widely used in classification tasks, but their application to forecasting models has been relatively underexplored \citep{wang2023counterfactual}.
In time series forecasting, exogenous variables make predictions more complex, which highlights the importance of setting range constraints \citep{papapetrou2024interpretable}.
ForecastCF is a model that generates counterfactual explanations to explain the predictions of time series forecasting models \citep{wang2023counterfactual}.
This model modifies the time series data through gradient-based perturbation to ensure the model’s predictions fall within a specified range, thus improving the model’s explainability.
Specifically, users define the desired prediction range, and the model generates predictions that satisfy these constraints, making the predictions more intuitive and understandable.}

\subsection{\revision{Spatio-temporal Time Series Forecasting}}\label{sec5-7}
\revision{As the structure of time series data becomes increasingly complex, spatial information is now frequently incorporated into the datasets.
Spatio-temporal time series data refers to datasets that simultaneously capture temporal variations and spatial distributions.
This type of data consists of values that change over time at specific locations, possessing both spatial interactions and temporal dependencies.
Due to these characteristics, it provides critical information not only from patterns extracted along the time axis but also from spatial dependencies caused by interactions between locations.
For example, region-based traffic flow, air pollution levels, climate change, and electricity consumption are representative cases.
The rapid advancements in technologies such as satellite data and GPS have led to an explosive increase in location-based data.
Therefore, spatio-temporal data modeling has become increasingly important, as many real-world problems cannot be adequately explained by temporal variations alone \citep{sun2024survey}.
Conventional multivariate time series models often fail to adequately address spatial correlations or network structures, necessitating their extension to spatio-temporal models.
To address this, numerous approaches utilizing GNNs and Transformers have been actively explored \citep{jin2023large, yang2024survey, chen2021nast, park2020st}. 
}

\revision{Several representative approaches for handling spatio-temporal data are as follows.
One is the Graph-Based Method for spatio-temporal modeling.
Data is represented as a graph, where spatial relationships between locations (nodes) are expressed as edges.
GCNs or graph-based RNNs are used to simultaneously learn spatial and temporal dependencies.
This approach effectively models non-euclidean structured data but has drawbacks, such as complex preprocessing and high computational costs.
These approaches are well-suited for applications such as traffic flow prediction by modeling road networks or electricity demand forecasting by modeling connections between substations.
\textbf{DCRNN} \citep{DCRNN} models traffic networks as directed graphs using Diffusion Convolution, effectively capturing spatial dependencies, including upstream and downstream node relationships.
It also leverages GRU to capture temporal patterns and non-linearities, making it a representative graph-based model for handling complex time series data.
\textbf{ST-GCN} \citep{ST-GCN} models skeleton dataset as graphs and extends GCN along the temporal axis to simultaneously model spatial and temporal dependencies.
\textbf{DB-STGCN} \citep{DB-STGCN} predicts train delays by integrating dynamic adjacency matrices generated through Dynamic Bayesian Networks (DBN) into ST-GCN to capture spatio-temporal dependencies.
It improves prediction accuracy and interpretability through the integration of attention mechanisms and the combination of GCN and GRU.
}

\revision{Another approach is the raster-based method.
Data is converted into a 2D grid (raster) or video format to learn spatial locations and temporal patterns.
Grid-based data representation: A specific space is divided into grids, and data over time is collected for each grid cell.
Video-based data representation: Data is transformed into a 3D tensor, including the time axis.
This format allows easy application of image processing techniques, such as CNNs, and 3D CNNs are particularly effective for directly learning spatio-temporal patterns.
However, there are limitations in learning overly complex spatial relationships.
They are suitable for modeling climate data from satellite imagery or recognizing actions in video data through spatiotemporal dynamics. 
\textbf{ConvLSTM} \citep{ConvLSTM} replaces the fully connected state transitions of conventional LSTM with convolutions to learn spatial correlations, making it a representative early raster-based method.
It represents all inputs, hidden states, and cell states as 3D tensors to maintain the grid-like spatial structure.
\textbf{ST-ResNet} \citep{ST-ResNet} models citywide crowd flows based on residual learning.
It partitions the city into a grid structure and represents the inflow and outflow of each grid cell as 2-channel image data.
Residual networks are employed to effectively learn interactions between both nearby and distant regions.
\textbf{SLCNN} \citep{SLCNN} utilizes Structure Learning Convolution (SLC) to define graphs as learnable parameters for the effective modeling of both global static structures and dynamic structures within datasets.
It replaces conventional CNN convolutions with SLC tailored to graph structures, generating graph-specific representations at each layer.
}

\revision{Recently, there have also been efforts to combine graph-based and raster-based methodologies to more effectively model both spatial structures and temporal dependencies.
They are ideal for representing traffic flow as graphs and enabling detailed analysis of each node in smart city data. 
\textbf{GMAN} \citep{GMAN} effectively learns the dynamic spatio-temporal relationships in traffic networks through Spatio-Temporal Embedding (STE) and ST-Attention Block.
It utilizes spatial attention to capture dynamic relationships between sensors, temporal attention to learn nonlinear temporal patterns, and gated fusion to integrate these features seamlessly.
\textbf{STSGCN} \citep{STSGCN} employs the Spatial-Temporal Synchronous Graph Convolutional Module (STSGCM) to synchronize and learn spatial and temporal dependencies.
It constructs a localized spatial-temporal graph to model relationships between neighboring nodes within the same time frame and nodes from past or future time frames simultaneously.
\textbf{GCN-SBULSTM} \citep{GCN-SBULSTM} proposes a model that combines GCN with Stacked Bidirectional Unidirectional LSTM (SBULSTM).
The Bidirectional LSTM captures temporal patterns, while GCN learns spatial dependencies, enabling integrated modeling of spatio-temporal interactions.
Additionally, sequence-based methods process spatio-temporal data in sequence form, focusing primarily on temporal dynamics. }

\revision{The primary objective of spatio-temporal time series prediction models is to address the challenging task of integrating complex data structures while simultaneously learning temporal dependencies and spatial interactions.
This enables accurate future state prediction and helps solve various real-world problems, facilitating precise decision-making.
}

\vspace{3ex}

\section{Conclusion}\label{sec6}

This survey has been prepared to broaden the expertise of existing researchers and to assist beginners in gaining a fundamental understanding amid the rapid growth of time series forecasting (TSF) research.
By integrating key concepts of time series and the latest techniques, we aim to provide researchers with clear direction and insights, fostering the continued advancement of this field.
This survey paper comprehensively reviews recent advancements in TSF, focusing on deep learning models.
\revision{It also incorporates key papers from major AI and ML conferences, as well as several notable papers from the arXiv preprint repository \href{https://arxiv.org}{(https://arxiv.org)}.
}

TSF is a highly regarded topic, gradually expanding from the prominence of Transformer models to various architecture-based models.
Fundamental deep learning models like MLPs, CNNs, RNNs, and GNNs are being reassessed, and Transformers are also advancing by overcoming their previous limitations.
\revision{Additionally, innovative models like Mamba are emerging, and models such as Diffusion are also entering the field, experiencing rapid growth.}
Furthermore, as the demand for foundation models in the field of TSF intensifies, pre-trained models based on large language models (LLMs) are emerging.
To create a powerful predictive model, researchers need to be aware of these trends and understand both the characteristics of time series data and the strengths and weaknesses of different architectures.

Taking it a step further, we aim to address common challenges in TSF and provide deep insights into future development directions.
In particular, we reviewed various studies that focus on approaches to channel correlation and address the challenges of distribution shift, which frequently occur in real-world scenarios.
Additionally, we emphasized the need for research that leverages causality to eliminate spurious correlations, thereby enabling a deeper understanding of the underlying essence.
\revision{We also discussed the importance of studies focused on effectively extracting features from complex time series data.
In addition, we explored model combination techniques aimed at enhancing model performance, as well as spatio-temporal TSF problems to address complex real-world issues. Furthermore, we provided insights into improving model reliability by addressing interpretability and explainability.}
Through this, we aim to present readers with potential directions for future research.

\paragraph{Limitations and Future Work}
This survey acknowledges several limitations that could be addressed in future research.
\begin{itemize}
    \item We skipped the detailed theoretical backgrounds of the models. This survey aims to provide a comprehensive overview, facilitating comparison and analysis, thereby allowing researchers to explore areas of interest more effectively. Although comprehensive information is primarily provided, readers can access additional details through reference links if needed.
    \item We left the specific differences in characteristics across various time series datasets for future work. Time series data is often domain-specific, requiring expert knowledge, which can be integrated into models to enhance performance. This underscores the importance of interdisciplinary collaboration, and future research should continue to develop in this direction. \revision{By integrating insights from a wide range of academic fields into TSF, it can contribute to the development of robust and generalized models.}
    \item \revision{The aspect of expanding Artificial General Intelligence (AGI) and Adaptive Modeling could be further explored. Research on techniques such as meta-learning, reinforcement learning, and neuro-symbolic AI can enhance the generalization and adaptability of models. In the current TSF field, time series-specific models and deep learning-based standard methodologies are dominant. However, as TSF problems become increasingly complex, research on generalization and adaptability holds great potential.} 
\end{itemize}

\bmhead{Acknowledgements}

We would like to express our heartfelt gratitude to Jisoo Mok, Hyeongrok Han, Bonggyun Kang, Junyong Ahn, Jiin Kim, Youngwoo Kimh, Hyungyu Lee, Juhyeon Shin, Jaihyun Lew, and Jieun Byeon from our research lab for their invaluable assistance in reviewing and providing feedback on this paper. Their insights and suggestions have significantly enhanced the quality of our work.

\clearpage
\bibliography{reference}

\begin{thebibliography}{312}
\providecommand{\natexlab}[1]{#1}
\providecommand{\url}[1]{{#1}}
\providecommand{\urlprefix}{URL }
\providecommand{\doi}[1]{\url{https://doi.org/#1}}
\providecommand{\eprint}[2][]{\url{#2}}
 \bibcommenthead

\bibitem[{Abu-Mostafa and Atiya(1996)}]{IntroductionToFinancialForecasting}
Abu-Mostafa YS, Atiya AF (1996) Introduction to financial forecasting. Applied intelligence 6:205--213

\bibitem[{Achiam et~al(2023)Achiam, Adler, Agarwal, Ahmad, Akkaya, Aleman, Almeida, Altenschmidt, Altman, Anadkat et~al}]{GPT4}
Achiam J, Adler S, Agarwal S, et~al (2023) Gpt-4 technical report. arXiv preprint arXiv:230308774

\bibitem[{Ahamed and Cheng(2024)}]{TimeMachine}
Ahamed MA, Cheng Q (2024) TimeMachine: A Time Series is Worth 4 Mambas for Long-Term Forecasting, IOS Press. \doi{10.3233/faia240677}, \urlprefix\url{http://dx.doi.org/10.3233/faia240677}

\bibitem[{Ahmed et~al(2022)Ahmed, Hassan, and Mstafa}]{ReviewOnDeepSequential}
Ahmed DM, Hassan MM, Mstafa RJ (2022) A review on deep sequential models for forecasting time series data. Applied Computational Intelligence and Soft Computing 2022(1):6596397

\bibitem[{Alcaraz and Strodthoff(2023)}]{SSSD}
Alcaraz JL, Strodthoff N (2023) Diffusion-based time series imputation and forecasting with structured state space models. Transactions on Machine Learning Research \urlprefix\url{https://openreview.net/forum?id=hHiIbk7ApW}

\bibitem[{Alghamdi et~al(2019)Alghamdi, Elgazzar, Bayoumi, Sharaf, and Shah}]{trafficForecasting}
Alghamdi T, Elgazzar K, Bayoumi M, et~al (2019) Forecasting traffic congestion using arima modeling. In: 2019 15th international wireless communications \& mobile computing conference (IWCMC), IEEE, pp 1227--1232

\bibitem[{Ansari et~al(2024)Ansari, Stella, Turkmen, Zhang, Mercado, Shen, Shchur, Rangapuram, Arango, Kapoor, Zschiegner, Maddix, Wang, Mahoney, Torkkola, Wilson, Bohlke-Schneider, and Wang}]{CHRONOS}
Ansari AF, Stella L, Turkmen AC, et~al (2024) Chronos: Learning the language of time series. Transactions on Machine Learning Research \urlprefix\url{https://openreview.net/forum?id=gerNCVqqtR}, expert Certification

\bibitem[{Athanasopoulos et~al(2011)Athanasopoulos, Hyndman, Song, and Wu}]{athanasopoulos2011tourism}
Athanasopoulos G, Hyndman RJ, Song H, et~al (2011) The tourism forecasting competition. International Journal of Forecasting 27(3):822--844

\bibitem[{Bai et~al(2018)Bai, Kolter, and Koltun}]{TCN}
Bai S, Kolter JZ, Koltun V (2018) An empirical evaluation of generic convolutional and recurrent networks for sequence modeling. arXiv preprint arXiv:180301271

\bibitem[{Barlin et~al(2013)Barlin, Zhou, Clair, Iasonos, Soslow, Alektiar, Hensley, Leitao~Jr, Barakat, and Abu-Rustum}]{barlin2013classification}
Barlin JN, Zhou Q, Clair CMS, et~al (2013) Classification and regression tree (cart) analysis of endometrial carcinoma: seeing the forest for the trees. Gynecologic oncology 130(3):452--456

\bibitem[{Behrouz et~al(2024)Behrouz, Santacatterina, and Zabih}]{Chimera}
Behrouz A, Santacatterina M, Zabih R (2024) Chimera: Effectively modeling multivariate time series with 2-dimensional state space models. In: The Thirty-eighth Annual Conference on Neural Information Processing Systems, \urlprefix\url{https://openreview.net/forum?id=ncYGjx2vnE}

\bibitem[{Beltagy et~al(2020)Beltagy, Peters, and Cohan}]{Beltagy2020Longformer}
Beltagy I, Peters ME, Cohan A (2020) Longformer: The long-document transformer. arXiv:200405150

\bibitem[{Benidis et~al(2022)Benidis, Rangapuram, Flunkert, Wang, Maddix, Turkmen, Gasthaus, Bohlke-Schneider, Salinas, Stella et~al}]{DLforTSF}
Benidis K, Rangapuram SS, Flunkert V, et~al (2022) Deep learning for time series forecasting: Tutorial and literature survey. ACM Computing Surveys 55(6):1--36

\bibitem[{Bergsma et~al(2023)Bergsma, Zeyl, and Guo}]{SutraNets}
Bergsma S, Zeyl T, Guo L (2023) Sutranets: sub-series autoregressive networks for long-sequence, probabilistic forecasting. Advances in Neural Information Processing Systems 36:30518--30533

\bibitem[{Box et~al(1970)Box, Jenkins, Reinsel et~al}]{box1970forecasting}
Box GE, Jenkins GM, Reinsel G, et~al (1970) Forecasting and control. Time Series Analysis 3(75):1970

\bibitem[{Breiman(1996)}]{BaggingPredictors}
Breiman L (1996) Bagging predictors. Machine learning 24:123--140

\bibitem[{Brown(1959)}]{brown1959statistical}
Brown RG (1959) Statistical forecasting for inventory control. (No Title)

\bibitem[{Brown et~al(2020)Brown, Mann, Ryder, Subbiah, Kaplan, Dhariwal, Neelakantan, Shyam, Sastry, Askell et~al}]{GPT3}
Brown T, Mann B, Ryder N, et~al (2020) Language models are few-shot learners. Advances in neural information processing systems 33:1877--1901

\bibitem[{Cai et~al(2024{\natexlab{a}})Cai, Liang, Liu, Feng, and Wu}]{MSGNet}
Cai W, Liang Y, Liu X, et~al (2024{\natexlab{a}}) Msgnet: Learning multi-scale inter-series correlations for multivariate time series forecasting. In: Proceedings of the AAAI Conference on Artificial Intelligence, pp 11141--11149

\bibitem[{Cai et~al(2024{\natexlab{b}})Cai, Wang, Wu, Chen, and Wu}]{ForecastGrapher}
Cai W, Wang K, Wu H, et~al (2024{\natexlab{b}}) Forecastgrapher: Redefining multivariate time series forecasting with graph neural networks. arXiv preprint arXiv:240518036

\bibitem[{Cai et~al(2024{\natexlab{c}})Cai, Zhu, Wang, and Yao}]{MambaTS}
Cai X, Zhu Y, Wang X, et~al (2024{\natexlab{c}}) Mambats: Improved selective state space models for long-term time series forecasting. arXiv preprint arXiv:240516440

\bibitem[{Cao et~al(2024)Cao, Tan, Gao, Xu, Chen, Heng, and Li}]{cao2024survey}
Cao H, Tan C, Gao Z, et~al (2024) A survey on generative diffusion models. IEEE Transactions on Knowledge and Data Engineering

\bibitem[{Center(2020)}]{center2020dominick}
Center JMK (2020) Dominick’s dataset

\bibitem[{Challu et~al(2023)Challu, Olivares, Oreshkin, Ramirez, Canseco, and Dubrawski}]{N-HITS}
Challu C, Olivares KG, Oreshkin BN, et~al (2023) Nhits: Neural hierarchical interpolation for time series forecasting. In: Proceedings of the AAAI conference on artificial intelligence, pp 6989--6997

\bibitem[{Chang et~al(2017)Chang, Zhang, Han, Yu, Guo, Tan, Cui, Witbrock, Hasegawa-Johnson, and Huang}]{DilatedRNN}
Chang S, Zhang Y, Han W, et~al (2017) Dilated recurrent neural networks. Advances in neural information processing systems 30

\bibitem[{Chen et~al(2022)Chen, Lundberg, and Lee}]{chen2022explaining}
Chen H, Lundberg SM, Lee SI (2022) Explaining a series of models by propagating shapley values. Nature communications 13(1):4512

\bibitem[{Chen et~al(2024{\natexlab{a}})Chen, Lenssen, Feng, Hu, Fey, Tassiulas, Leskovec, and Ying}]{CCM}
Chen J, Lenssen JE, Feng A, et~al (2024{\natexlab{a}}) From similarity to superiority: Channel clustering for time series forecasting. In: The Thirty-eighth Annual Conference on Neural Information Processing Systems, \urlprefix\url{https://openreview.net/forum?id=MDgn9aazo0}

\bibitem[{Chen et~al(2021{\natexlab{a}})Chen, Chen, Xu, Zhang, Huang, and Knoll}]{chen2021nast}
Chen K, Chen G, Xu D, et~al (2021{\natexlab{a}}) Nast: Non-autoregressive spatial-temporal transformer for time series forecasting. arXiv preprint arXiv:210205624

\bibitem[{Chen et~al(2021{\natexlab{b}})Chen, Fu, and Wang}]{GCN-SBULSTM}
Chen P, Fu X, Wang X (2021{\natexlab{b}}) A graph convolutional stacked bidirectional unidirectional-lstm neural network for metro ridership prediction. IEEE Transactions on Intelligent Transportation Systems 23(7):6950--6962

\bibitem[{Chen et~al(2024{\natexlab{b}})Chen, ZHANG, Cheng, Shu, Wang, Wen, Yang, and Guo}]{pathformer}
Chen P, ZHANG Y, Cheng Y, et~al (2024{\natexlab{b}}) Pathformer: Multi-scale transformers with adaptive pathways for time series forecasting. In: The Twelfth International Conference on Learning Representations, \urlprefix\url{https://openreview.net/forum?id=lJkOCMP2aW}

\bibitem[{Chen and Guestrin(2016)}]{xgboost}
Chen T, Guestrin C (2016) Xgboost: A scalable tree boosting system. In: Proceedings of the 22nd acm sigkdd international conference on knowledge discovery and data mining, pp 785--794

\bibitem[{Chen et~al(2024{\natexlab{c}})Chen, Li, Liu, and Li}]{BiTGraph}
Chen X, Li X, Liu B, et~al (2024{\natexlab{c}}) Biased temporal convolution graph network for time series forecasting with missing values. In: The Twelft International Conference on Learning Representations

\bibitem[{Chen et~al(2021{\natexlab{c}})Chen, Zou, Li, Li, Yang, and Chen}]{chen2021multiple}
Chen Y, Zou X, Li K, et~al (2021{\natexlab{c}}) Multiple local 3d cnns for region-based prediction in smart cities. Information Sciences 542:476--491

\bibitem[{Chen et~al(2024{\natexlab{d}})Chen, Ren, Wang, Fang, Sun, and Li}]{contiformer}
Chen Y, Ren K, Wang Y, et~al (2024{\natexlab{d}}) Contiformer: Continuous-time transformer for irregular time series modeling. Advances in Neural Information Processing Systems 36

\bibitem[{Chen et~al(2023)Chen, Ma, Li, Wang, and Li}]{LSTFwithDeep}
Chen Z, Ma M, Li T, et~al (2023) Long sequence time-series forecasting with deep learning: A survey. Information Fusion 97:101819

\bibitem[{Cheng et~al(2024{\natexlab{a}})Cheng, Wen, Liu, and Sun}]{RobustTSF}
Cheng H, Wen Q, Liu Y, et~al (2024{\natexlab{a}}) Robusttsf: Towards theory and design of robust time series forecasting with anomalies. ICRL

\bibitem[{Cheng et~al(2024{\natexlab{b}})Cheng, Yang, Pan, Liu, and Li}]{ConvTimeNet}
Cheng M, Yang J, Pan T, et~al (2024{\natexlab{b}}) Convtimenet: A deep hierarchical fully convolutional model for multivariate time series analysis. arXiv preprint arXiv:240301493

\bibitem[{Cheng et~al(2024{\natexlab{c}})Cheng, Chen, Li, Luo, Wang, Zhao, and Yan}]{GridTST}
Cheng X, Chen X, Li S, et~al (2024{\natexlab{c}}) Leveraging 2d information for long-term time series forecasting with vanilla transformers. arXiv preprint arXiv:240513810

\bibitem[{Chicco et~al(2021)Chicco, Warrens, and Jurman}]{chicco2021coefficient}
Chicco D, Warrens MJ, Jurman G (2021) The coefficient of determination r-squared is more informative than smape, mae, mape, mse and rmse in regression analysis evaluation. Peerj computer science 7:e623

\bibitem[{Cho et~al(2014)Cho, van Merri{\"e}nboer, Gulcehre, Bahdanau, Bougares, Schwenk, and Bengio}]{GRU}
Cho K, van Merri{\"e}nboer B, Gulcehre C, et~al (2014) Learning phrase representations using {RNN} encoder{--}decoder for statistical machine translation. In: Moschitti A, Pang B, Daelemans W (eds) Proceedings of the 2014 Conference on Empirical Methods in Natural Language Processing ({EMNLP}). Association for Computational Linguistics, Doha, Qatar, pp 1724--1734, \doi{10.3115/v1/D14-1179}, \urlprefix\url{https://aclanthology.org/D14-1179/}

\bibitem[{Clark et~al(2020)Clark, Luong, Le, and Manning}]{ELECTRA}
Clark K, Luong MT, Le QV, et~al (2020) Electra: Pre-training text encoders as discriminators rather than generators. In: International Conference on Learning Representations, \urlprefix\url{https://openreview.net/forum?id=r1xMH1BtvB}

\bibitem[{Cortes(1995)}]{cortes1995support}
Cortes C (1995) Support-vector networks. Machine Learning

\bibitem[{Coskunuzer et~al(2024)Coskunuzer, Segovia-Dominguez, Chen, and Gel}]{TMP-Nets}
Coskunuzer B, Segovia-Dominguez I, Chen Y, et~al (2024) Time-aware knowledge representations of dynamic objects with multidimensional persistence. In: Proceedings of the AAAI Conference on Artificial Intelligence, pp 11678--11686

\bibitem[{Crabb{\'e} et~al(2024)Crabb{\'e}, Huynh, Stanczuk, and van~der Schaar}]{DiffusionInfrequencyDomain}
Crabb{\'e} J, Huynh N, Stanczuk JP, et~al (2024) Time series diffusion in the frequency domain. In: Forty-first International Conference on Machine Learning, \urlprefix\url{https://openreview.net/forum?id=W9GaJUVLCT}

\bibitem[{Cryer(1986)}]{cryer1986time}
Cryer JD (1986) Time series analysis, vol 286. Duxbury Press Boston

\bibitem[{Dai et~al(2019)Dai, Yang, Yang, Carbonell, Le, and Salakhutdinov}]{transformer-xl}
Dai Z, Yang Z, Yang Y, et~al (2019) Transformer-{XL}: Attentive language models beyond a fixed-length context. In: Korhonen A, Traum D, M{\`a}rquez L (eds) Proceedings of the 57th Annual Meeting of the Association for Computational Linguistics. Association for Computational Linguistics, Florence, Italy, pp 2978--2988, \doi{10.18653/v1/P19-1285}, \urlprefix\url{https://aclanthology.org/P19-1285}

\bibitem[{Danese and Kalchschmidt(2011)}]{DANESE2011204}
Danese P, Kalchschmidt M (2011) The role of the forecasting process in improving forecast accuracy and operational performance. International Journal of Production Economics 131(1):204--214. \doi{https://doi.org/10.1016/j.ijpe.2010.09.006}, \urlprefix\url{https://www.sciencedirect.com/science/article/pii/S0925527310003282}, innsbruck 2008

\bibitem[{Das et~al(2024)Das, Kong, Sen, and Zhou}]{TimesFM}
Das A, Kong W, Sen R, et~al (2024) A decoder-only foundation model for time-series forecasting. In: Forty-first International Conference on Machine Learning, \urlprefix\url{https://openreview.net/forum?id=jn2iTJas6h}

\bibitem[{Devlin et~al(2019)Devlin, Chang, Lee, and Toutanova}]{BERT}
Devlin J, Chang MW, Lee K, et~al (2019) {BERT}: Pre-training of deep bidirectional transformers for language understanding. In: Burstein J, Doran C, Solorio T (eds) Proceedings of the 2019 Conference of the North {A}merican Chapter of the Association for Computational Linguistics: Human Language Technologies, Volume 1 (Long and Short Papers). Association for Computational Linguistics, Minneapolis, Minnesota, pp 4171--4186, \doi{10.18653/v1/N19-1423}, \urlprefix\url{https://aclanthology.org/N19-1423}

\bibitem[{Dimri et~al(2020)Dimri, Ahmad, and Sharif}]{climateForecasting}
Dimri T, Ahmad S, Sharif M (2020) Time series analysis of climate variables using seasonal arima approach. Journal of Earth System Science 129:1--16

\bibitem[{Dosovitskiy et~al(2021)Dosovitskiy, Beyer, Kolesnikov, Weissenborn, Zhai, Unterthiner, Dehghani, Minderer, Heigold, Gelly, Uszkoreit, and Houlsby}]{ViT}
Dosovitskiy A, Beyer L, Kolesnikov A, et~al (2021) An image is worth 16x16 words: Transformers for image recognition at scale. In: International Conference on Learning Representations, \urlprefix\url{https://openreview.net/forum?id=YicbFdNTTy}

\bibitem[{Dragomiretskiy and Zosso(2013)}]{VMD}
Dragomiretskiy K, Zosso D (2013) Variational mode decomposition. IEEE transactions on signal processing 62(3):531--544

\bibitem[{Dubey et~al(2024)Dubey, Jauhri, Pandey, Kadian, Al-Dahle, Letman, Mathur, Schelten, Yang, Fan et~al}]{dubey2024llama}
Dubey A, Jauhri A, Pandey A, et~al (2024) The llama 3 herd of models. arXiv preprint arXiv:240721783

\bibitem[{Dudek(2024)}]{probstacking2}
Dudek G (2024) Stacking for probabilistic short-term load forecasting. In: International Conference on Computational Science, Springer, pp 3--18

\bibitem[{Ekambaram et~al(2023)Ekambaram, Jati, Nguyen, Sinthong, and Kalagnanam}]{TSMixer}
Ekambaram V, Jati A, Nguyen N, et~al (2023) Tsmixer: Lightweight mlp-mixer model for multivariate time series forecasting. In: Proceedings of the 29th ACM SIGKDD Conference on Knowledge Discovery and Data Mining, pp 459--469

\bibitem[{Ekambaram et~al(2024)Ekambaram, Jati, Dayama, Mukherjee, Nguyen, Gifford, Reddy, and Kalagnanam}]{TTMs}
Ekambaram V, Jati A, Dayama P, et~al (2024) Tiny time mixers (ttms): Fast pre-trained models for enhanced zero/few-shot forecasting of multivariate time series. CoRR

\bibitem[{Eldele et~al(2021)Eldele, Ragab, Chen, Wu, Kwoh, Li, and Guan}]{TS-TCC}
Eldele E, Ragab M, Chen Z, et~al (2021) Time-series representation learning via temporal and contextual contrasting. Proceedings of the Thirtieth International Joint Conference on Artificial Intelligence (IJCAI-21)

\bibitem[{Eldele et~al(2024)Eldele, Ragab, Chen, Wu, and Li}]{TSLANet}
Eldele E, Ragab M, Chen Z, et~al (2024) Tslanet: Rethinking transformers for time series representation learning. International Conference on Machine Learning

\bibitem[{Fan et~al(2021)Fan, Xiong, Mangalam, Li, Yan, Malik, and Feichtenhofer}]{fan2021multiscale}
Fan H, Xiong B, Mangalam K, et~al (2021) Multiscale vision transformers. In: Proceedings of the IEEE/CVF international conference on computer vision, pp 6824--6835

\bibitem[{Fan et~al(2023)Fan, Wang, Wang, Wang, Zhou, and Fu}]{Dish-TS}
Fan W, Wang P, Wang D, et~al (2023) Dish-ts: a general paradigm for alleviating distribution shift in time series forecasting. In: Proceedings of the AAAI Conference on Artificial Intelligence, pp 7522--7529

\bibitem[{Fan et~al(2024{\natexlab{a}})Fan, Yi, Ye, Ning, Zhang, and An}]{DERITS}
Fan W, Yi K, Ye H, et~al (2024{\natexlab{a}}) Deep frequency derivative learning for non-stationary time series forecasting. IJCAI

\bibitem[{Fan et~al(2024{\natexlab{b}})Fan, Wu, Xu, Huang, Liu, and Bian}]{MG-TSD}
Fan X, Wu Y, Xu C, et~al (2024{\natexlab{b}}) {MG}-{TSD}: Multi-granularity time series diffusion models with guided learning process. In: The Twelfth International Conference on Learning Representations, \urlprefix\url{https://openreview.net/forum?id=CZiY6OLktd}

\bibitem[{Feng et~al(2024{\natexlab{a}})Feng, Chen, and Song}]{feng2024learning}
Feng R, Chen M, Song Y (2024{\natexlab{a}}) Learning traffic as videos: Short-term traffic flow prediction using mixed-pointwise convolution and channel attention mechanism. Expert Systems with Applications 240:122468

\bibitem[{Feng et~al(2024{\natexlab{b}})Feng, Miao, Zhang, and Zhao}]{LDT}
Feng S, Miao C, Zhang Z, et~al (2024{\natexlab{b}}) Latent diffusion transformer for probabilistic time series forecasting. In: Proceedings of the AAAI Conference on Artificial Intelligence, pp 11979--11987

\bibitem[{Freund and Schapire(1997)}]{freund1997decision}
Freund Y, Schapire RE (1997) A decision-theoretic generalization of on-line learning and an application to boosting. Journal of computer and system sciences 55(1):119--139

\bibitem[{Friedman(2001)}]{friedman2001greedy}
Friedman JH (2001) Greedy function approximation: a gradient boosting machine. Annals of statistics pp 1189--1232

\bibitem[{Fu et~al(2023)Fu, Dao, Saab, Thomas, Rudra, and R{\'e}}]{H3}
Fu DY, Dao T, Saab KK, et~al (2023) Hungry hungry hippos: Towards language modeling with state space models. In The International Conference on Learning Representations (ICLR)

\bibitem[{Fukushima(1980)}]{Neocognitron}
Fukushima K (1980) Neocognitron: A self-organizing neural network model for a mechanism of pattern recognition unaffected by shift in position. Biological cybernetics 36(4):193--202

\bibitem[{Gallifant et~al(2024)Gallifant, Fiske, Levites~Strekalova, Osorio-Valencia, Parke, Mwavu, Martinez, Gichoya, Ghassemi, Demner-Fushman et~al}]{gallifant2024peer}
Gallifant J, Fiske A, Levites~Strekalova YA, et~al (2024) Peer review of gpt-4 technical report and systems card. PLOS Digital Health 3(1):e0000417

\bibitem[{Ganaie et~al(2022)Ganaie, Hu, Malik, Tanveer, and Suganthan}]{EnsembleDL}
Ganaie MA, Hu M, Malik AK, et~al (2022) Ensemble deep learning: A review. Engineering Applications of Artificial Intelligence 115:105151

\bibitem[{Garc{\'\i}a and Aznarte(2020)}]{garcia2020shapley}
Garc{\'\i}a MV, Aznarte JL (2020) Shapley additive explanations for no2 forecasting. Ecological Informatics 56:101039

\bibitem[{Godahewa et~al(2021{\natexlab{a}})Godahewa, Bergmeir, Webb, Abolghasemi, Hyndman, and Montero-Manso}]{godahewa_2021_4656027}
Godahewa R, Bergmeir C, Webb G, et~al (2021{\natexlab{a}}) Solar power dataset (4 seconds observations). \doi{10.5281/zenodo.4656027}, \urlprefix\url{https://doi.org/10.5281/zenodo.4656027}

\bibitem[{Godahewa et~al(2021{\natexlab{b}})Godahewa, Bergmeir, Webb, Abolghasemi, Hyndman, and Montero-Manso}]{godahewa_2021_4654909}
Godahewa R, Bergmeir C, Webb G, et~al (2021{\natexlab{b}}) Wind farms dataset (with missing values). \doi{10.5281/zenodo.4654909}, \urlprefix\url{https://doi.org/10.5281/zenodo.4654909}

\bibitem[{Godahewa et~al(2021{\natexlab{c}})Godahewa, Bergmeir, Webb, Abolghasemi, Hyndman, and Montero-Manso}]{godahewa_2021_4654858}
Godahewa R, Bergmeir C, Webb G, et~al (2021{\natexlab{c}}) Wind farms dataset (without missing values). \doi{10.5281/zenodo.4654858}, \urlprefix\url{https://doi.org/10.5281/zenodo.4654858}

\bibitem[{Godahewa et~al(2021{\natexlab{d}})Godahewa, Bergmeir, Webb, Abolghasemi, Hyndman, and Montero-Manso}]{godahewa_2021_4656032}
Godahewa R, Bergmeir C, Webb G, et~al (2021{\natexlab{d}}) Wind power dataset (4 seconds observations). \doi{10.5281/zenodo.4656032}, \urlprefix\url{https://doi.org/10.5281/zenodo.4656032}

\bibitem[{Godahewa et~al(2021{\natexlab{e}})Godahewa, Bergmeir, Webb, Hyndman, and Montero-Manso}]{godahewa_2021_4659727}
Godahewa R, Bergmeir C, Webb G, et~al (2021{\natexlab{e}}) Australian electricity demand dataset. \doi{10.5281/zenodo.4659727}, \urlprefix\url{https://doi.org/10.5281/zenodo.4659727}

\bibitem[{Godahewa et~al(2021{\natexlab{f}})Godahewa, Bergmeir, Webb, Hyndman, and Montero-Manso}]{godahewa_2021_5121965}
Godahewa R, Bergmeir C, Webb G, et~al (2021{\natexlab{f}}) Bitcoin dataset with missing values. \doi{10.5281/zenodo.5121965}, \urlprefix\url{https://doi.org/10.5281/zenodo.5121965}

\bibitem[{Godahewa et~al(2021{\natexlab{g}})Godahewa, Bergmeir, Webb, Hyndman, and Montero-Manso}]{godahewa_2021_5122101}
Godahewa R, Bergmeir C, Webb G, et~al (2021{\natexlab{g}}) Bitcoin dataset without missing values. \doi{10.5281/zenodo.5122101}, \urlprefix\url{https://doi.org/10.5281/zenodo.5122101}

\bibitem[{Godahewa et~al(2021{\natexlab{h}})Godahewa, Bergmeir, Webb, Hyndman, and Montero-Manso}]{godahewa_2021_5122114}
Godahewa R, Bergmeir C, Webb G, et~al (2021{\natexlab{h}}) Rideshare dataset with missing values. \doi{10.5281/zenodo.5122114}, \urlprefix\url{https://doi.org/10.5281/zenodo.5122114}

\bibitem[{Godahewa et~al(2021{\natexlab{i}})Godahewa, Bergmeir, Webb, Hyndman, and Montero-Manso}]{godahewa_2021_5122232}
Godahewa R, Bergmeir C, Webb G, et~al (2021{\natexlab{i}}) Rideshare dataset without missing values. \doi{10.5281/zenodo.5122232}, \urlprefix\url{https://doi.org/10.5281/zenodo.5122232}

\bibitem[{Godahewa et~al(2021{\natexlab{j}})Godahewa, Bergmeir, Webb, Hyndman, and Montero-Manso}]{godahewa_2021_5129073}
Godahewa R, Bergmeir C, Webb G, et~al (2021{\natexlab{j}}) Temperature rain dataset with missing values. \doi{10.5281/zenodo.5129073}, \urlprefix\url{https://doi.org/10.5281/zenodo.5129073}

\bibitem[{Godahewa et~al(2021{\natexlab{k}})Godahewa, Bergmeir, Webb, Hyndman, and Montero-Manso}]{godahewa_2021_5129091}
Godahewa R, Bergmeir C, Webb G, et~al (2021{\natexlab{k}}) Temperature rain dataset without missing values. \doi{10.5281/zenodo.5129091}, \urlprefix\url{https://doi.org/10.5281/zenodo.5129091}

\bibitem[{Godahewa et~al(2021{\natexlab{l}})Godahewa, Bergmeir, Webb, Hyndman, and Montero-Manso}]{godahewa2021monash}
Godahewa RW, Bergmeir C, Webb GI, et~al (2021{\natexlab{l}}) Monash time series forecasting archive. In: Thirty-fifth Conference on Neural Information Processing Systems Datasets and Benchmarks Track (Round 2), \urlprefix\url{https://openreview.net/forum?id=wEc1mgAjU-}

\bibitem[{Gong et~al(2023)Gong, Tang, and Liang}]{PatchMixer}
Gong Z, Tang Y, Liang J (2023) Patchmixer: A patch-mixing architecture for long-term time series forecasting. arXiv preprint arXiv:231000655

\bibitem[{Granger(1969)}]{Granger}
Granger CW (1969) Investigating causal relations by econometric models and cross-spectral methods. Econometrica: journal of the Econometric Society pp 424--438

\bibitem[{Graziani et~al(2023)Graziani, Dutkiewicz, Calvaresi, Amorim, Yordanova, Vered, Nair, Abreu, Blanke, Pulignano et~al}]{graziani2023global}
Graziani M, Dutkiewicz L, Calvaresi D, et~al (2023) A global taxonomy of interpretable ai: unifying the terminology for the technical and social sciences. Artificial intelligence review 56(4):3473--3504

\bibitem[{Gruver et~al(2023)Gruver, Finzi, Qiu, and Wilson}]{LLMTime}
Gruver N, Finzi MA, Qiu S, et~al (2023) Large language models are zero-shot time series forecasters. In: Thirty-seventh Conference on Neural Information Processing Systems, \urlprefix\url{https://openreview.net/forum?id=md68e8iZK1}

\bibitem[{Gu and Dao(2024)}]{Mamba}
Gu A, Dao T (2024) Mamba: Linear-time sequence modeling with selective state spaces. In: First Conference on Language Modeling, \urlprefix\url{https://openreview.net/forum?id=tEYskw1VY2}

\bibitem[{Gu et~al(2021)Gu, Johnson, Goel, Saab, Dao, Rudra, and R{\'e}}]{S4}
Gu A, Johnson I, Goel K, et~al (2021) Combining recurrent, convolutional, and continuous-time models with linear state space layers. Advances in neural information processing systems 34:572--585

\bibitem[{Gu et~al(2022)Gu, Goel, and Re}]{gu2021efficiently}
Gu A, Goel K, Re C (2022) Efficiently modeling long sequences with structured state spaces. In: International Conference on Learning Representations, \urlprefix\url{https://openreview.net/forum?id=uYLFoz1vlAC}

\bibitem[{Hahn et~al(2023)Hahn, Langer, Meyes, and Meisen}]{datasetsurvey}
Hahn Y, Langer T, Meyes R, et~al (2023) Time series dataset survey for forecasting with deep learning. Forecasting 5(1):315--335

\bibitem[{Hamon et~al(2022)Hamon, Junklewitz, Sanchez, Malgieri, and De~Hert}]{hamon2022bridging}
Hamon R, Junklewitz H, Sanchez I, et~al (2022) Bridging the gap between ai and explainability in the gdpr: towards trustworthiness-by-design in automated decision-making. IEEE Computational Intelligence Magazine 17(1):72--85

\bibitem[{Han et~al(2023{\natexlab{a}})Han, Kim, Choi, and Yoon}]{han2023impact}
Han H, Kim S, Choi HS, et~al (2023{\natexlab{a}}) On the impact of knowledge distillation for model interpretability. In: Proceedings of the 40th International Conference on Machine Learning. JMLR.org, ICML'23

\bibitem[{Han et~al(2023{\natexlab{b}})Han, Park, Min, Kim, Kim, Park, Kim, Park, An, Lee et~al}]{han2023improving}
Han H, Park S, Min S, et~al (2023{\natexlab{b}}) Improving generalization performance of electrocardiogram classification models. Physiological Measurement 44(5):054003

\bibitem[{Han et~al(2024)Han, Chen, Ye, and Zhan}]{SOFTS}
Han L, Chen XY, Ye HJ, et~al (2024) {SOFTS}: Efficient multivariate time series forecasting with series-core fusion. In: The Thirty-eighth Annual Conference on Neural Information Processing Systems, \urlprefix\url{https://openreview.net/forum?id=89AUi5L1uA}

\bibitem[{Hasson et~al(2023)Hasson, Maddix, Wang, Gupta, and Park}]{theoreticalEnsemble}
Hasson H, Maddix DC, Wang B, et~al (2023) Theoretical guarantees of learning ensembling strategies with applications to time series forecasting. In: International Conference on Machine Learning, PMLR, pp 12616--12632

\bibitem[{Ho et~al(2020)Ho, Jain, and Abbeel}]{DDPM}
Ho J, Jain A, Abbeel P (2020) Denoising diffusion probabilistic models. Advances in neural information processing systems 33:6840--6851

\bibitem[{Hochreiter and Schmidhuber(1997)}]{LSTM}
Hochreiter S, Schmidhuber J (1997) Long short-term memory. Neural computation 9(8):1735--1780

\bibitem[{Holt(1957)}]{holt1957forecasting}
Holt CC (1957) Forecasting trends and seasonals by exponentially weighted averages. carnegie institute of technology. Pittsburgh ONR memorandum

\bibitem[{Hopfield(1982)}]{RNN}
Hopfield JJ (1982) Neural networks and physical systems with emergent collective computational abilities. Proceedings of the national academy of sciences 79(8):2554--2558

\bibitem[{Hou and Yu(2024)}]{RWKV-TS}
Hou H, Yu FR (2024) Rwkv-ts: Beyond traditional recurrent neural network for time series tasks. arXiv preprint arXiv:240109093

\bibitem[{Hounie et~al(2024)Hounie, Porras-Valenzuela, and Ribeiro}]{PDLS}
Hounie I, Porras-Valenzuela J, Ribeiro A (2024) Transformers with loss shaping constraints for long-term time series forecasting. In: Forty-first International Conference on Machine Learning

\bibitem[{Hu et~al(2024{\natexlab{a}})Hu, Lan, Zhou, Wen, and Liang}]{Time-SSM}
Hu J, Lan D, Zhou Z, et~al (2024{\natexlab{a}}) Time-ssm: Simplifying and unifying state space models for time series forecasting. CoRR abs/2405.16312. \urlprefix\url{https://doi.org/10.48550/arXiv.2405.16312}

\bibitem[{Hu et~al(2024{\natexlab{b}})Hu, Liu, Zhu, Cheng, and Dai}]{AMD}
Hu Y, Liu P, Zhu P, et~al (2024{\natexlab{b}}) Adaptive multi-scale decomposition framework for time series forecasting. CoRR abs/2406.03751. \urlprefix\url{https://doi.org/10.48550/arXiv.2406.03751}

\bibitem[{Huang et~al(1998)Huang, Shen, Long, Wu, Shih, Zheng, Yen, Tung, and Liu}]{EMD}
Huang NE, Shen Z, Long SR, et~al (1998) The empirical mode decomposition and the hilbert spectrum for nonlinear and non-stationary time series analysis. Proceedings of the Royal Society of London Series A: mathematical, physical and engineering sciences 454(1971):903--995

\bibitem[{Huang et~al(2024{\natexlab{a}})Huang, Shen, Zhang, Cheng, Ding, Zhou, and Wang}]{HDMixer}
Huang Q, Shen L, Zhang R, et~al (2024{\natexlab{a}}) Hdmixer: Hierarchical dependency with extendable patch for multivariate time series forecasting. In: Proceedings of the AAAI Conference on Artificial Intelligence, pp 12608--12616

\bibitem[{Huang et~al(2024{\natexlab{b}})Huang, Liu, Zhang, Li, Li, and Zhang}]{CrossWaveNet}
Huang S, Liu Y, Zhang F, et~al (2024{\natexlab{b}}) Crosswavenet: A dual-channel network with deep cross-decomposition for long-term time series forecasting. Expert Systems with Applications 238:121642

\bibitem[{Huber(1992)}]{HuberLoss}
Huber PJ (1992) Robust estimation of a location parameter. In: Breakthroughs in statistics: Methodology and distribution. Springer, p 492--518

\bibitem[{Ilbert et~al(2024)Ilbert, Odonnat, Feofanov, Virmaux, Paolo, Palpanas, and Redko}]{SAMformer}
Ilbert R, Odonnat A, Feofanov V, et~al (2024) {SAM}former: Unlocking the potential of transformers in time series forecasting with sharpness-aware minimization and channel-wise attention. In: Salakhutdinov R, Kolter Z, Heller K, et~al (eds) Proceedings of the 41st International Conference on Machine Learning, Proceedings of Machine Learning Research, vol 235. PMLR, pp 20924--20954, \urlprefix\url{https://proceedings.mlr.press/v235/ilbert24a.html}

\bibitem[{Ioffe and Szegedy(2015)}]{BatchNorm}
Ioffe S, Szegedy C (2015) Batch normalization: Accelerating deep network training by reducing internal covariate shift. In: International conference on machine learning, pmlr, pp 448--456

\bibitem[{Jhin et~al(2024)Jhin, Kim, and Park}]{CONTIME}
Jhin SY, Kim S, Park N (2024) Addressing prediction delays in time series forecasting: A continuous gru approach with derivative regularization. In: Proceedings of the 30th ACM SIGKDD Conference on Knowledge Discovery and Data Mining, pp 1234--1245

\bibitem[{Jia et~al(2024)Jia, Lin, Hao, Lin, Guo, and Wan}]{WITRAN}
Jia Y, Lin Y, Hao X, et~al (2024) Witran: Water-wave information transmission and recurrent acceleration network for long-range time series forecasting. Advances in Neural Information Processing Systems 36

\bibitem[{Jin et~al(2023)Jin, Wen, Liang, Zhang, Xue, Wang, Zhang, Wang, Chen, Li et~al}]{jin2023large}
Jin M, Wen Q, Liang Y, et~al (2023) Large models for time series and spatio-temporal data: A survey and outlook. arXiv preprint arXiv:231010196

\bibitem[{Jin et~al(2024)Jin, Wang, Ma, Chu, Zhang, Shi, Chen, Liang, Li, Pan, and Wen}]{TimeLLM}
Jin M, Wang S, Ma L, et~al (2024) {Time-LLM}: Time series forecasting by reprogramming large language models. In: International Conference on Learning Representations (ICLR)

\bibitem[{Jung et~al(2020)Jung, Lee, Yi, and Yoon}]{jung2020icaps}
Jung D, Lee J, Yi J, et~al (2020) icaps: An interpretable classifier via disentangled capsule networks. In: European Conference on Computer Vision, Springer, pp 314--330

\bibitem[{Kalman(1960)}]{statespacemodel}
Kalman RE (1960) A new approach to linear filtering and prediction problems. Transactions of the ASME–Journal of Basic Engineering 82(1):35--45

\bibitem[{Kang et~al(2024)Kang, Lee, Kim, Chung, and Yoon}]{2024BSA}
Kang BG, Lee D, Kim H, et~al (2024) Introducing spectral attention for long-range dependency in time series forecasting. In: Advances in Neural Information Processing Systems

\bibitem[{Ke et~al(2017)Ke, Meng, Finley, Wang, Chen, Ma, Ye, and Liu}]{ke2017lightgbm}
Ke G, Meng Q, Finley T, et~al (2017) Lightgbm: A highly efficient gradient boosting decision tree. Advances in neural information processing systems 30

\bibitem[{Khosravi et~al(2011)Khosravi, Mazloumi, Nahavandi, Creighton, and Van~Lint}]{khosravi2011prediction}
Khosravi A, Mazloumi E, Nahavandi S, et~al (2011) Prediction intervals to account for uncertainties in travel time prediction. IEEE Transactions on Intelligent Transportation Systems 12(2):537--547

\bibitem[{Kim and Baek(2022)}]{kim2022bagging}
Kim D, Baek JG (2022) Bagging ensemble-based novel data generation method for univariate time series forecasting. Expert Systems with Applications 203:117366

\bibitem[{Kim et~al(2024)Kim, Park, Lee, and Kim}]{kim2024self}
Kim D, Park J, Lee J, et~al (2024) Are self-attentions effective for time series forecasting? In: The Thirty-eighth Annual Conference on Neural Information Processing Systems, \urlprefix\url{https://openreview.net/forum?id=iN43sJoib7}

\bibitem[{Kim et~al(2020)Kim, Yi, Kim, and Yoon}]{kim2020interpretation}
Kim S, Yi J, Kim E, et~al (2020) Interpretation of {NLP} models through input marginalization. In: Webber B, Cohn T, He Y, et~al (eds) Proceedings of the 2020 Conference on Empirical Methods in Natural Language Processing (EMNLP). Association for Computational Linguistics, Online, pp 3154--3167, \doi{10.18653/v1/2020.emnlp-main.255}, \urlprefix\url{https://aclanthology.org/2020.emnlp-main.255/}

\bibitem[{Kim et~al(2021)Kim, Kim, Tae, Park, Choi, and Choo}]{RevIN}
Kim T, Kim J, Tae Y, et~al (2021) Reversible instance normalization for accurate time-series forecasting against distribution shift. In: International Conference on Learning Representations

\bibitem[{Kipf and Welling(2017)}]{GCN}
Kipf TN, Welling M (2017) Semi-supervised classification with graph convolutional networks. In: International Conference on Learning Representations, \urlprefix\url{https://openreview.net/forum?id=SJU4ayYgl}

\bibitem[{Kitaev et~al(2020)Kitaev, Kaiser, and Levskaya}]{Reformer}
Kitaev N, Kaiser L, Levskaya A (2020) Reformer: The efficient transformer. In: International Conference on Learning Representations, \urlprefix\url{https://openreview.net/forum?id=rkgNKkHtvB}

\bibitem[{Koenker and Bassett~Jr(1978)}]{koenker1978regression}
Koenker R, Bassett~Jr G (1978) Regression quantiles. Econometrica: journal of the Econometric Society pp 33--50

\bibitem[{Kollovieh et~al(2024)Kollovieh, Ansari, Bohlke-Schneider, Zschiegner, Wang, and Wang}]{TSDiff}
Kollovieh M, Ansari AF, Bohlke-Schneider M, et~al (2024) Predict, refine, synthesize: Self-guiding diffusion models for probabilistic time series forecasting. Advances in Neural Information Processing Systems 36

\bibitem[{Kong et~al(2021)Kong, Ping, Huang, Zhao, and Catanzaro}]{kong2020diffwave}
Kong Z, Ping W, Huang J, et~al (2021) Diffwave: A versatile diffusion model for audio synthesis. In: International Conference on Learning Representations, \urlprefix\url{https://openreview.net/forum?id=a-xFK8Ymz5J}

\bibitem[{Kontopoulou et~al(2023)Kontopoulou, Panagopoulos, Kakkos, and Matsopoulos}]{kontopoulou2023review}
Kontopoulou VI, Panagopoulos AD, Kakkos I, et~al (2023) A review of arima vs. machine learning approaches for time series forecasting in data driven networks. Future Internet 15(8):255

\bibitem[{LeCun et~al(1998)LeCun, Bottou, Bengio, and Haffner}]{LeNet}
LeCun Y, Bottou L, Bengio Y, et~al (1998) Gradient-based learning applied to document recognition. Proceedings of the IEEE 86(11):2278--2324

\bibitem[{Li et~al(2023)Li, Li, Savarese, and Hoi}]{BLIP2}
Li J, Li D, Savarese S, et~al (2023) Blip-2: Bootstrapping language-image pre-training with frozen image encoders and large language models. In: International conference on machine learning, PMLR, pp 19730--19742

\bibitem[{Li et~al(2024{\natexlab{a}})Li, Xu, Ding, Liu, and Ran}]{DB-STGCN}
Li J, Xu X, Ding X, et~al (2024{\natexlab{a}}) Bayesian spatio-temporal graph convolutional network for railway train delay prediction. IEEE Transactions on Intelligent Transportation Systems

\bibitem[{Li et~al(2019)Li, Jin, Xuan, Zhou, Chen, Wang, and Yan}]{LogSparseTransformer}
Li S, Jin X, Xuan Y, et~al (2019) Enhancing the locality and breaking the memory bottleneck of transformer on time series forecasting. Advances in neural information processing systems 32

\bibitem[{Li et~al(2018)Li, Yu, Shahabi, and Liu}]{DCRNN}
Li Y, Yu R, Shahabi C, et~al (2018) Diffusion convolutional recurrent neural network: Data-driven traffic forecasting. In: International Conference on Learning Representations, \urlprefix\url{https://openreview.net/forum?id=SJiHXGWAZ}

\bibitem[{Li et~al(2024{\natexlab{b}})Li, Chen, Hu, Chen, baolin sun, and Zhou}]{TMDM}
Li Y, Chen W, Hu X, et~al (2024{\natexlab{b}}) Transformer-modulated diffusion models for probabilistic multivariate time series forecasting. In: The Twelfth International Conference on Learning Representations, \urlprefix\url{https://openreview.net/forum?id=qae04YACHs}

\bibitem[{Li et~al(2024{\natexlab{c}})Li, Xu, and Anastasiu}]{DAN}
Li Y, Xu J, Anastasiu D (2024{\natexlab{c}}) Learning from polar representation: An extreme-adaptive model for long-term time series forecasting. In: Proceedings of the AAAI Conference on Artificial Intelligence, pp 171--179

\bibitem[{Li et~al(2024{\natexlab{d}})Li, Qin, Cheng, and Tan}]{FTMixer}
Li Z, Qin Y, Cheng X, et~al (2024{\natexlab{d}}) Ftmixer: Frequency and time domain representations fusion for time series modeling. arXiv preprint arXiv:240515256

\bibitem[{Liang et~al(2024{\natexlab{a}})Liang, Jiang, Sun, and Lu}]{Bi-Mamba+}
Liang A, Jiang X, Sun Y, et~al (2024{\natexlab{a}}) Bi-mamba4ts: Bidirectional mamba for time series forecasting. arXiv preprint arXiv:240415772

\bibitem[{Liang et~al(2024{\natexlab{b}})Liang, Zhang, Yuan, Zhang, and Zhang}]{minusformer}
Liang D, Zhang H, Yuan D, et~al (2024{\natexlab{b}}) Minusformer: Improving time series forecasting by progressively learning residuals. arXiv preprint arXiv:240202332

\bibitem[{Liang et~al(2024{\natexlab{c}})Liang, Wen, Nie, Jiang, Jin, Song, Pan, and Wen}]{FMforTS}
Liang Y, Wen H, Nie Y, et~al (2024{\natexlab{c}}) Foundation models for time series analysis: A tutorial and survey. arXiv preprint arXiv:240314735

\bibitem[{Lim and Zohren(2021)}]{TSFwithDL}
Lim B, Zohren S (2021) Time-series forecasting with deep learning: a survey. Philosophical Transactions of the Royal Society A 379(2194):20200209

\bibitem[{Lim et~al(2021)Lim, Ar{\i}k, Loeff, and Pfister}]{lim2021temporal}
Lim B, Ar{\i}k S{\"O}, Loeff N, et~al (2021) Temporal fusion transformers for interpretable multi-horizon time series forecasting. International Journal of Forecasting 37(4):1748--1764

\bibitem[{Lin and Huang(2022)}]{ensemble_app}
Lin K, Huang C (2022) Ensemble learning applications in multiple industries: A review. Inf Dyn Appl 1(1):44--58

\bibitem[{Lin et~al(2024{\natexlab{a}})Lin, Li, Li, Li, and Gao}]{DiffForTS}
Lin L, Li Z, Li R, et~al (2024{\natexlab{a}}) Diffusion models for time-series applications: a survey. Frontiers of Information Technology \& Electronic Engineering 25(1):19--41

\bibitem[{Lin et~al(2024{\natexlab{b}})Lin, Lin, Wu, Chen, and Yang}]{SparseTSF}
Lin S, Lin W, Wu W, et~al (2024{\natexlab{b}}) Sparsetsf: Modeling long-term time series forecasting with 1k parameters. International Conference on Machine Learning

\bibitem[{Lin et~al(2024{\natexlab{c}})Lin, Lin, Wu, Wang, and Wang}]{Petformer}
Lin S, Lin W, Wu W, et~al (2024{\natexlab{c}}) {PET}former: Long-term time series forecasting via placeholder-enhanced transformer. \urlprefix\url{https://openreview.net/forum?id=u3RJbzzBZj}

\bibitem[{Lipton(2018)}]{lipton2018mythos}
Lipton ZC (2018) The mythos of model interpretability: In machine learning, the concept of interpretability is both important and slippery. Queue 16(3):31--57

\bibitem[{Liu et~al(2024{\natexlab{a}})Liu, Li, Wu, and Lee}]{Llava}
Liu H, Li C, Wu Q, et~al (2024{\natexlab{a}}) Visual instruction tuning. Advances in neural information processing systems 36

\bibitem[{Liu et~al(2024{\natexlab{b}})Liu, Liu, Woo, Wang, Hooi, Xiong, and Sahoo}]{uniTST}
Liu J, Liu C, Woo G, et~al (2024{\natexlab{b}}) Unitst: Effectively modeling inter-series and intra-series dependencies for multivariate time series forecasting. arXiv preprint arXiv:240604975

\bibitem[{Liu et~al(2021{\natexlab{a}})Liu, Yu, Liao, Li, Lin, Liu, and Dustdar}]{Pyraformer}
Liu S, Yu H, Liao C, et~al (2021{\natexlab{a}}) Pyraformer: Low-complexity pyramidal attention for long-range time series modeling and forecasting. In: International conference on learning representations

\bibitem[{Liu et~al(2022)Liu, Wu, Wang, and Long}]{nonstationaryTransformer}
Liu Y, Wu H, Wang J, et~al (2022) Non-stationary transformers: Exploring the stationarity in time series forecasting. Advances in Neural Information Processing Systems 35:9881--9893

\bibitem[{Liu et~al(2024{\natexlab{c}})Liu, Hu, Zhang, Wu, Wang, Ma, and Long}]{itransformer}
Liu Y, Hu T, Zhang H, et~al (2024{\natexlab{c}}) itransformer: Inverted transformers are effective for time series forecasting. In: The Twelfth International Conference on Learning Representations, \urlprefix\url{https://openreview.net/forum?id=JePfAI8fah}

\bibitem[{Liu et~al(2024{\natexlab{d}})Liu, Li, Wang, and Long}]{Koopa}
Liu Y, Li C, Wang J, et~al (2024{\natexlab{d}}) Koopa: Learning non-stationary time series dynamics with koopman predictors. Advances in Neural Information Processing Systems 36

\bibitem[{Liu et~al(2021{\natexlab{b}})Liu, Zhu, Gao, and Xu}]{ForecastMethodsforTS}
Liu Z, Zhu Z, Gao J, et~al (2021{\natexlab{b}}) Forecast methods for time series data: a survey. Ieee Access 9:91896--91912

\bibitem[{Liu et~al(2024{\natexlab{e}})Liu, Cheng, Li, Huang, Liu, Xie, and Chen}]{SAN}
Liu Z, Cheng M, Li Z, et~al (2024{\natexlab{e}}) Adaptive normalization for non-stationary time series forecasting: A temporal slice perspective. Advances in Neural Information Processing Systems 36

\bibitem[{Lu et~al(2024)Lu, Han, Sun, and Yang}]{CATS}
Lu J, Han X, Sun Y, et~al (2024) Cats: Enhancing multivariate time series forecasting by constructing auxiliary time series as exogenous variables. International Conference on Machine Learning

\bibitem[{Lu~Han(2024)}]{CapacityRobustnessTrade-off}
Lu~Han DCZHan-Jia~Ye (2024) The capacity and robustness trade-off: Revisiting the channel independent strategy for multivariate time series forecasting. IEEE Transactions on Knowledge and Data Engineering pp 1--14

\bibitem[{Lundberg and Lee(2017)}]{lundberg2017unified}
Lundberg SM, Lee SI (2017) A unified approach to interpreting model predictions. In: Guyon I, Luxburg UV, Bengio S, et~al (eds) Advances in Neural Information Processing Systems, vol~30. Curran Associates, Inc., \urlprefix\url{https://proceedings.neurips.cc/paper_files/paper/2017/file/8a20a8621978632d76c43dfd28b67767-Paper.pdf}

\bibitem[{Luo and Wang(2024)}]{ModernTCN}
Luo D, Wang X (2024) Moderntcn: A modern pure convolution structure for general time series analysis. In: The Twelfth International Conference on Learning Representations

\bibitem[{Ma et~al(2024)Ma, Li, Fang, Zhao, and Zhang}]{U-Mixer}
Ma X, Li X, Fang L, et~al (2024) U-mixer: An unet-mixer architecture with stationarity correction for time series forecasting. In: Proceedings of the AAAI Conference on Artificial Intelligence, pp 14255--14262

\bibitem[{Ma et~al(2020)Ma, Guo, Ren, Tang, and Yin}]{DyGNN}
Ma Y, Guo Z, Ren Z, et~al (2020) Streaming graph neural networks. In: Proceedings of the 43rd international ACM SIGIR conference on research and development in information retrieval, pp 719--728

\bibitem[{Makridakis and Hibon(2000)}]{makridakis2000m3}
Makridakis S, Hibon M (2000) The m3-competition: results, conclusions and implications. International journal of forecasting 16(4):451--476

\bibitem[{Makridakis et~al(1982)Makridakis, Andersen, Carbone, Fildes, Hibon, Lewandowski, Newton, Parzen, and Winkler}]{makridakis1982accuracy}
Makridakis S, Andersen A, Carbone R, et~al (1982) The accuracy of extrapolation (time series) methods: Results of a forecasting competition. Journal of forecasting 1(2):111--153

\bibitem[{Makridakis et~al(2020)Makridakis, Spiliotis, and Assimakopoulos}]{makridakis2020m4}
Makridakis S, Spiliotis E, Assimakopoulos V (2020) The m4 competition: 100,000 time series and 61 forecasting methods. International Journal of Forecasting 36(1):54--74

\bibitem[{Marisca et~al(2024)Marisca, Alippi, and Bianchi}]{HD-TTS}
Marisca I, Alippi C, Bianchi FM (2024) Graph-based forecasting with missing data through spatiotemporal downsampling. International Conference on Machine Learning

\bibitem[{Masini et~al(2023)Masini, Medeiros, and Mendes}]{MLforTS}
Masini RP, Medeiros MC, Mendes EF (2023) Machine learning advances for time series forecasting. Journal of economic surveys 37(1):76--111

\bibitem[{Massaoudi et~al(2021)Massaoudi, Refaat, Chihi, Trabelsi, Oueslati, and Abu-Rub}]{LGBM-XGB-MLP}
Massaoudi M, Refaat SS, Chihi I, et~al (2021) A novel stacked generalization ensemble-based hybrid lgbm-xgb-mlp model for short-term load forecasting. Energy 214:118874

\bibitem[{Matheson and Winkler(1976)}]{matheson1976scoring}
Matheson JE, Winkler RL (1976) Scoring rules for continuous probability distributions. Management science 22(10):1087--1096

\bibitem[{McCracken and Ng(2016)}]{mccracken2016fred}
McCracken MW, Ng S (2016) Fred-md: A monthly database for macroeconomic research. Journal of Business \& Economic Statistics 34(4):574--589

\bibitem[{McLeod and Gweon(2013)}]{mcleod2013optimal}
McLeod A, Gweon H (2013) Optimal deseasonalization for monthly and daily geophysical time series. Journal of Environmental statistics 4(11):1--11

\bibitem[{Meijer and Chen(2024)}]{RiseofDiff}
Meijer C, Chen LY (2024) The rise of diffusion models in time-series forecasting. arXiv preprint arXiv:240103006

\bibitem[{Mounir et~al(2023)Mounir, Ouadi, and Jrhilifa}]{EMD-BI-LSTM}
Mounir N, Ouadi H, Jrhilifa I (2023) Short-term electric load forecasting using an emd-bi-lstm approach for smart grid energy management system. Energy and Buildings 288:113022

\bibitem[{Mu et~al(2023)Mu, Jiang, Yuan, Cui, and Qin}]{mu2023nao}
Mu B, Jiang X, Yuan S, et~al (2023) Nao seasonal forecast using a multivariate air--sea coupled deep learning model combined with causal discovery. Atmosphere 14(5):792

\bibitem[{Nason et~al(2000)Nason, Von~Sachs, and Kroisandt}]{nason2000wavelet}
Nason GP, Von~Sachs R, Kroisandt G (2000) Wavelet processes and adaptive estimation of the evolutionary wavelet spectrum. Journal of the Royal Statistical Society: Series B (Statistical Methodology) 62(2):271--292

\bibitem[{Nawrot et~al(2021)Nawrot, Tworkowski, Tyrolski, Kaiser, Wu, Szegedy, and Michalewski}]{nawrot2021hierarchical}
Nawrot P, Tworkowski S, Tyrolski M, et~al (2021) Hierarchical transformers are more efficient language models. arXiv preprint arXiv:211013711

\bibitem[{Ni et~al(2024)Ni, Yu, Liu, Li, and Lin}]{basisformer}
Ni Z, Yu H, Liu S, et~al (2024) Basisformer: Attention-based time series forecasting with learnable and interpretable basis. Advances in Neural Information Processing Systems 36

\bibitem[{Nie et~al(2024)Nie, Mei, Qin, Sun, and Ma}]{C-LoRA}
Nie T, Mei Y, Qin G, et~al (2024) Channel-aware low-rank adaptation in time series forecasting. Conference on Information and Knowledge Management

\bibitem[{Nie et~al(2023)Nie, Nguyen, Sinthong, and Kalagnanam}]{PatchTST}
Nie Y, Nguyen NH, Sinthong P, et~al (2023) A time series is worth 64 words: Long-term forecasting with transformers. In: The Eleventh International Conference on Learning Representations, \urlprefix\url{https://openreview.net/forum?id=Jbdc0vTOcol}

\bibitem[{Nti et~al(2020)Nti, Teimeh, Nyarko-Boateng, and Adekoya}]{ElectricityForecasting}
Nti IK, Teimeh M, Nyarko-Boateng O, et~al (2020) Electricity load forecasting: a systematic review. Journal of Electrical Systems and Information Technology 7:1--19

\bibitem[{Oord et~al(2018)Oord, Li, and Vinyals}]{CPC}
Oord Avd, Li Y, Vinyals O (2018) Representation learning with contrastive predictive coding. arXiv preprint arXiv:180703748

\bibitem[{Oreshkin et~al(2020)Oreshkin, Carpov, Chapados, and Bengio}]{N-BEATS}
Oreshkin BN, Carpov D, Chapados N, et~al (2020) N-beats: Neural basis expansion analysis for interpretable time series forecasting. In: International Conference on Learning Representations, \urlprefix\url{https://openreview.net/forum?id=r1ecqn4YwB}

\bibitem[{Pan and Yang(2009)}]{TransferLearning}
Pan SJ, Yang Q (2009) A survey on transfer learning. IEEE Transactions on knowledge and data engineering 22(10):1345--1359

\bibitem[{Papapetrou and Lee(2024)}]{papapetrou2024interpretable}
Papapetrou P, Lee Z (2024) Interpretable and explainable time series mining. In: 2024 IEEE 11th International Conference on Data Science and Advanced Analytics (DSAA), IEEE, pp 1--3

\bibitem[{Park et~al(2020)Park, Lee, Bahng, Tae, Jin, Kim, Ko, and Choo}]{park2020st}
Park C, Lee C, Bahng H, et~al (2020) St-grat: A novel spatio-temporal graph attention networks for accurately forecasting dynamically changing road speed. In: Proceedings of the 29th ACM international conference on information \& knowledge management, pp 1215--1224

\bibitem[{Passalis et~al(2019)Passalis, Tefas, Kanniainen, Gabbouj, and Iosifidis}]{DAIN}
Passalis N, Tefas A, Kanniainen J, et~al (2019) Deep adaptive input normalization for time series forecasting. IEEE transactions on neural networks and learning systems 31(9):3760--3765

\bibitem[{Patro and Agneeswaran(2024{\natexlab{a}})}]{mamba360}
Patro BN, Agneeswaran VS (2024{\natexlab{a}}) Mamba-360: Survey of state space models as transformer alternative for long sequence modelling: Methods, applications, and challenges. arXiv preprint arXiv:240416112

\bibitem[{Patro and Agneeswaran(2024{\natexlab{b}})}]{SiMBA}
Patro BN, Agneeswaran VS (2024{\natexlab{b}}) Simba: Simplified mamba-based architecture for vision and multivariate time series. arXiv preprint arXiv:240315360

\bibitem[{Patwardhan et~al(2023)Patwardhan, Marrone, and Sansone}]{patwardhan2023transformers}
Patwardhan N, Marrone S, Sansone C (2023) Transformers in the real world: A survey on nlp applications. Information 14(4):242

\bibitem[{Pavlyshenko(2020)}]{probstacking1}
Pavlyshenko BM (2020) Using bayesian regression for stacking time series predictive models. In: 2020 IEEE Third International Conference on Data Stream Mining \& Processing (DSMP), pp 305--309, \doi{10.1109/DSMP47368.2020.9204312}

\bibitem[{Pearl(1995)}]{Do-Calculus}
Pearl J (1995) Causal diagrams for empirical research. Biometrika 82(4):669--688

\bibitem[{Pearl(1998)}]{DAGs}
Pearl J (1998) Graphs, causality, and structural equation models. Sociological Methods \& Research 27(2):226--284

\bibitem[{Pearl et~al(2000)}]{SCM}
Pearl J, et~al (2000) Models, reasoning and inference. Cambridge, UK: CambridgeUniversityPress 19(2):3

\bibitem[{Peng et~al(2023)Peng, Zhang, and Tian}]{VMD_app}
Peng Zj, Zhang C, Tian Yx (2023) Crude oil price time series forecasting: a novel approach based on variational mode decomposition, time-series imaging, and deep learning. IEEE Access

\bibitem[{Petropoulos et~al(2018)Petropoulos, Hyndman, and Bergmeir}]{whyBaggingForTS}
Petropoulos F, Hyndman RJ, Bergmeir C (2018) Exploring the sources of uncertainty: Why does bagging for time series forecasting work? European Journal of Operational Research 268(2):545--554

\bibitem[{Piao et~al(2024)Piao, Chen, Murayama, Matsubara, and Sakurai}]{fredformer}
Piao X, Chen Z, Murayama T, et~al (2024) Fredformer: Frequency debiased transformer for time series forecasting. In: KDD, pp 2400--2410, \urlprefix\url{https://doi.org/10.1145/3637528.3671928}

\bibitem[{Portet(2020)}]{portet2020primer}
Portet S (2020) A primer on model selection using the akaike information criterion. Infectious Disease Modelling 5:111--128

\bibitem[{Qi et~al(2024)Qi, Wen, Li, Yang, Li, Rao, Pan, and Xu}]{InfoTime}
Qi S, Wen L, Li Y, et~al (2024) Enhancing multivariate time series forecasting with mutual information-driven cross-variable and temporal modeling. arXiv preprint arXiv:240300869

\bibitem[{Qian et~al(2023)Qian, Wang, Wu, Zhu, and Shi}]{qian2023causality}
Qian J, Wang Q, Wu Y, et~al (2023) Causality-based deep learning forecast of the kuroshio volume transport in the east china sea. Earth and Space Science 10(2):e2022EA002722

\bibitem[{Qin et~al(2017)Qin, Song, Chen, Cheng, Jiang, and Cottrell}]{DA-RNN}
Qin Y, Song D, Chen H, et~al (2017) A dual-stage attention-based recurrent neural network for time series prediction. In: Proceedings of the Twenty-Sixth International Joint Conference on Artificial Intelligence, {IJCAI-17}, pp 2627--2633, \doi{10.24963/ijcai.2017/366}, \urlprefix\url{https://doi.org/10.24963/ijcai.2017/366}

\bibitem[{Quinlan(1986)}]{quinlan1986induction}
Quinlan JR (1986) Induction of decision trees. Machine learning 1:81--106

\bibitem[{Radford et~al(2021)Radford, Kim, Hallacy, Ramesh, Goh, Agarwal, Sastry, Askell, Mishkin, Clark et~al}]{CLIP}
Radford A, Kim JW, Hallacy C, et~al (2021) Learning transferable visual models from natural language supervision. In: International conference on machine learning, PMLR, pp 8748--8763

\bibitem[{Ramesh et~al(2022)Ramesh, Dhariwal, Nichol, Chu, and Chen}]{DALL-E2}
Ramesh A, Dhariwal P, Nichol A, et~al (2022) Hierarchical text-conditional image generation with clip latents. arXiv preprint arXiv:220406125 1(2):3

\bibitem[{Rasul et~al(2021)Rasul, Seward, Schuster, and Vollgraf}]{TimeGrad}
Rasul K, Seward C, Schuster I, et~al (2021) Autoregressive denoising diffusion models for multivariate probabilistic time series forecasting. In: International Conference on Machine Learning, PMLR, pp 8857--8868

\bibitem[{Rasul et~al(2024)Rasul, Ashok, Williams, Ghonia, Bhagwatkar, Khorasani, Bayazi, Adamopoulos, Riachi, Hassen, Biloš, Garg, Schneider, Chapados, Drouin, Zantedeschi, Nevmyvaka, and Rish}]{LagLlama}
Rasul K, Ashok A, Williams AR, et~al (2024) Lag-llama: Towards foundation models for probabilistic time series forecasting. \eprint{2310.08278}

\bibitem[{Ribeiro et~al(2016)Ribeiro, Singh, and Guestrin}]{ribeiro2016should}
Ribeiro MT, Singh S, Guestrin C (2016) " why should i trust you?" explaining the predictions of any classifier. In: Proceedings of the 22nd ACM SIGKDD international conference on knowledge discovery and data mining, pp 1135--1144

\bibitem[{Rombach et~al(2022)Rombach, Blattmann, Lorenz, Esser, and Ommer}]{StableDiffusion}
Rombach R, Blattmann A, Lorenz D, et~al (2022) High-resolution image synthesis with latent diffusion models. In: Proceedings of the IEEE/CVF conference on computer vision and pattern recognition, pp 10684--10695

\bibitem[{Rosenbaum and Rubin(1983)}]{PSM}
Rosenbaum PR, Rubin DB (1983) The central role of the propensity score in observational studies for causal effects. Biometrika 70(1):41--55

\bibitem[{Rossi et~al(2020)Rossi, Chamberlain, Frasca, Eynard, Monti, and Bronstein}]{TGN}
Rossi E, Chamberlain B, Frasca F, et~al (2020) Temporal graph networks for deep learning on dynamic graphs. In: ICML 2020 Workshop on Graph Representation Learning

\bibitem[{Rumelhart et~al(1986)Rumelhart, Hinton, and Williams}]{MLP}
Rumelhart DE, Hinton GE, Williams RJ (1986) Learning representations by back-propagating errors. nature 323(6088):533--536

\bibitem[{Runge et~al(2019)Runge, Nowack, Kretschmer, Flaxman, and Sejdinovic}]{runge2019detecting}
Runge J, Nowack P, Kretschmer M, et~al (2019) Detecting and quantifying causal associations in large nonlinear time series datasets. Science advances 5(11):eaau4996

\bibitem[{Saharia et~al(2022)Saharia, Chan, Saxena, Li, Whang, Denton, Ghasemipour, Gontijo~Lopes, Karagol~Ayan, Salimans et~al}]{Imagen}
Saharia C, Chan W, Saxena S, et~al (2022) Photorealistic text-to-image diffusion models with deep language understanding. Advances in neural information processing systems 35:36479--36494

\bibitem[{{\c{S}}AHiN et~al(2024){\c{S}}AHiN, Arslan, and {\"O}zdemir}]{csahin2024unlocking}
{\c{S}}AHiN E, Arslan NN, {\"O}zdemir D (2024) Unlocking the black box: an in-depth review on interpretability, explainability, and reliability in deep learning. Neural Computing and Applications pp 1--107

\bibitem[{Sankaranarayanan et~al(2018)Sankaranarayanan, Balaji, Castillo, and Chellappa}]{GAN}
Sankaranarayanan S, Balaji Y, Castillo CD, et~al (2018) Generate to adapt: Aligning domains using generative adversarial networks. In: Proceedings of the IEEE conference on computer vision and pattern recognition, pp 8503--8512

\bibitem[{Scarselli et~al(2008)Scarselli, Gori, Tsoi, Hagenbuchner, and Monfardini}]{GNN}
Scarselli F, Gori M, Tsoi AC, et~al (2008) The graph neural network model. IEEE transactions on neural networks 20(1):61--80

\bibitem[{Sch{\"o}lkopf et~al(2021)Sch{\"o}lkopf, Locatello, Bauer, Ke, Kalchbrenner, Goyal, and Bengio}]{scholkopf2021toward}
Sch{\"o}lkopf B, Locatello F, Bauer S, et~al (2021) Toward causal representation learning. Proceedings of the IEEE 109(5):612--634

\bibitem[{Shabani et~al(2023)Shabani, Abdi, Meng, and Sylvain}]{scaleformer}
Shabani MA, Abdi AH, Meng L, et~al (2023) Scaleformer: Iterative multi-scale refining transformers for time series forecasting. In: The Eleventh International Conference on Learning Representations, \urlprefix\url{https://openreview.net/forum?id=sCrnllCtjoE}

\bibitem[{Shafer(1976)}]{evidencetheory}
Shafer G (1976) A Mathematical Theory of Evidence. Princeton University Press

\bibitem[{Sharma et~al(2022)Sharma, Dwivedi, and Metri}]{sharma2022incorporating}
Sharma K, Dwivedi YK, Metri B (2022) Incorporating causality in energy consumption forecasting using deep neural networks. Annals of Operations Research pp 1--36

\bibitem[{Sharma et~al(2021)Sharma, Mangla, Mohanty, and Pattanaik}]{stackedEnsemble}
Sharma N, Mangla M, Mohanty SN, et~al (2021) Employing stacked ensemble approach for time series forecasting. International Journal of Information Technology 13:2075--2080

\bibitem[{Shen and Kwok(2023)}]{TimeDiff}
Shen L, Kwok J (2023) Non-autoregressive conditional diffusion models for time series prediction. In: International Conference on Machine Learning, PMLR, pp 31016--31029

\bibitem[{Shen et~al(2024)Shen, Chen, and Kwok}]{mr-Diff}
Shen L, Chen W, Kwok J (2024) Multi-resolution diffusion models for time series forecasting. In: The Twelfth International Conference on Learning Representations

\bibitem[{Shi et~al(2015)Shi, Chen, Wang, Yeung, Wong, and Woo}]{ConvLSTM}
Shi X, Chen Z, Wang H, et~al (2015) Convolutional lstm network: A machine learning approach for precipitation nowcasting. Advances in neural information processing systems 28

\bibitem[{Shih et~al(2019)Shih, Sun, and Lee}]{TPA-LSTM}
Shih SY, Sun FK, Lee Hy (2019) Temporal pattern attention for multivariate time series forecasting. Machine Learning 108:1421--1441

\bibitem[{Shu and Lampos(2024)}]{DeformTime}
Shu Y, Lampos V (2024) Deformtime: Capturing variable dependencies with deformable attention for time series forecasting. arXiv preprint arXiv:240607438

\bibitem[{Smyl(2020)}]{ESRNN}
Smyl S (2020) A hybrid method of exponential smoothing and recurrent neural networks for time series forecasting. International journal of forecasting 36(1):75--85

\bibitem[{Smyl et~al(2024)Smyl, Oreshkin, Pe{\l}ka, and Dudek}]{smyl2024any}
Smyl S, Oreshkin BN, Pe{\l}ka P, et~al (2024) Any-quantile probabilistic forecasting of short-term electricity demand. arXiv preprint arXiv:240417451

\bibitem[{Sohl-Dickstein et~al(2015)Sohl-Dickstein, Weiss, Maheswaranathan, and Ganguli}]{sohl2015deep}
Sohl-Dickstein J, Weiss E, Maheswaranathan N, et~al (2015) Deep unsupervised learning using nonequilibrium thermodynamics. In: International conference on machine learning, PMLR, pp 2256--2265

\bibitem[{Song et~al(2020)Song, Lin, Guo, and Wan}]{STSGCN}
Song C, Lin Y, Guo S, et~al (2020) Spatial-temporal synchronous graph convolutional networks: A new framework for spatial-temporal network data forecasting. In: Proceedings of the AAAI conference on artificial intelligence, pp 914--921

\bibitem[{Song and Ermon(2019)}]{song2019generative}
Song Y, Ermon S (2019) Generative modeling by estimating gradients of the data distribution. Advances in neural information processing systems 32

\bibitem[{Song et~al(2021)Song, Sohl-Dickstein, Kingma, Kumar, Ermon, and Poole}]{SDE}
Song Y, Sohl-Dickstein J, Kingma DP, et~al (2021) Score-based generative modeling through stochastic differential equations. In: International Conference on Learning Representations, \urlprefix\url{https://openreview.net/forum?id=PxTIG12RRHS}

\bibitem[{Soyiri and Reidpath(2013)}]{healthForecasting}
Soyiri IN, Reidpath DD (2013) An overview of health forecasting. Environmental health and preventive medicine 18:1--9

\bibitem[{Sparks et~al(2020)Sparks, Carroll, Goldie, Marchiori, Melloy, Padgham, Parsonage, Pembleton, and Balamuta}]{sparks2020bomrang}
Sparks AH, Carroll J, Goldie J, et~al (2020) Bomrang: Australian government bureau of meteorology (bom) data client. R package version 07 0 URL https://CRAN R-project org/package= bomrang

\bibitem[{Srivastava et~al(2014)Srivastava, Hinton, Krizhevsky, Sutskever, and Salakhutdinov}]{Dropout}
Srivastava N, Hinton G, Krizhevsky A, et~al (2014) Dropout: a simple way to prevent neural networks from overfitting. The journal of machine learning research 15(1):1929--1958

\bibitem[{{\v{S}}t{\v{e}}pni{\v{c}}ka and Burda(2017)}]{vstvepnivcka2017results}
{\v{S}}t{\v{e}}pni{\v{c}}ka M, Burda M (2017) On the results and observations of the time series forecasting competition cif 2016. In: 2017 IEEE International Conference on Fuzzy Systems (FUZZ-IEEE), IEEE, pp 1--6

\bibitem[{Sui et~al(2024)Sui, Zhang, Zhou, Liao, and Wei}]{ensemble_stock}
Sui M, Zhang C, Zhou L, et~al (2024) An ensemble approach to stock price prediction using deep learning and time series models. In: 2024 IEEE 6th International Conference on Power, Intelligent Computing and Systems (ICPICS), IEEE, pp 793--797

\bibitem[{Sun et~al(2024{\natexlab{a}})Sun, Hao, Zou, and Shen}]{sun2024survey}
Sun F, Hao W, Zou A, et~al (2024{\natexlab{a}}) A survey on spatio-temporal series prediction with deep learning: taxonomy, applications, and future directions. Neural Computing and Applications pp 1--25

\bibitem[{Sun et~al(2024{\natexlab{b}})Sun, Xie, Chen, Eldele, and Hu}]{HCAN}
Sun Y, Xie Z, Chen D, et~al (2024{\natexlab{b}}) Hierarchical classification auxiliary network for time series forecasting. arXiv preprint arXiv:240518975

\bibitem[{Taieb et~al(2012)Taieb, Bontempi, Atiya, and Sorjamaa}]{taieb2012review}
Taieb SB, Bontempi G, Atiya AF, et~al (2012) A review and comparison of strategies for multi-step ahead time series forecasting based on the nn5 forecasting competition. Expert systems with applications 39(8):7067--7083

\bibitem[{Tang and Zhang(2024)}]{PDMLP}
Tang P, Zhang W (2024) Pdmlp: Patch-based decomposed mlp for long-term time series forecastin. arXiv preprint arXiv:240513575

\bibitem[{Tashiro et~al(2021)Tashiro, Song, Song, and Ermon}]{CSDI}
Tashiro Y, Song J, Song Y, et~al (2021) Csdi: Conditional score-based diffusion models for probabilistic time series imputation. Advances in Neural Information Processing Systems 34:24804--24816

\bibitem[{Tay et~al(2020)Tay, Dehghani, Bahri, and Metzler}]{EfficientTransformers}
Tay Y, Dehghani M, Bahri D, et~al (2020) Efficient transformers: A survey. ACM Computing Surveys 55:1 -- 28. \urlprefix\url{https://api.semanticscholar.org/CorpusID:221702858}

\bibitem[{Thompson(1990)}]{thompson1990mse}
Thompson PA (1990) An mse statistic for comparing forecast accuracy across series. International Journal of Forecasting 6(2):219--227

\bibitem[{Touvron et~al(2023)Touvron, Martin, Stone, Albert, Almahairi, Babaei, Bashlykov, Batra, Bhargava, Bhosale et~al}]{Llama2}
Touvron H, Martin L, Stone K, et~al (2023) Llama 2: Open foundation and fine-tuned chat models. arXiv preprint arXiv:230709288

\bibitem[{Truchan et~al(2024)Truchan, Kalfar, and Ahmadi}]{ltboost}
Truchan H, Kalfar C, Ahmadi Z (2024) Ltboost: Boosted hybrids of ensemble linear and gradient algorithms for the long-term time series forecasting. In: Proceedings of the 33rd ACM International Conference on Information and Knowledge Management, pp 2271--2281

\bibitem[{Ulyanov et~al(2016)Ulyanov, Vedaldi, and Lempitsky}]{InstanceNorm}
Ulyanov D, Vedaldi A, Lempitsky V (2016) Instance normalization: The missing ingredient for fast stylization. arXiv preprint arXiv:160708022

\bibitem[{Van Den~Oord et~al(2016)Van Den~Oord, Dieleman, Zen, Simonyan, Vinyals, Graves, Kalchbrenner, Senior, Kavukcuoglu et~al}]{WaveNet}
Van Den~Oord A, Dieleman S, Zen H, et~al (2016) Wavenet: A generative model for raw audio. arXiv preprint arXiv:160903499 12

\bibitem[{Vapnik et~al(1996)Vapnik, Golowich, and Smola}]{vapnik1996support}
Vapnik V, Golowich S, Smola A (1996) Support vector method for function approximation, regression estimation and signal processing. Advances in neural information processing systems 9

\bibitem[{Vaswani et~al(2017)Vaswani, Shazeer, Parmar, Uszkoreit, Jones, Gomez, Kaiser, and Polosukhin}]{Transformer}
Vaswani A, Shazeer N, Parmar N, et~al (2017) Attention is all you need. Advances in neural information processing systems 30

\bibitem[{Veličković et~al(2018)Veličković, Cucurull, Casanova, Romero, Liò, and Bengio}]{GAT}
Veličković P, Cucurull G, Casanova A, et~al (2018) Graph attention networks. In: International Conference on Learning Representations, \urlprefix\url{https://openreview.net/forum?id=rJXMpikCZ}

\bibitem[{Wachter et~al(2017)Wachter, Mittelstadt, and Russell}]{wachter2017counterfactual}
Wachter S, Mittelstadt B, Russell C (2017) Counterfactual explanations without opening the black box: Automated decisions and the gdpr. Harv JL \& Tech 31:841

\bibitem[{Wang et~al(2022{\natexlab{a}})Wang, Jatowt, and Yoshikawa}]{TimeBERT}
Wang J, Jatowt A, Yoshikawa M (2022{\natexlab{a}}) Timebert: Extending pre-trained language representations with temporal information. arXiv preprint arXiv:220413032

\bibitem[{Wang et~al(2022{\natexlab{b}})Wang, Adiga, Chen, Sadilek, Venkatramanan, and Marathe}]{wang2022causalgnn}
Wang L, Adiga A, Chen J, et~al (2022{\natexlab{b}}) Causalgnn: Causal-based graph neural networks for spatio-temporal epidemic forecasting. In: Proceedings of the AAAI conference on artificial intelligence, pp 12191--12199

\bibitem[{Wang et~al(2024{\natexlab{a}})Wang, Wu, Shi, Hu, Luo, Ma, Zhang, and Zhou}]{TimeMixer}
Wang S, Wu H, Shi X, et~al (2024{\natexlab{a}}) Timemixer: Decomposable multiscale mixing for time series forecasting. International Conference on Learning Representations

\bibitem[{Wang et~al(2023{\natexlab{a}})Wang, Li, Ding, Nie, Chen, Dong, and Wang}]{MMD}
Wang W, Li H, Ding Z, et~al (2023{\natexlab{a}}) Rethinking maximum mean discrepancy for visual domain adaptation. IEEE Transactions on Neural Networks and Learning Systems 34(1):264--277. \doi{10.1109/TNNLS.2021.3093468}

\bibitem[{Wang et~al(2024{\natexlab{b}})Wang, Zhou, Wen, Gao, Ding, and Jin}]{CARD}
Wang X, Zhou T, Wen Q, et~al (2024{\natexlab{b}}) Card: Channel aligned robust blend transformer for time series forecasting. In: The Twelfth International Conference on Learning Representations

\bibitem[{Wang et~al(2024{\natexlab{c}})Wang, Wu, Dong, Qin, Zhang, Liu, Qiu, Wang, and Long}]{TimeXer}
Wang Y, Wu H, Dong J, et~al (2024{\natexlab{c}}) Timexer: Empowering transformers for time series forecasting with exogenous variables. In: The Thirty-eighth Annual Conference on Neural Information Processing Systems, \urlprefix\url{https://openreview.net/forum?id=INAeUQ04lT}

\bibitem[{Wang et~al(2023{\natexlab{b}})Wang, Miliou, Samsten, and Papapetrou}]{wang2023counterfactual}
Wang Z, Miliou I, Samsten I, et~al (2023{\natexlab{b}}) Counterfactual explanations for time series forecasting. In: 2023 IEEE International Conference on Data Mining (ICDM), IEEE, pp 1391--1396

\bibitem[{Wang et~al(2024{\natexlab{d}})Wang, Kong, Feng, Wang, Zhao, Wang, and Zhang}]{S-Mamba}
Wang Z, Kong F, Feng S, et~al (2024{\natexlab{d}}) Is mamba effective for time series forecasting? arXiv preprint arXiv:240311144

\bibitem[{Wang et~al(2024{\natexlab{e}})Wang, Ruan, Huang, Zhou, Zhang, Wang, Wang, Huang, and Liu}]{TSP}
Wang Z, Ruan S, Huang T, et~al (2024{\natexlab{e}}) A lightweight multi-layer perceptron for efficient multivariate time series forecasting. Knowledge-Based Systems 288:111463

\bibitem[{Wei et~al(2022)Wei, Niu, Tang, Wang, Hu, and Liang}]{wei2022npenas}
Wei C, Niu C, Tang Y, et~al (2022) Npenas: Neural predictor guided evolution for neural architecture search. IEEE Transactions on Neural Networks and Learning Systems 34(11):8441--8455

\bibitem[{Wen et~al(2023)Wen, Zhou, Zhang, Chen, Ma, Yan, and Sun}]{TrasformersInTS}
Wen Q, Zhou T, Zhang C, et~al (2023) Transformers in time series: A survey. In: Elkind E (ed) Proceedings of the Thirty-Second International Joint Conference on Artificial Intelligence, {IJCAI-23}. International Joint Conferences on Artificial Intelligence Organization, pp 6778--6786, \doi{10.24963/ijcai.2023/759}, \urlprefix\url{https://doi.org/10.24963/ijcai.2023/759}, survey Track

\bibitem[{Wen et~al(2017)Wen, Torkkola, Narayanaswamy, and Madeka}]{MQ-RNN}
Wen R, Torkkola K, Narayanaswamy B, et~al (2017) A multi-horizon quantile recurrent forecaster. arXiv preprint arXiv:171111053

\bibitem[{Winters(1960)}]{winters1960forecasting}
Winters PR (1960) Forecasting sales by exponentially weighted moving averages. Management science 6(3):324--342

\bibitem[{Wolpert(1992)}]{wolpert1992stacked}
Wolpert DH (1992) Stacked generalization. Neural networks 5(2):241--259

\bibitem[{Wong et~al(2016)Wong, Gatt, Stamatescu, and McDonnell}]{DataAugmentation}
Wong SC, Gatt A, Stamatescu V, et~al (2016) Understanding data augmentation for classification: when to warp? In: 2016 international conference on digital image computing: techniques and applications (DICTA), IEEE, pp 1--6

\bibitem[{Woo et~al(2024)Woo, Liu, Kumar, Xiong, Savarese, and Sahoo}]{Uni2TS}
Woo G, Liu C, Kumar A, et~al (2024) Unified training of universal time series forecasting transformers. In: Forty-first International Conference on Machine Learning

\bibitem[{Wu and Wang(2024)}]{wu2024two}
Wu B, Wang L (2024) Two-stage decomposition and temporal fusion transformers for interpretable wind speed forecasting. Energy 288:129728

\bibitem[{Wu et~al(2024{\natexlab{a}})Wu, Yu, Peng, and Wang}]{wu2024interpretable}
Wu B, Yu S, Peng L, et~al (2024{\natexlab{a}}) Interpretable wind speed forecasting with meteorological feature exploring and two-stage decomposition. Energy 294:130782

\bibitem[{Wu et~al(2021)Wu, Xu, Wang, and Long}]{Autoformer}
Wu H, Xu J, Wang J, et~al (2021) Autoformer: Decomposition transformers with auto-correlation for long-term series forecasting. Advances in neural information processing systems 34:22419--22430

\bibitem[{Wu et~al(2023)Wu, Hu, Liu, Zhou, Wang, and Long}]{Timesnet}
Wu H, Hu T, Liu Y, et~al (2023) Timesnet: Temporal 2d-variation modeling for general time series analysis. International Conference on Learning Representations

\bibitem[{Wu and Huang(2009)}]{EEMD}
Wu Z, Huang NE (2009) Ensemble empirical mode decomposition: a noise-assisted data analysis method. Advances in adaptive data analysis 1(01):1--41

\bibitem[{Wu et~al(2024{\natexlab{b}})Wu, Gong, and Zhang}]{DTMamba}
Wu Z, Gong Y, Zhang A (2024{\natexlab{b}}) Dtmamba: Dual twin mamba for time series forecasting. arXiv preprint arXiv:240507022

\bibitem[{Xiong et~al(2024)Xiong, Tang, Ma, Xu, and Li}]{TDTLoss}
Xiong Q, Tang K, Ma M, et~al (2024) Tdt loss takes it all: Integrating temporal dependencies among targets into non-autoregressive time series forecasting. arXiv preprint arXiv:240604777

\bibitem[{Xu et~al(2019)Xu, Sheng, Li, Cheng, and Wu}]{EmdModeMixing}
Xu B, Sheng Y, Li P, et~al (2019) Causes and classification of emd mode mixing. Vibroengineering Procedia 22:158--164

\bibitem[{Xu et~al(2022)Xu, Chen, Xu, Zhou, Yu, Zheng, Xuan, and Yang}]{SignalMix}
Xu X, Chen Z, Xu D, et~al (2022) Mixing signals: Data augmentation approach for deep learning based modulation recognition. arXiv preprint arXiv:220403737

\bibitem[{Xu et~al(2024{\natexlab{a}})Xu, Liang, Huang, Lan, and Shu}]{Mambaformer}
Xu X, Liang Y, Huang B, et~al (2024{\natexlab{a}}) Integrating mamba and transformer for long-short range time series forecasting. arXiv preprint arXiv:240414757

\bibitem[{Xu et~al(2024{\natexlab{b}})Xu, Bian, Zhong, Wen, and Xu}]{TGTSF}
Xu Z, Bian Y, Zhong J, et~al (2024{\natexlab{b}}) Beyond trend and periodicity: Guiding time series forecasting with textual cues. arXiv preprint arXiv:240513522

\bibitem[{Xu et~al(2024{\natexlab{c}})Xu, Zeng, and Xu}]{FITS}
Xu Z, Zeng A, Xu Q (2024{\natexlab{c}}) Fits: Modeling time series with $10 k $ parameters. International Conference on Learning Representations

\bibitem[{Xua and Yang(2024)}]{xua2024interpretability}
Xua B, Yang G (2024) Interpretability research of deep learning: A literature survey. Information Fusion p 102721

\bibitem[{Xue and Salim(2023)}]{Promptcast}
Xue H, Salim FD (2023) Promptcast: A new prompt-based learning paradigm for time series forecasting. IEEE Transactions on Knowledge and Data Engineering

\bibitem[{Yan et~al(2018)Yan, Xiong, and Lin}]{ST-GCN}
Yan S, Xiong Y, Lin D (2018) Spatial temporal graph convolutional networks for skeleton-based action recognition. In: Proceedings of the AAAI conference on artificial intelligence

\bibitem[{Yan et~al(2021)Yan, Zhang, Zhou, Zhan, and Xia}]{Scoregrad}
Yan T, Zhang H, Zhou T, et~al (2021) Scoregrad: Multivariate probabilistic time series forecasting with continuous energy-based generative models. arXiv preprint arXiv:210610121

\bibitem[{Yan et~al(2024)Yan, Gong, Yongping, Zhan, and Xia}]{D3M}
Yan T, Gong H, Yongping H, et~al (2024) Probabilistic time series modeling with decomposable denoising diffusion model. In: Salakhutdinov R, Kolter Z, Heller K, et~al (eds) Proceedings of the 41st International Conference on Machine Learning, Proceedings of Machine Learning Research, vol 235. PMLR, pp 55759--55777, \urlprefix\url{https://proceedings.mlr.press/v235/yan24b.html}

\bibitem[{Yang et~al(2023{\natexlab{a}})Yang, Li, Wang, Zang, Pang, and Wang}]{WaveForM}
Yang F, Li X, Wang M, et~al (2023{\natexlab{a}}) Waveform: Graph enhanced wavelet learning for long sequence forecasting of multivariate time series. In: Proceedings of the AAAI Conference on Artificial Intelligence, pp 10754--10761

\bibitem[{Yang et~al(2023{\natexlab{b}})Yang, Xu, Bai, Ma, and Su}]{yang2023investigating}
Yang M, Xu C, Bai Y, et~al (2023{\natexlab{b}}) Investigating black-box model for wind power forecasting using local interpretable model-agnostic explanations algorithm: Why should a model be trusted? CSEE Journal of Power and Energy Systems

\bibitem[{Yang and Yang(2020)}]{EEMD_app}
Yang Y, Yang Y (2020) Hybrid method for short-term time series forecasting based on eemd. IEEE Access 8:61915--61928

\bibitem[{Yang et~al(2024{\natexlab{a}})Yang, Jin, Wen, Zhang, Liang, Ma, Wang, Liu, Yang, Xu et~al}]{yang2024survey}
Yang Y, Jin M, Wen H, et~al (2024{\natexlab{a}}) A survey on diffusion models for time series and spatio-temporal data. arXiv preprint arXiv:240418886

\bibitem[{Yang et~al(2024{\natexlab{b}})Yang, Zhu, and Chen}]{VCformer}
Yang Y, Zhu Q, Chen J (2024{\natexlab{b}}) Vcformer: Variable correlation transformer with inherent lagged correlation for multivariate time series forecasting. In: Larson K (ed) Proceedings of the Thirty-Third International Joint Conference on Artificial Intelligence, {IJCAI-24}. International Joint Conferences on Artificial Intelligence Organization, pp 5335--5343, \doi{10.24963/ijcai.2024/590}, \urlprefix\url{https://doi.org/10.24963/ijcai.2024/590}, main Track

\bibitem[{Ye et~al(2024)Ye, Zhang, Yi, Yu, Li, Li, and Tsung}]{surveyOfTSFM}
Ye J, Zhang W, Yi K, et~al (2024) A survey of time series foundation models: Generalizing time series representation with large language mode. arXiv preprint arXiv:240502358

\bibitem[{Yi et~al(2024)Yi, Zhang, Fan, Wang, Wang, He, An, Lian, Cao, and Niu}]{FreTS}
Yi K, Zhang Q, Fan W, et~al (2024) Frequency-domain mlps are more effective learners in time series forecasting. Advances in Neural Information Processing Systems 36

\bibitem[{Yu et~al(2023)Yu, Wang, Shao, Sun, Wu, and Xu}]{DSformer}
Yu C, Wang F, Shao Z, et~al (2023) Dsformer: A double sampling transformer for multivariate time series long-term prediction. In: Proceedings of the 32nd ACM international conference on information and knowledge management, pp 3062--3072

\bibitem[{Yu et~al(2024)Yu, Zou, Hu, Aviles-Rivero, Qin, and Wang}]{Leddam}
Yu G, Zou J, Hu X, et~al (2024) Revitalizing multivariate time series forecasting: Learnable decomposition with inter-series dependencies and intra-series variations modeling. International Conference on Machine Learning

\bibitem[{Yuan and Qiao(2024)}]{Diffusion-ts}
Yuan X, Qiao Y (2024) Diffusion-{TS}: Interpretable diffusion for general time series generation. In: The Twelfth International Conference on Learning Representations, \urlprefix\url{https://openreview.net/forum?id=4h1apFjO99}

\bibitem[{Zeng et~al(2023)Zeng, Chen, Zhang, and Xu}]{DLinear}
Zeng A, Chen M, Zhang L, et~al (2023) Are transformers effective for time series forecasting? In: Proceedings of the AAAI conference on artificial intelligence, pp 11121--11128

\bibitem[{Zeng et~al(2024)Zeng, Liu, Zheng, and Kong}]{C-Mamba}
Zeng C, Liu Z, Zheng G, et~al (2024) C-mamba: Channel correlation enhanced state space models for multivariate time series forecasting. arXiv preprint arXiv:240605316

\bibitem[{Zhan et~al(2024)Zhan, He, Li, and Deng}]{TEFN}
Zhan T, He Y, Li Z, et~al (2024) Time evidence fusion network: Multi-source view in long-term time series forecasting. arXiv preprint arXiv:240506419

\bibitem[{Zhang and Wang(2024)}]{ACNet}
Zhang D, Wang Y (2024) Adaptive extraction network for multivariate long sequence time-series forecasting. arXiv preprint arXiv:240512038

\bibitem[{Zhang et~al(2017)Zhang, Zheng, and Qi}]{ST-ResNet}
Zhang J, Zheng Y, Qi D (2017) Deep spatio-temporal residual networks for citywide crowd flows prediction. In: Proceedings of the AAAI conference on artificial intelligence

\bibitem[{Zhang et~al(2024{\natexlab{a}})Zhang, Wen, Zhang, Cai, Jin, Liu, Zhang, Liang, Pang, Song et~al}]{zhang2024self}
Zhang K, Wen Q, Zhang C, et~al (2024{\natexlab{a}}) Self-supervised learning for time series analysis: Taxonomy, progress, and prospects. IEEE Transactions on Pattern Analysis and Machine Intelligence

\bibitem[{Zhang et~al(2024{\natexlab{b}})Zhang, Zou, and Tang}]{zhang2024caformer}
Zhang K, Zou X, Tang Y (2024{\natexlab{b}}) Caformer: Rethinking time series analysis from causal perspective. arXiv preprint arXiv:240308572

\bibitem[{Zhang et~al(2024{\natexlab{c}})Zhang, Sun, and Liang}]{PA-RNN}
Zhang M, Sun Y, Liang F (2024{\natexlab{c}}) Sparse deep learning for time series data: theory and applications. Advances in Neural Information Processing Systems 36

\bibitem[{Zhang et~al(2020)Zhang, Chang, Meng, Xiang, and Pan}]{SLCNN}
Zhang Q, Chang J, Meng G, et~al (2020) Spatio-temporal graph structure learning for traffic forecasting. In: Proceedings of the AAAI conference on artificial intelligence, pp 1177--1185

\bibitem[{Zhang et~al(2021{\natexlab{a}})Zhang, Chen, Xiao, Zhang, and Feng}]{zhang2021hybrid}
Zhang S, Chen Y, Xiao J, et~al (2021{\natexlab{a}}) Hybrid wind speed forecasting model based on multivariate data secondary decomposition approach and deep learning algorithm with attention mechanism. Renewable Energy 174:688--704

\bibitem[{Zhang et~al(2021{\natexlab{b}})Zhang, Chen, Zhang, and Feng}]{zhang2021novel}
Zhang S, Chen Y, Zhang W, et~al (2021{\natexlab{b}}) A novel ensemble deep learning model with dynamic error correction and multi-objective ensemble pruning for time series forecasting. Information Sciences 544:427--445

\bibitem[{Zhang et~al(2024{\natexlab{d}})Zhang, Chowdhury, Gupta, and Shang}]{LLMforTSsurvey}
Zhang X, Chowdhury RR, Gupta RK, et~al (2024{\natexlab{d}}) Large language models for time series: A survey. In: Larson K (ed) Proceedings of the Thirty-Third International Joint Conference on Artificial Intelligence, {IJCAI-24}. International Joint Conferences on Artificial Intelligence Organization, pp 8335--8343, \doi{10.24963/ijcai.2024/921}, \urlprefix\url{https://doi.org/10.24963/ijcai.2024/921}, survey Track

\bibitem[{Zhang and Yan(2023)}]{Crossformer}
Zhang Y, Yan J (2023) Crossformer: Transformer utilizing cross-dimension dependency for multivariate time series forecasting. In: The eleventh international conference on learning representations

\bibitem[{Zhang et~al(2024{\natexlab{e}})Zhang, Ma, Pal, Zhang, and Coates}]{MTST}
Zhang Y, Ma L, Pal S, et~al (2024{\natexlab{e}}) Multi-resolution time-series transformer for long-term forecasting. In: International Conference on Artificial Intelligence and Statistics, PMLR, pp 4222--4230

\bibitem[{Zhao et~al(2020)Zhao, Ma, and Ermon}]{zhao2020individual}
Zhao S, Ma T, Ermon S (2020) Individual calibration with randomized forecasting. In: International Conference on Machine Learning, PMLR, pp 11387--11397

\bibitem[{Zheng et~al(2020)Zheng, Fan, Wang, and Qi}]{GMAN}
Zheng C, Fan X, Wang C, et~al (2020) Gman: A graph multi-attention network for traffic prediction. In: Proceedings of the AAAI conference on artificial intelligence, pp 1234--1241

\bibitem[{Zhou et~al(2021)Zhou, Zhang, Peng, Zhang, Li, Xiong, and Zhang}]{Informer}
Zhou H, Zhang S, Peng J, et~al (2021) Informer: Beyond efficient transformer for long sequence time-series forecasting. In: Proceedings of the AAAI conference on artificial intelligence, pp 11106--11115

\bibitem[{Zhou et~al(2022)Zhou, Ma, Wen, Wang, Sun, and Jin}]{Fedformer}
Zhou T, Ma Z, Wen Q, et~al (2022) Fedformer: Frequency enhanced decomposed transformer for long-term series forecasting. In: International conference on machine learning, PMLR, pp 27268--27286

\bibitem[{Zhou et~al(2023)Zhou, Niu, Wang, Sun, and Jin}]{OneFitsAll}
Zhou T, Niu P, Wang X, et~al (2023) One fits all: Power general time series analysis by pretrained {LM}. In: Thirty-seventh Conference on Neural Information Processing Systems, \urlprefix\url{https://openreview.net/forum?id=gMS6FVZvmF}

\end{thebibliography}



\end{document}